%% file: main.tex
\documentclass{article}

\usepackage{microtype}
\usepackage{graphicx}
\usepackage{subfigure}
\usepackage{booktabs} %

\usepackage{hyperref}

\usepackage[accepted]{icml2023}

\usepackage{amsmath}
\usepackage{amssymb}
\usepackage{mathtools}
\usepackage{amsthm}

\usepackage[capitalize,noabbrev]{cleveref}

\theoremstyle{plain}
\newtheorem{theorem}{Theorem}[section]

\theoremstyle{definition}
\newtheorem{definition}[theorem]{Definition}

\theoremstyle{remark}

\usepackage[textsize=tiny]{todonotes}

\icmltitlerunning{The Dormant Neuron Phenomenon in Deep Reinforcement Learning}

\begin{document}

\twocolumn[
\icmltitle{The Dormant Neuron Phenomenon in Deep Reinforcement Learning}

\icmlsetsymbol{equal}{*}

\begin{icmlauthorlist}
\icmlauthor{Ghada Sokar}{uni,go}
\icmlauthor{Rishabh Agarwal}{comp,mila}
\icmlauthor{Pablo Samuel Castro}{comp,equal}
\icmlauthor{Utku Evci}{comp,equal}
\end{icmlauthorlist}

\icmlaffiliation{uni}{Eindhoven University of Technology, The Netherlands}
\icmlaffiliation{go}{Work done while the author was intern at Google DeepMind}
\icmlaffiliation{comp}{Google DeepMind}
\icmlaffiliation{mila}{Mila}

\icmlcorrespondingauthor{Ghada Sokar}{g.a.z.n.sokar@tue.nl}
\icmlcorrespondingauthor{Rishabh Agarwal}{
rishabhagarwal@google.com}
\icmlcorrespondingauthor{Pablo Samuel Castro}{psc@google.com}
\icmlcorrespondingauthor{Utku Evci}{evci@google.com}

\icmlkeywords{Machine Learning, ICML}

\vskip 0.3in
]

\printAffiliationsAndNotice{\icmlEqualContribution} %

\begin{abstract} 
In this work we identify the {\em dormant neuron phenomenon} in deep reinforcement learning, where an agent's network suffers from an increasing number of inactive neurons, thereby affecting network expressivity. We demonstrate the presence of this phenomenon across a variety of algorithms and environments, and highlight its effect on learning. To address this issue, we propose a simple and effective method ({\em ReDo}) that {\em Re}cycles {\em Do}rmant neurons throughout training. Our experiments demonstrate that {\em ReDo} maintains the expressive power of networks by reducing the number of dormant neurons and results in improved performance.
\end{abstract}
\input{1.introduction.tex}
\input{2.background.tex}
\input{3.analysis.tex}

\input{4.method.tex}

\input{5.results.tex}

\input{6.related_work.tex}

\input{7.conclusion}

\paragraph{Societal impact.} Although the work presented here is mostly of an academic nature, it aids in the development of more capable autonomous agents. While our contributions do not directly contribute to any negative societal impacts, we urge the community to consider these when building on our research.

\input{acknowledgements}

\bibliography{main}
\bibliographystyle{icml2023}

\newpage
\appendix
\onecolumn
\input{8.appendix.tex}

\end{document}

%% file: 1.introduction.tex
\section{Introduction}

The use of deep neural networks as function approximators for value-based reinforcement learning (RL) has been one of the core elements that has enabled scaling RL to complex decision-making problems \cite{mnih2015human,silver2016mastering,bellemare2020autonomous}. However, their use can lead to training difficulties that are not present in traditional RL settings. Numerous improvements have been integrated with RL methods to address training instability, such as the use of target networks, prioritized experience replay, multi-step targets, among others \cite{hessel2018rainbow}. In parallel, there have been recent efforts devoted to better understanding the behavior of deep neural networks under the learning dynamics of RL \cite{Hasselt2018deadlytriad,fu2019diagnosing,kumar2020implicit,bengio2020interference,lyle2021understanding,araujo2021lifting}.

Recent work in so-called ``scaling laws'' for supervised learning problems suggest that, in these settings, there is a positive correlation between performance and the number of parameters \cite{hestness2017deep,kaplan2020scaling,zhai2022scaling}. In RL, however, there is evidence that the networks {\em lose} their expressivity and ability to fit new targets over time, despite being over-parameterized \citep{kumar2020implicit,lyle2021understanding}; this issue has been partly mitigated by perturbing the learned parameters. \citet{igl2020transient} and \citet{nikishin2022primacy} periodically \textit{reset} some, or all, of the layers of an agent’s neural networks, leading to improved performance. These approaches, however, are somewhat drastic: reinitializing the weights can cause the network to ``forget'' previously learned knowledge and require many gradient updates to recover.

\begin{figure}[t]
    \centering
    \includegraphics[width=\linewidth]{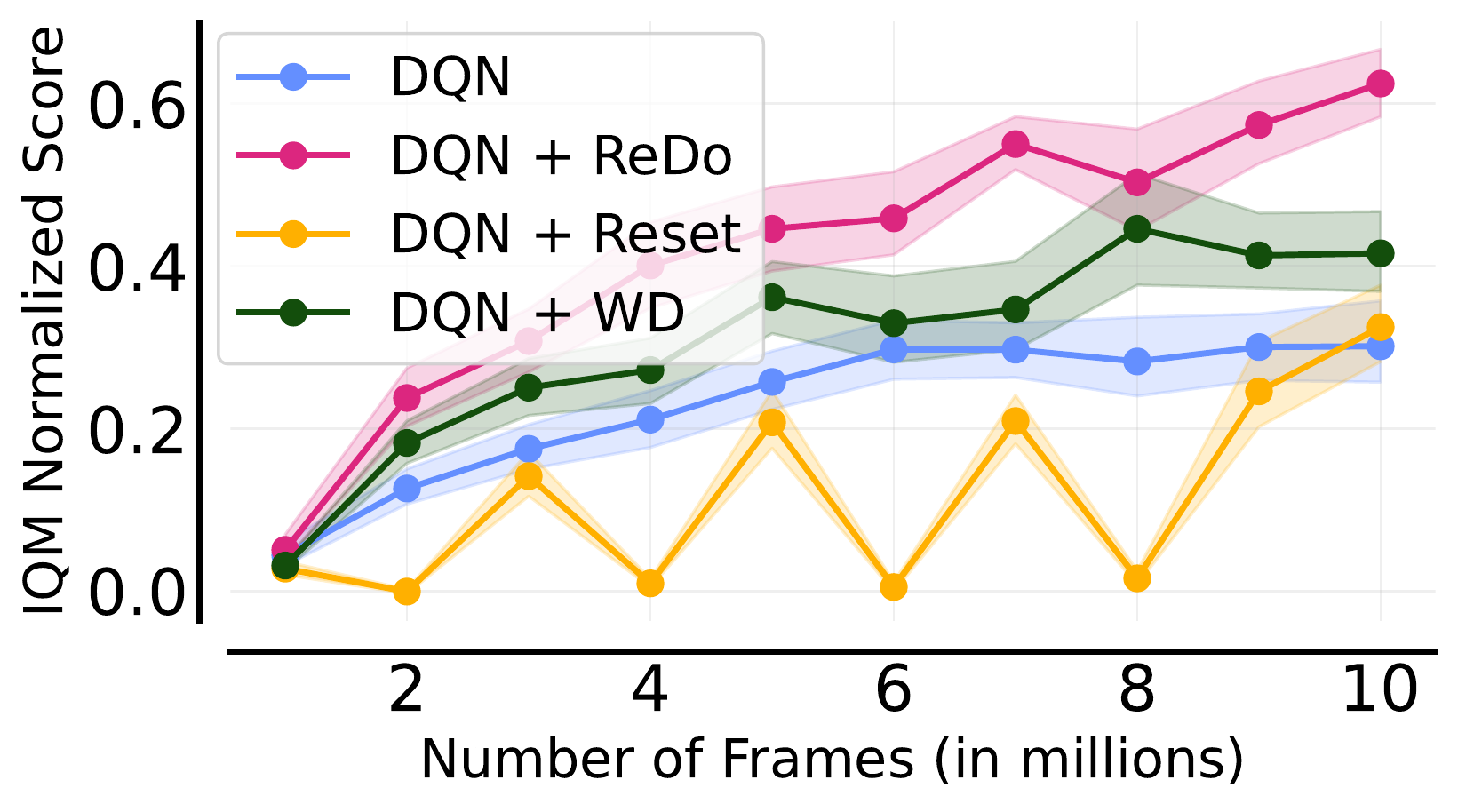}
    \caption{\textbf{Sample efficiency curves}
    for DQN, with a replay ratio of 1, when using
     network resets~\citep{nikishin2022primacy}, weight decay (WD), and our proposed {\em ReDo}.
    Shaded regions show 95\% CIs. The figure shows interquartile mean (IQM) human-normalized scores over the course of training, aggregated across 17 Atari games and 5 runs per game. Among all algorithms, {\em DQN+ReDo} performs the best. }
    \label{fig:main_figure_1}
    \vspace{-0.35cm}
\end{figure}

In this work, we seek to understand the underlying reasons behind the loss of expressivity during the training of RL agents. The observed decrease in the learning ability over time raises the following question: {\em Do RL agents use neural network parameters to their full potential?} To answer this, we analyze neuron activity throughout training and track \emph{dormant} neurons: neurons that have become practically inactive through low activations. 
Our analyses reveal that the number of dormant neurons increases as training progresses, an effect we coin the ``{\em dormant neuron phenomenon}''. 
Specifically, we find that while agents start the training with a small number of dormant neurons, this number increases as training progresses. The effect is exacerbated by the number of gradient updates taken per data collection step. This is in contrast with supervised learning, where the number of dormant neurons remains low throughout training.

We demonstrate the presence of the dormant neuron phenomenon across different algorithms and domains: in two value-based algorithms on the Arcade Learning Environment \citep{bellemare2013arcade} (DQN \cite{mnih2015human} and DrQ($\epsilon$) \cite{yarats2021image,agarwal2021deep}), and with an actor-critic method (SAC \cite{haarnoja2018soft}) evaluated on the MuJoCo suite \cite{todorov2012mujoco}. To address this issue, we propose {\em Re}cycling {\em Do}rmant neurons~({\em ReDo}), a simple and effective method to avoid network under-utilization during training without sacrificing previously learned knowledge: we explicitly limit the spread of dormant neurons by ``recycling'' them to an active state. {\em ReDo} consistently maintains the capacity of the network throughout training and improves the agent's performance (see \autoref{fig:main_figure_1}). Our contributions in this work can be summarized as follows: 
\begin{itemize}
    \item We demonstrate the existence of the dormant neuron phenomenon in deep RL.
    \item We investigate the underlying causes of this phenomenon and show its negative effect on the learning ability of deep RL agents. 
    \item We propose {\em Re}cycling {\em Do}rmant neurons ({\em ReDo}), a simple method to reduce the number of dormant neurons and maintain network expressivity during training.
    \item We demonstrate the effectiveness of {\em ReDo} in maximizing network utilization and improving performance.
\end{itemize}

%% file: 2.background.tex
\section{Background}
We consider a Markov decision process \cite{puterman2014markov}, $\mathcal{M}=\langle\mathcal{S}, \mathcal{A}, \mathcal{R}, \mathcal{P}, \gamma \rangle$, defined by a state space $\mathcal{S}$, an action space $\mathcal{A}$, a reward function $\mathcal{R}:\mathcal{S}\times\mathcal{A}\rightarrow\mathbb{R}$, a transition probability distribution $\mathcal{P}(s'|s, a)$ indicating the probability of transitioning to state $s'$ after taking action $a$ from state $s$, and a discounting factor $\gamma \in [0,1)$. An agent's behaviour is formalized as a policy $\pi:\mathcal{S}\rightarrow Dist(\mathcal{A})$; given any state $s\in\mathcal{S}$ and action $a\in\mathcal{A}$, the value of choosing $a$ from $s$ and following $\pi$ afterwards is given by $Q^\pi(s,a) = \mathbb{E}[\sum_{t=0}^\infty \gamma^t \mathcal{R}(s_t,a_t)] $. The goal in RL is to find a policy $\pi^*$ that maximizes this value: for any $\pi$, $Q^{\pi^*} := Q^* \geq Q^{\pi}$.

In deep reinforcement learning, the $Q$-function is represented using a neural network $Q_\theta$ with parameters $\theta$. During training, an agent interacts with the environment and collects trajectories of the form $(s, a, r, s')\in\mathcal{S}\times\mathcal{A}\times\mathbb{R}\times\mathcal{S}$. These samples are typically stored in a \textit{replay buffer} \cite{lin1992self}, from which batches are sampled to update the parameters of $Q_\theta$ using gradient descent. The optimization performed aims to minimize the temporal difference loss \cite{sutton1988learning}: \(\mathcal{L} = Q_\theta(s, a) - Q^\mathcal{T}_\theta(s, a)\); here, $Q^\mathcal{T}_\theta(s, a)$ is the bootstrap target \([\mathcal{R}(s, a) + \gamma \max_{a'\in\mathcal{A}}Q_{\tilde{\theta}}(s', a')]\) and $Q_{\tilde{\theta}}$ is a delayed version of $Q_{\theta}$ that is known as the \textit{target} network.

The number of gradient updates performed per environment step is known as the \textit{replay ratio}. This is a key design choice that has a substantial impact on performance \cite{van2019use,fedus2020revisiting,kumar2021dr3, nikishin2022primacy}. Increasing the replay ratio can increase the sample-efficiency of RL agents as more parameter updates per sampled trajectory are performed. However, prior works have shown that training agents with a high replay ratio can cause training instabilities, ultimately resulting in decreased agent performance \citep{nikishin2022primacy}.

One important aspect of reinforcement learning, when contrasted with supervised learning, is that RL agents train on highly non-stationary data, where the non-stationarity is coming in a few forms \cite{igl2020transient}, but we focus on two of the most salient ones.\\
{\bf Input data non-stationarity:} The data the agent trains on is collected in an online manner by interacting with the environment using its current policy $\pi$; this data is then used to update the policy, which affects the distribution of future samples.\\
{\bf Target non-stationarity:} The learning target used by RL agents is based on its own estimate $Q_{\tilde{\theta}}$, which is changing as learning progresses.
 

%% file: 3.analysis.tex
\section{The Dormant Neuron Phenomenon}
\label{sec:analysis}

Prior work has highlighted the fact that networks used in online RL tend to {\em lose} their expressive ability; in this section we demonstrate that {\em dormant neurons} play an important role in this finding.

\begin{definition}
Given an input distribution $D$, let $h^\ell_i(x)$ denote the activation of neuron $i$ in layer $\ell$ under input $x\in D$ and $H^\ell$ be the number of neurons in layer $\ell$. We define the score of a neuron $i$ (in layer $\ell$) via the normalized average of its activation as follows:
\begin{align}
    \label{eqn:neuronScore}
    s^\ell_i = \frac{\mathbb{E}_{x\in D}|h^\ell_i(x)|}{\frac{1}{H^\ell}\sum_{k\in h}\mathbb{E}_{x\in D} |h^\ell_k(x)|}
\end{align}
We say a neuron $i$ in layer $\ell$ is {\bf $\tau$-dormant} if $s^\ell_i \leq \tau$.
\end{definition}

We normalize the scores such that they sum to 1 within a layer. This makes the comparison of neurons in different layers possible. The threshold $\tau$ allows us to detect neurons with low activations. Even though these low activation neurons could, in theory, impact the learned functions when recycled, their impact is expected to be less than the neurons with high activations. 

\begin{figure}[t]
\vskip 0.2in
\begin{center}
{
\includegraphics[width=0.49\columnwidth]{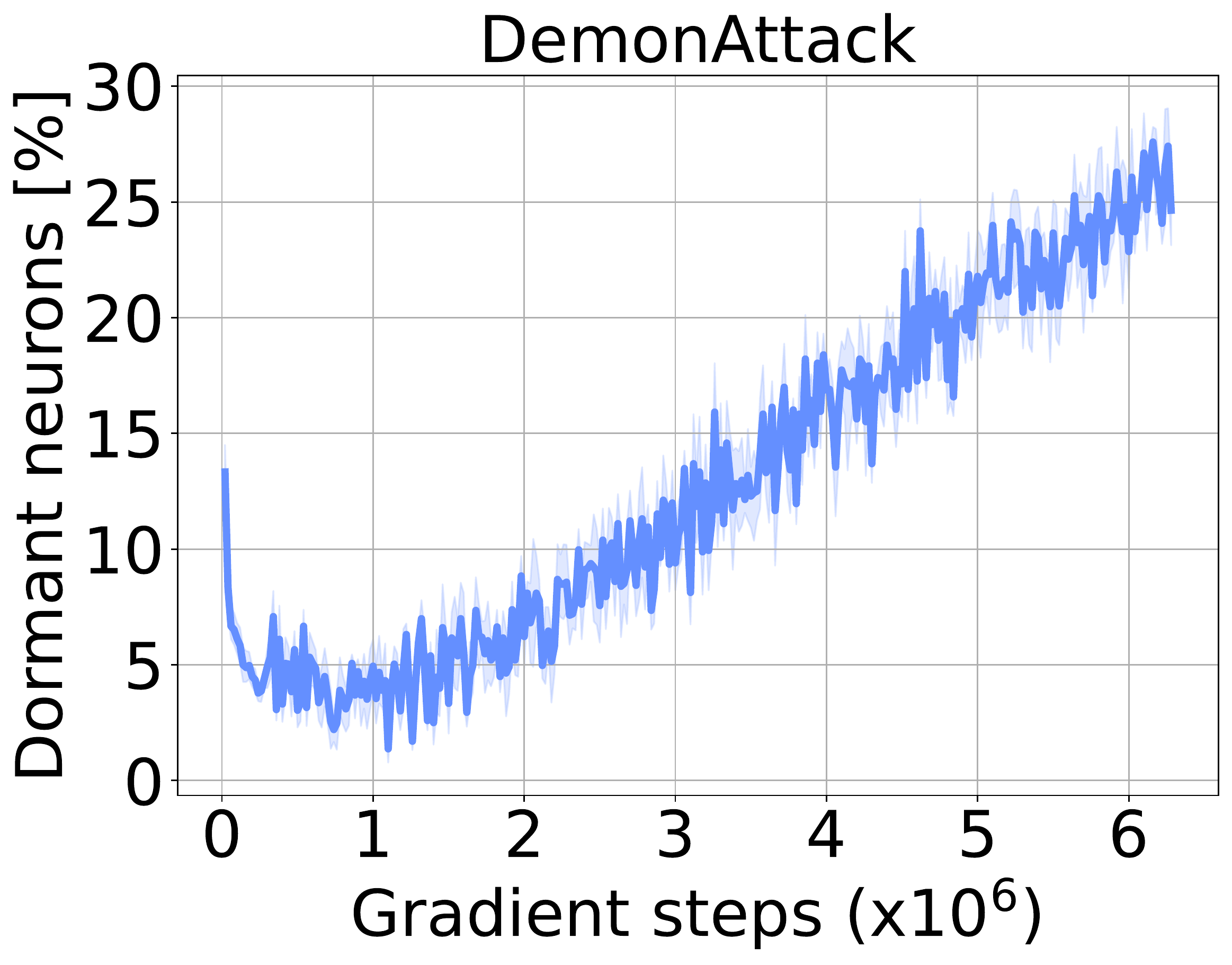}\hfill
\includegraphics[width=0.49\columnwidth]{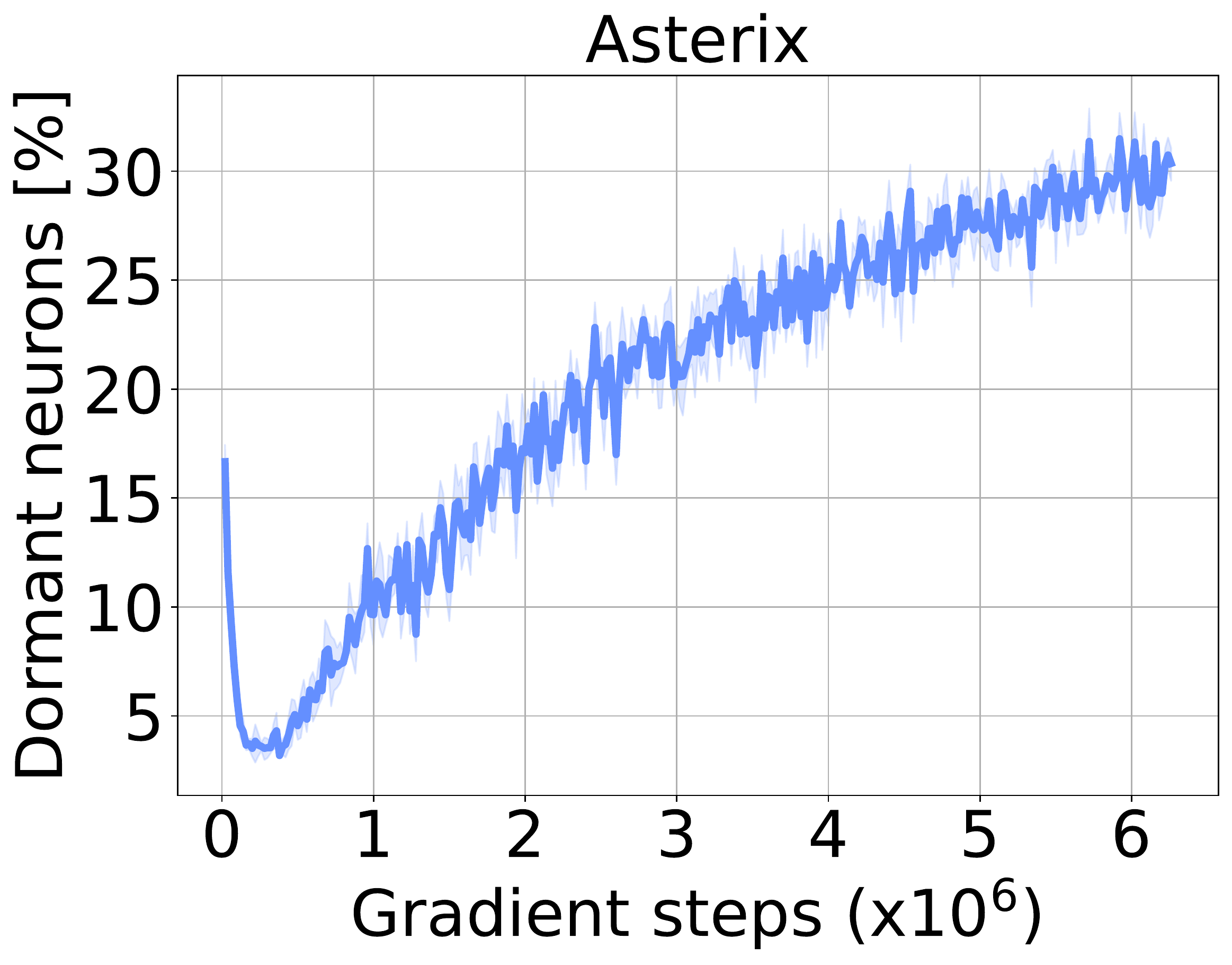}
}
\caption{The percentage of dormant neurons increases throughout training for DQN agents.}
\label{fig:dormant_increase}
\end{center}
\vskip -0.2in
\end{figure}

\begin{definition}
An algorithm exhibits the {\bf dormant neuron phenomenon} if the number of $\tau$-dormant neurons in its neural network increases steadily throughout training.
\end{definition}

An algorithm exhibiting the dormant neuron phenomenon is not using its network's capacity to its full potential, and this under-utilization worsens over time.

The remainder of this section focuses first on demonstrating that RL agents suffer from the dormant neuron phenomenon, and then on understanding the underlying causes for it. Specifically, we analyze DQN \citep{mnih2015human}, a foundational agent on which most modern value-based agents are based. To do so, we run our evaluations on the Arcade Learning Environment \citep{bellemare2013arcade} using 5 independent seeds for each experiment, and reporting 95\% confidence intervals. For clarity, we focus our analyses on two representative games (DemonAttack and Asterix), but include others in the appendix. In these initial analyses we focus solely on $\tau=0$ dormancy, but loosen this threshold when benchmarking our algorithm in sections~\ref{sec:method} and \ref{sec:experiments}. Additionally, we present analyses on an actor-critic method (SAC \cite{haarnoja2018soft}) and a modern sample-efficient agent (DrQ($\epsilon$)~\cite{yarats2021image}) in Appendix \ref{appendix:dormant_phenomenon_more_domains}.

\paragraph{The dormant neuron phenomenon is present in deep RL agents.} 
We begin our analyses by tracking the number of dormant neurons during DQN training. In \autoref{fig:dormant_increase}, we observe that the percentage of dormant neurons steadily increases throughout training. This observation is consistent across different algorithms and environments, as can be seen in Appendix~\ref{appendix:dormant_phenomenon_more_domains}.

\begin{figure}[t]
\vskip 0.2in
\begin{center}
\centerline{
\includegraphics[width=0.6\columnwidth]{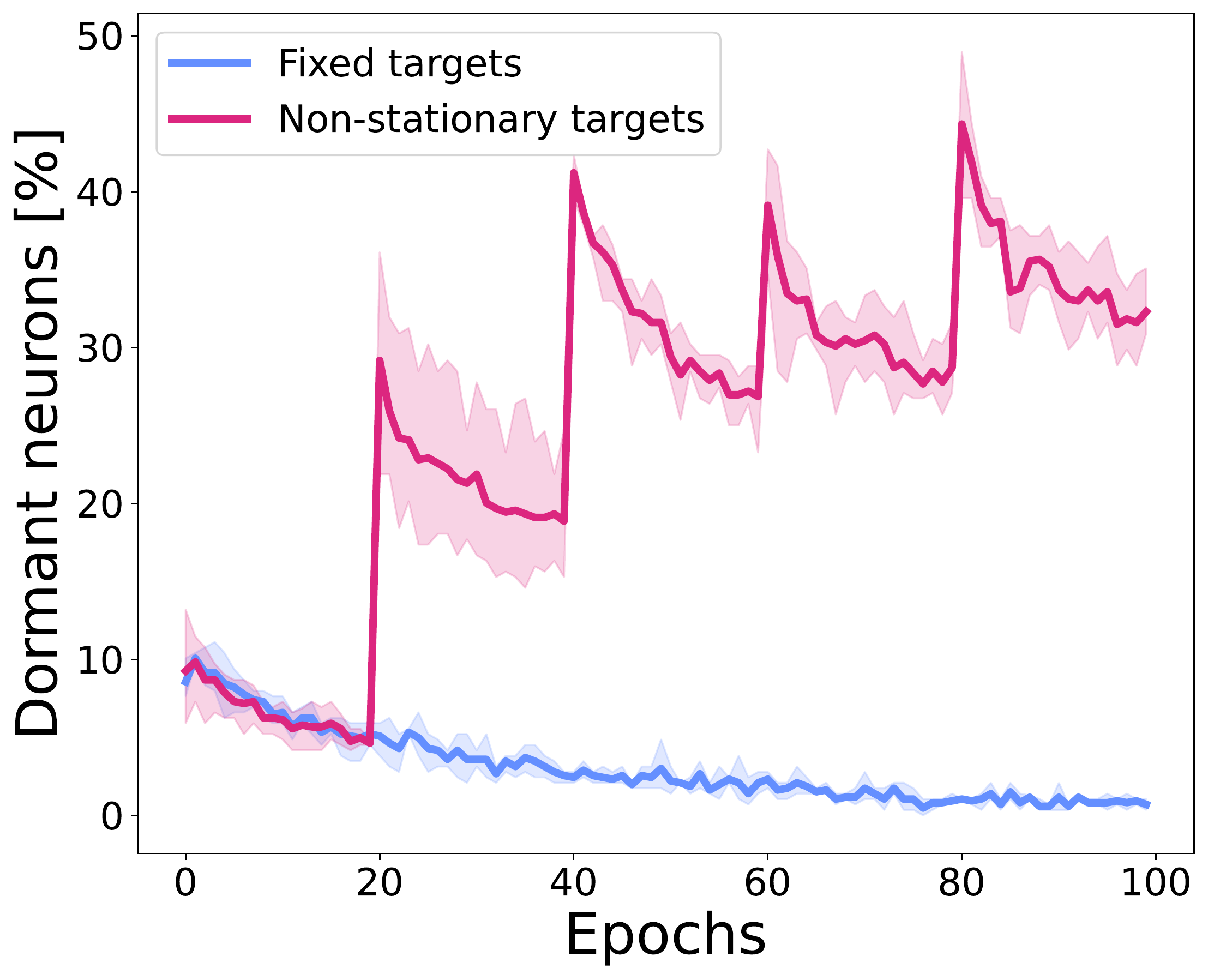}
}
\caption{Percentage of dormant neurons when training on CIFAR-10 with fixed and non-stationary targets. Averaged over 3 independent seeds with shaded areas reporting 95\% confidence intervals. The percentage of dormant neurons increases with non-stationary targets.}
\label{fig:non_stationary_targets}
\end{center}
\vskip -0.2in
\end{figure}

\begin{figure}[t]
\vskip 0.2in
\begin{center}
\centerline{
\includegraphics[width=0.49\columnwidth]{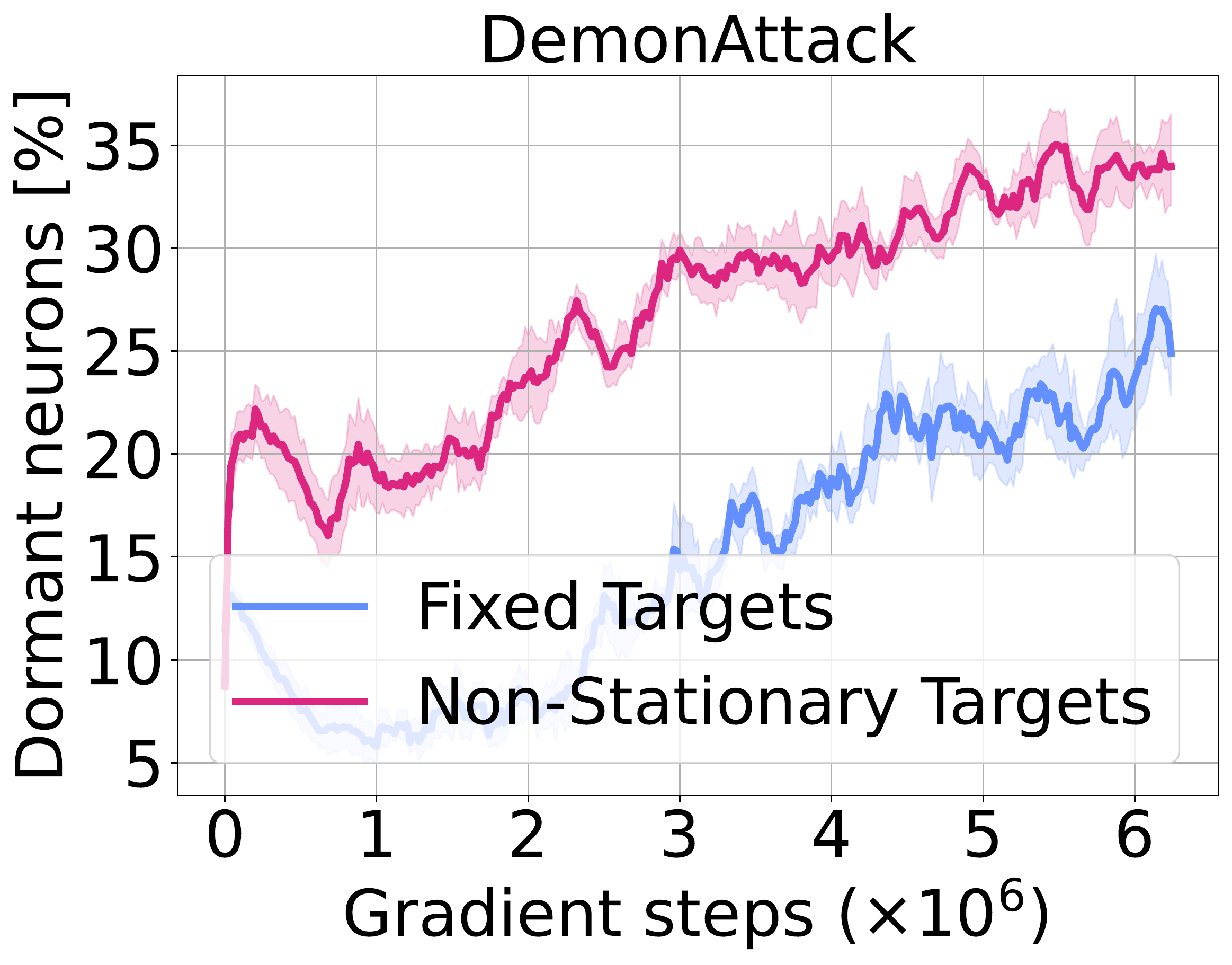}\hfill
\includegraphics[width=0.49\columnwidth]{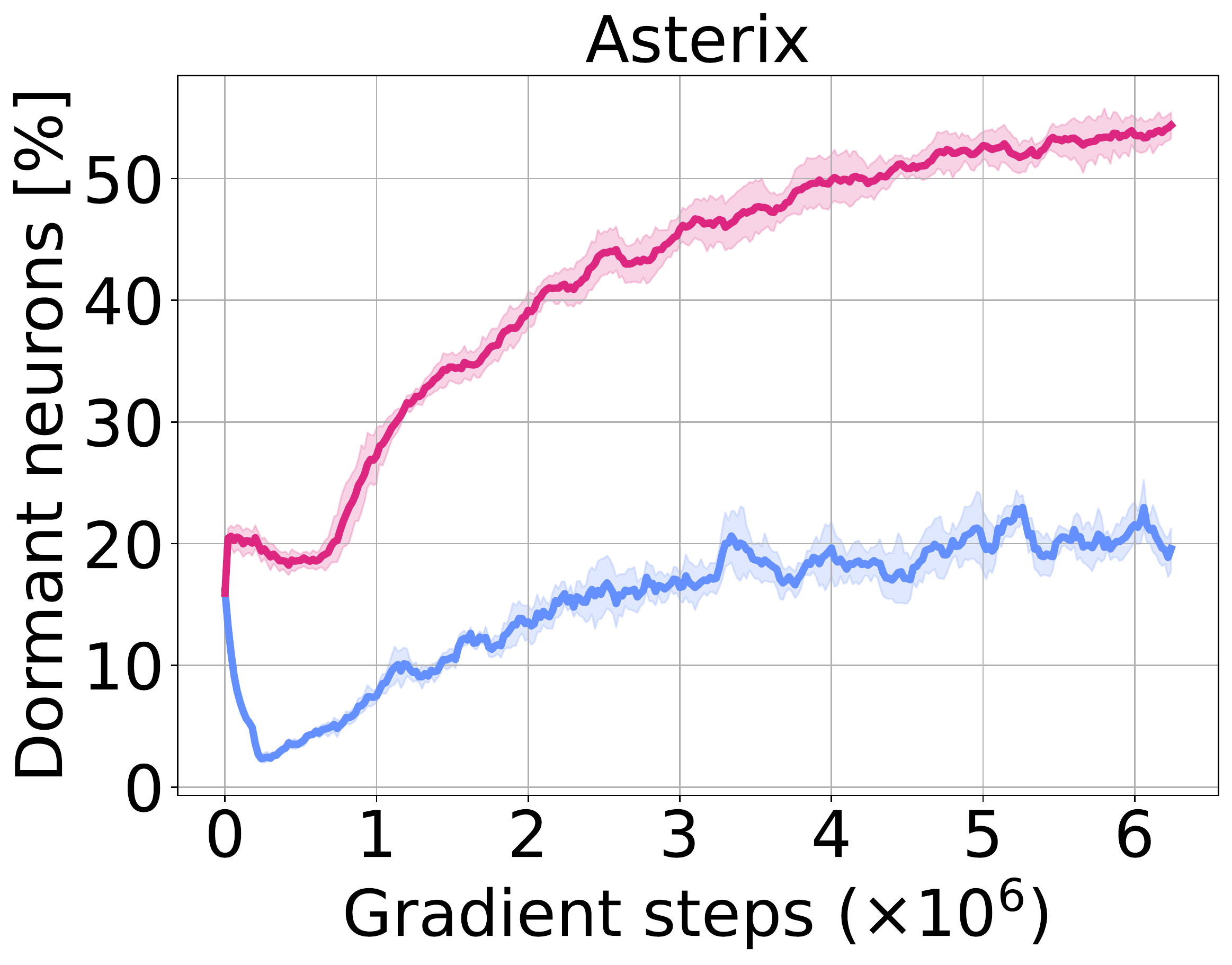}
}
\caption{{\em Offline RL.} Dormant neurons throughout training with standard moving targets and fixed (random) targets. The phenomenon is still present in offline RL, where the training data is fixed.}
\label{fig:dormant_offline}
\end{center}
\vskip -0.2in
\end{figure}

\paragraph{Target non-stationarity exacerbates dormant neurons.} We hypothesize that the non-stationarity of training deep RL agents is one of the causes for the dormant neuron phenomenon. To evaluate this hypothesis, we consider two supervised learning scenarios using the standard CIFAR-10 dataset \cite{krizhevsky2009learning}: (1) training a network with \textit{fixed targets}, and (2) training a network with \textit{non-stationary targets}, where the labels are shuffled throughout training (see Appendix~\ref{appendix:experimental_details} for details). As \autoref{fig:non_stationary_targets} shows, the number of dormant neurons {\em decreases} over time with fixed targets, but {\em increases} over time with non-stationary targets. Indeed, the sharp increases in the figure correspond to the points in training when the labels are shuffled. These findings suggest that the continuously changing targets in deep RL are a significant factor for the presence of the phenomenon.

\paragraph{Input non-stationarity does not appear to be a major factor.}
To investigate whether the non-stationarity due to online data collection plays a role in exacerbating the phenomenon, we measure the number of dormant neurons in the \textit{offline} RL setting, where an agent is trained on a fixed dataset (we used the dataset provided by~\citet{agarwal2020optimistic}). In \autoref{fig:dormant_offline} we can see that the phenomenon remains in this setting, suggesting that input non-stationary is not one of the primary contributing factors. To further analyze the source of dormant neurons in this setting, we train RL agents with {\em fixed} random targets (ablating the non-stationarity in inputs and targets). The decrease in the number of dormant neurons observed in this case (\autoref{fig:dormant_offline}) supports our hypothesis that target non-stationarity in RL training is the primary source of the dormant neuron phenomenon.

\begin{figure}[t]
\vskip 0.2in
\begin{center}
\centerline{
\includegraphics[width=0.49\columnwidth]{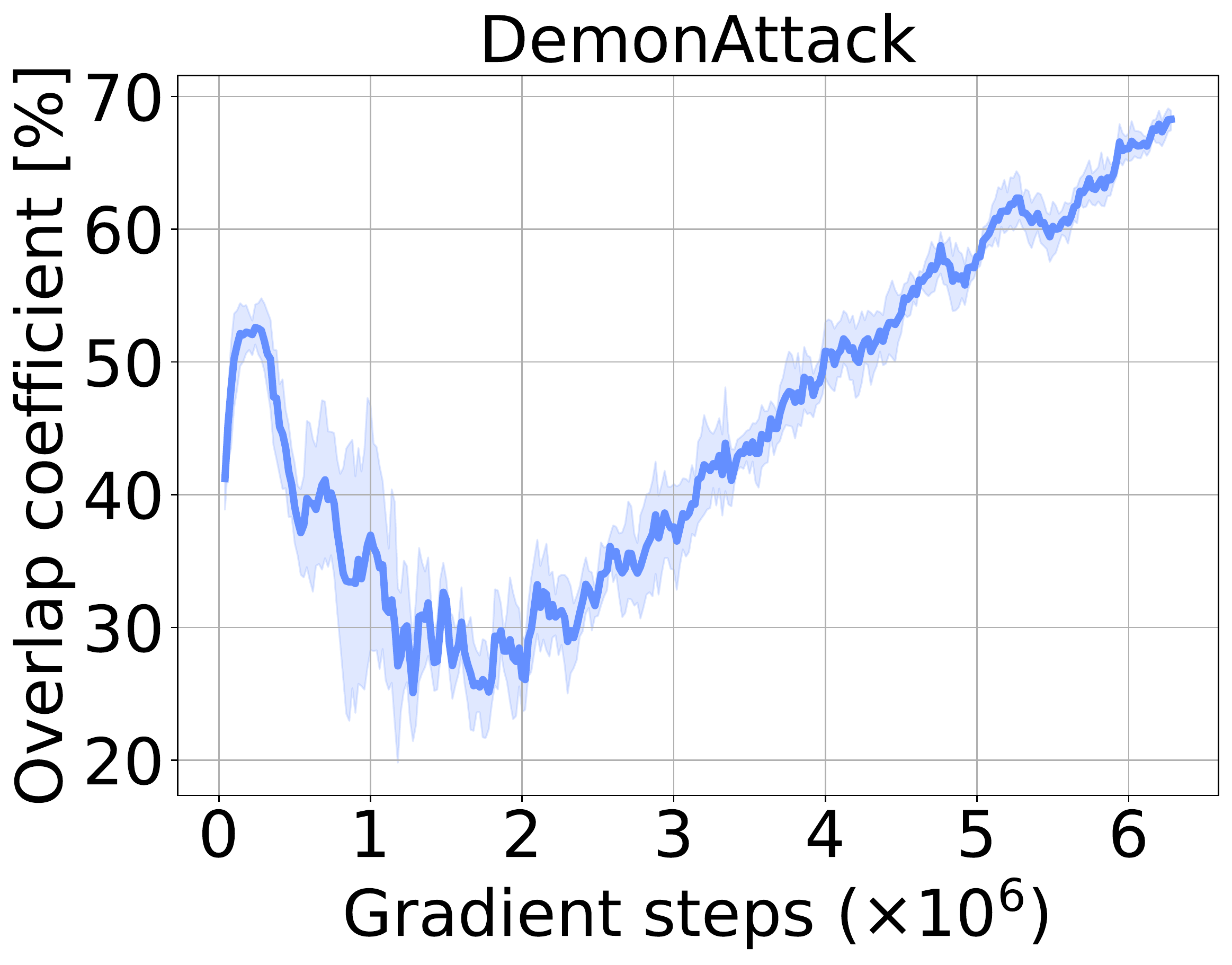}
\hspace{0.2cm}
\includegraphics[width=0.49\columnwidth]{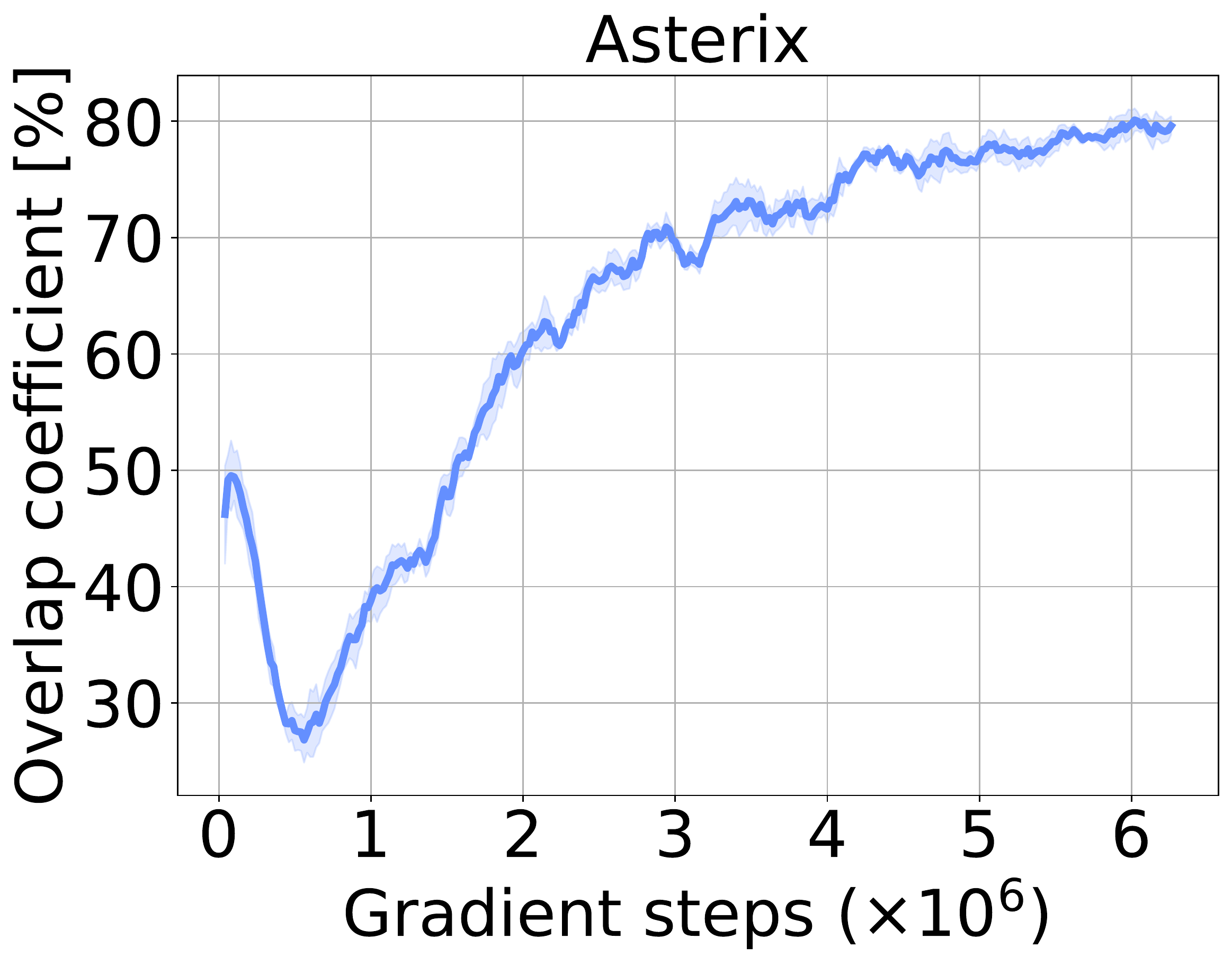}
}
\caption{The overlap coefficient of dormant neurons throughout training. There is an increase in the number of dormant neurons that remain dormant.}
\label{fig:intersected_deadneurons_percent}
\end{center}
\vskip -0.2in
\end{figure}

\begin{figure}[t]
\vskip 0.2in
\begin{center}
{
\includegraphics[width=0.49\columnwidth]{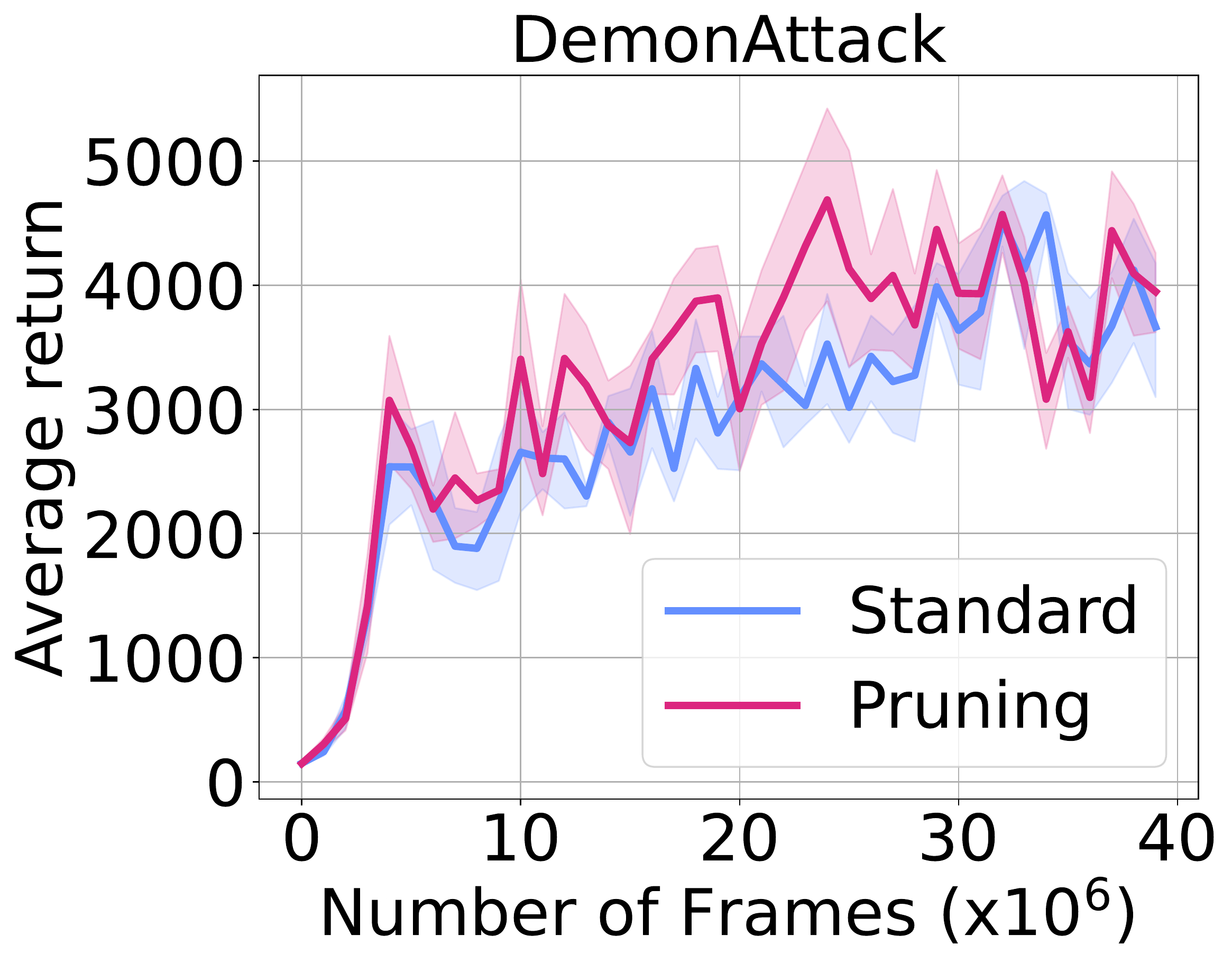}\hfill
\includegraphics[width=0.49\columnwidth]{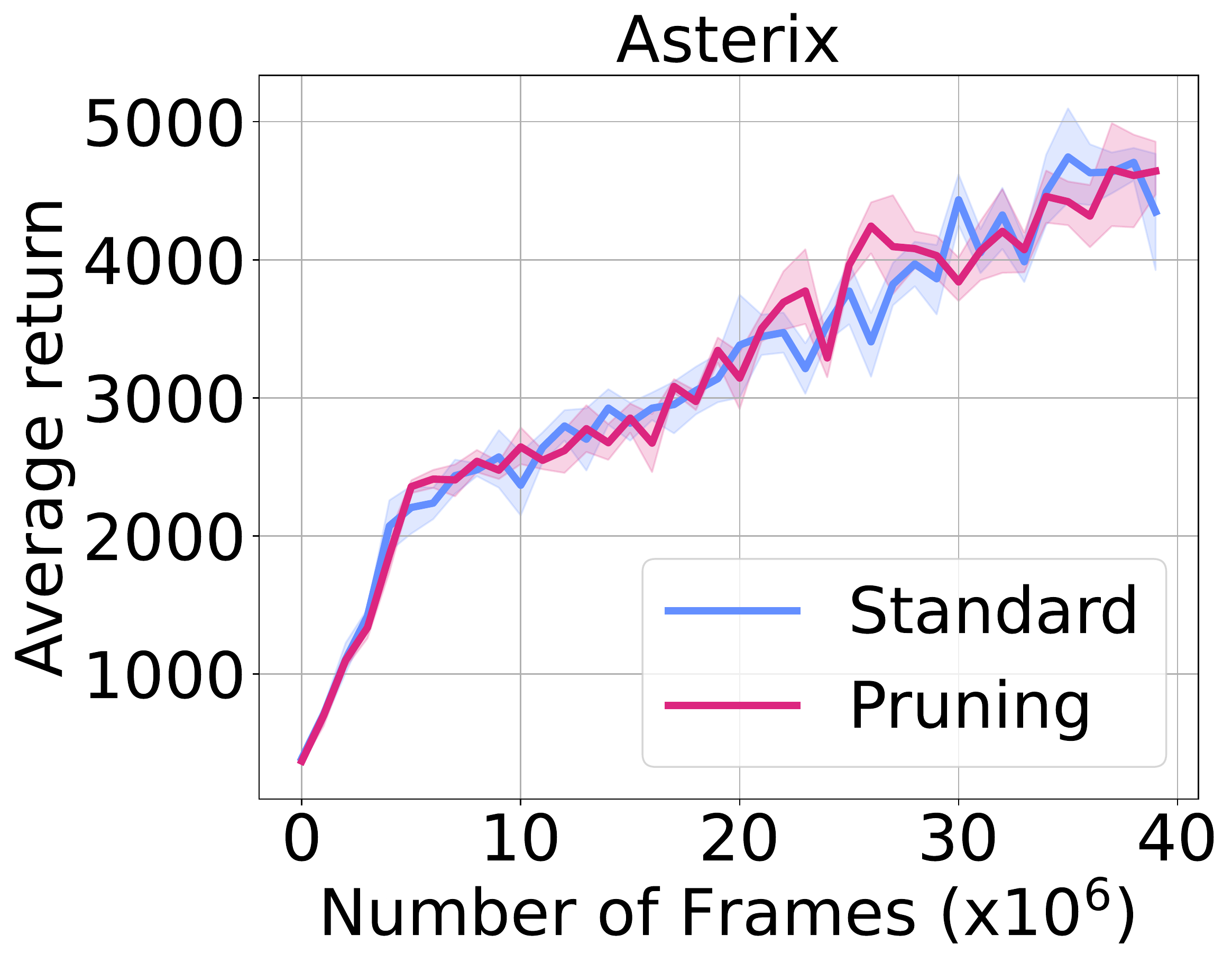}
}
\caption{Pruning dormant neurons during training does not affect the performance of an agent.}
\label{fig:pruning}
\end{center}
\vskip -0.2in
\end{figure}

\paragraph{Dormant neurons remain dormant.}
To investigate whether dormant neurons ``reactivate'' as training progresses, we track the overlap in the set of dormant neurons. \autoref{fig:intersected_deadneurons_percent} plots the overlap coefficient between the set of dormant neurons in the penultimate layer at the current iteration, and the historical set of dormant neurons.\footnote{The overlap coefficient between two sets $X$ and $Y$ is defined as $overlap(X, Y) = \frac{|X\cap Y|}{\min(|X|, |Y|)}$.} The increase shown in the figure strongly suggests that once a neuron becomes dormant, it remains that way for the rest of training. 
To further investigate this, we explicitly {\em prune} any neuron found dormant throughout training, to check whether their removal affects the agent's overall performance. As \autoref{fig:pruning} shows, their removal does {\em not} affect the agent's performance, further confirming that dormant neurons remain dormant.

\begin{figure}[t]
\vskip 0.2in
\begin{center}
\begin{subfigure}
{
\includegraphics[width=0.49\columnwidth]{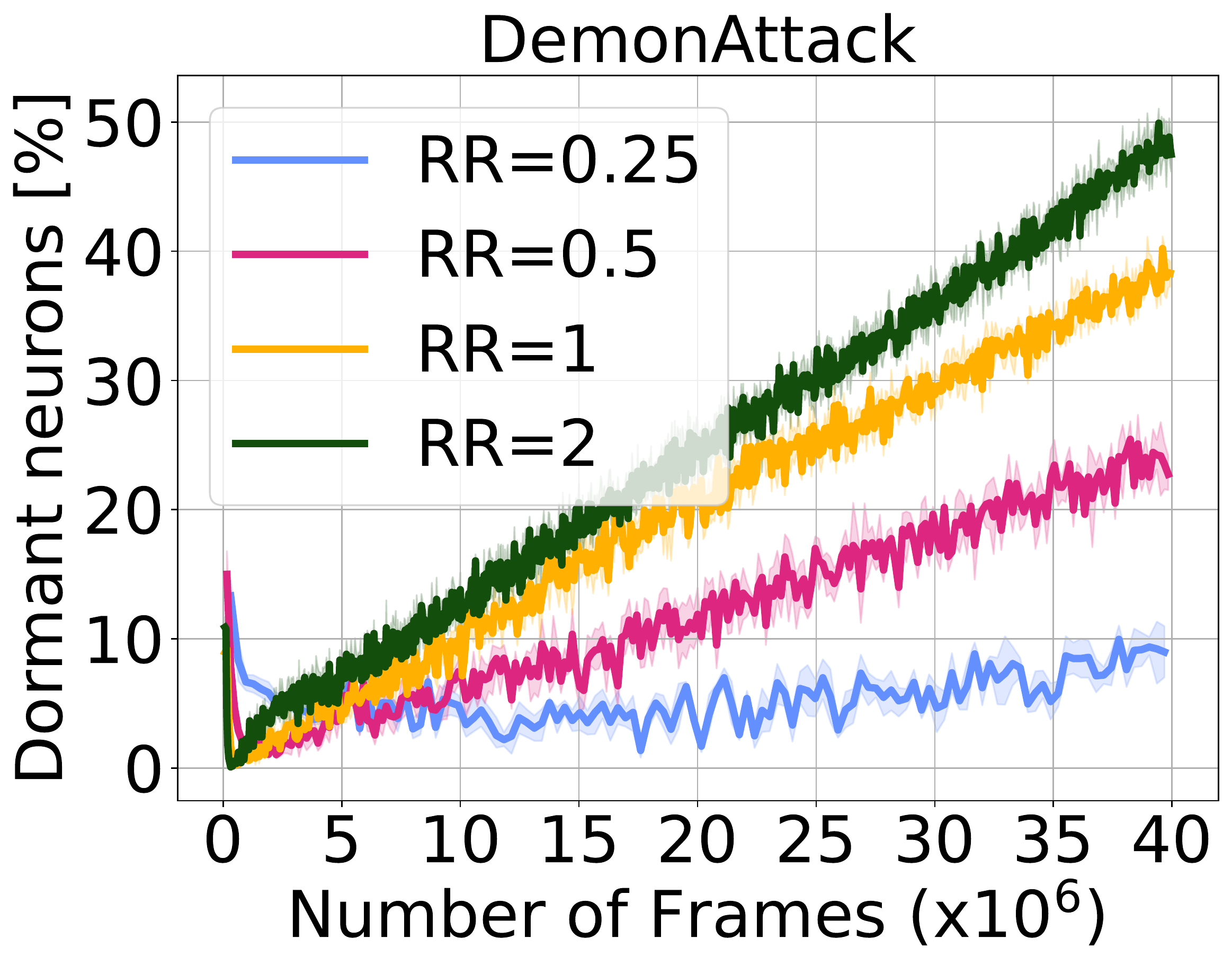}
\hspace{0.2cm}
\includegraphics[width=0.49\columnwidth]{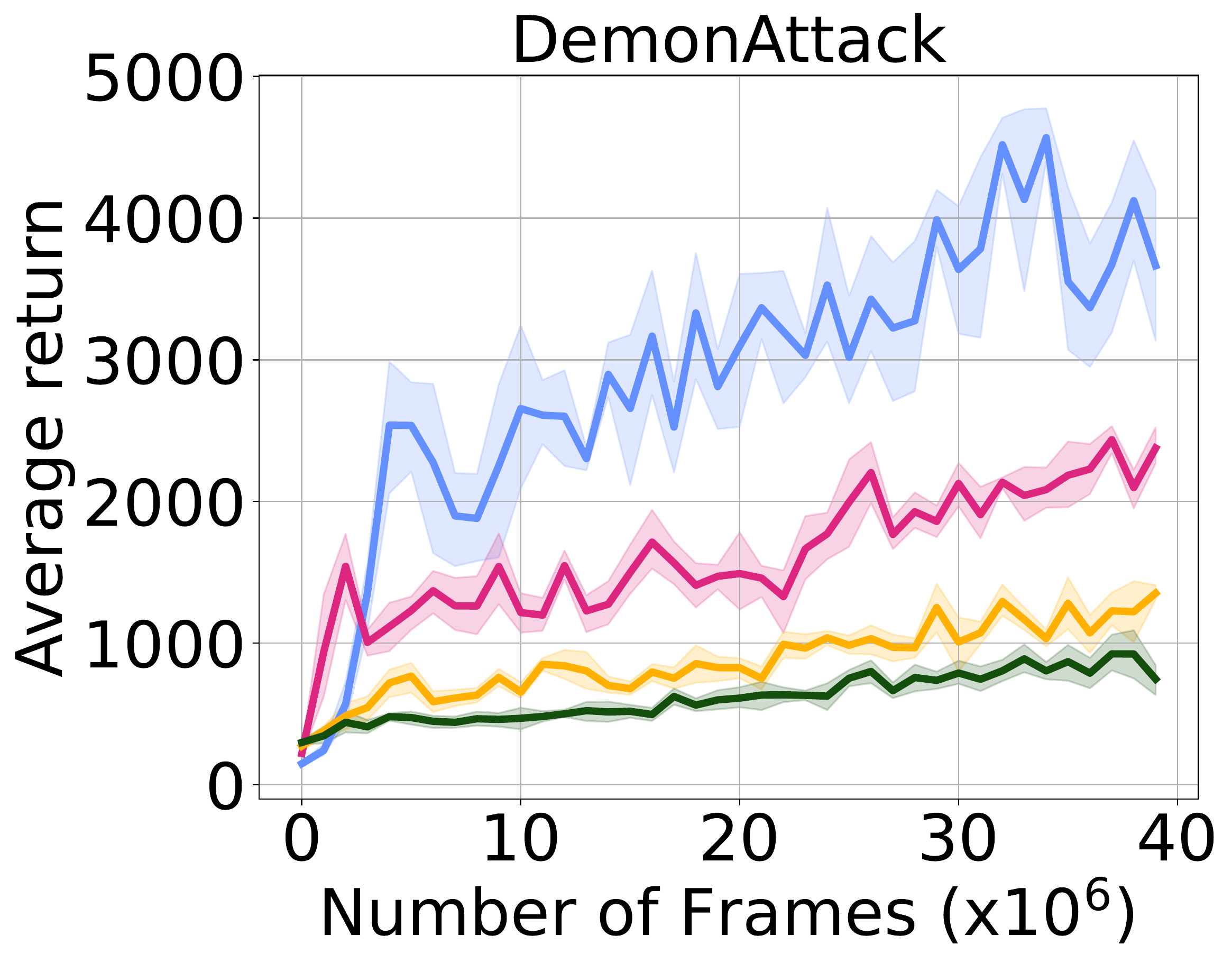}
}
\end{subfigure}
\caption{The rate of increase in dormant neurons with varying replay ratio (RR) (left). As the replay ratio increases, the number of dormant neurons also increases. The higher percentage of dormant neurons correlates with the performance drop that occurs when the replay ratio is increased (right).}
\label{fig:dormant_in_high_RR}
\end{center}
\vskip -0.2in
\end{figure}

\paragraph{More gradient updates leads to more dormant neurons.}
Although an increase in replay ratio can seem appealing from a data-efficiency point of view (as more gradient updates per environment step are taken), it has been shown to cause overfitting and performance collapse \cite{kumar2020implicit,nikishin2022primacy}. In \autoref{fig:dormant_in_high_RR} we measure neuron dormancy while varying the replay ratio, and observe a strong correlation between replay ratio and the fraction of neurons turning dormant. Although difficult to assert conclusively, this finding could account for the difficulty in training RL agents with higher replay ratios; indeed, we will demonstrate in Section \ref{sec:experiments} that recycling dormant neurons and activating them can mitigate this instability, leading to better results.

\begin{figure}[t]
\vskip 0.2in
\begin{center}
\centerline{
\includegraphics[width=0.48\columnwidth]{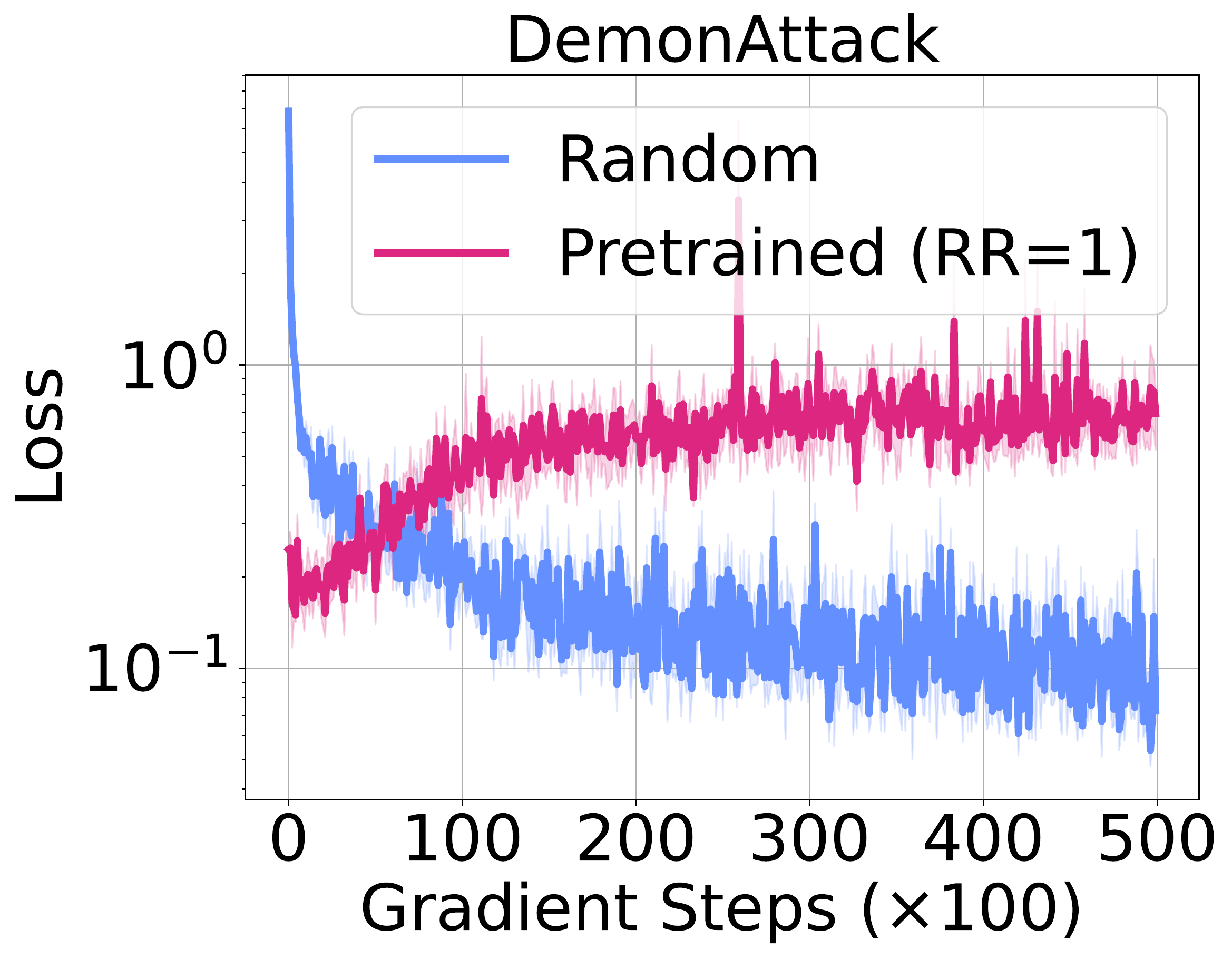}\hfill
\includegraphics[width=0.48\columnwidth]{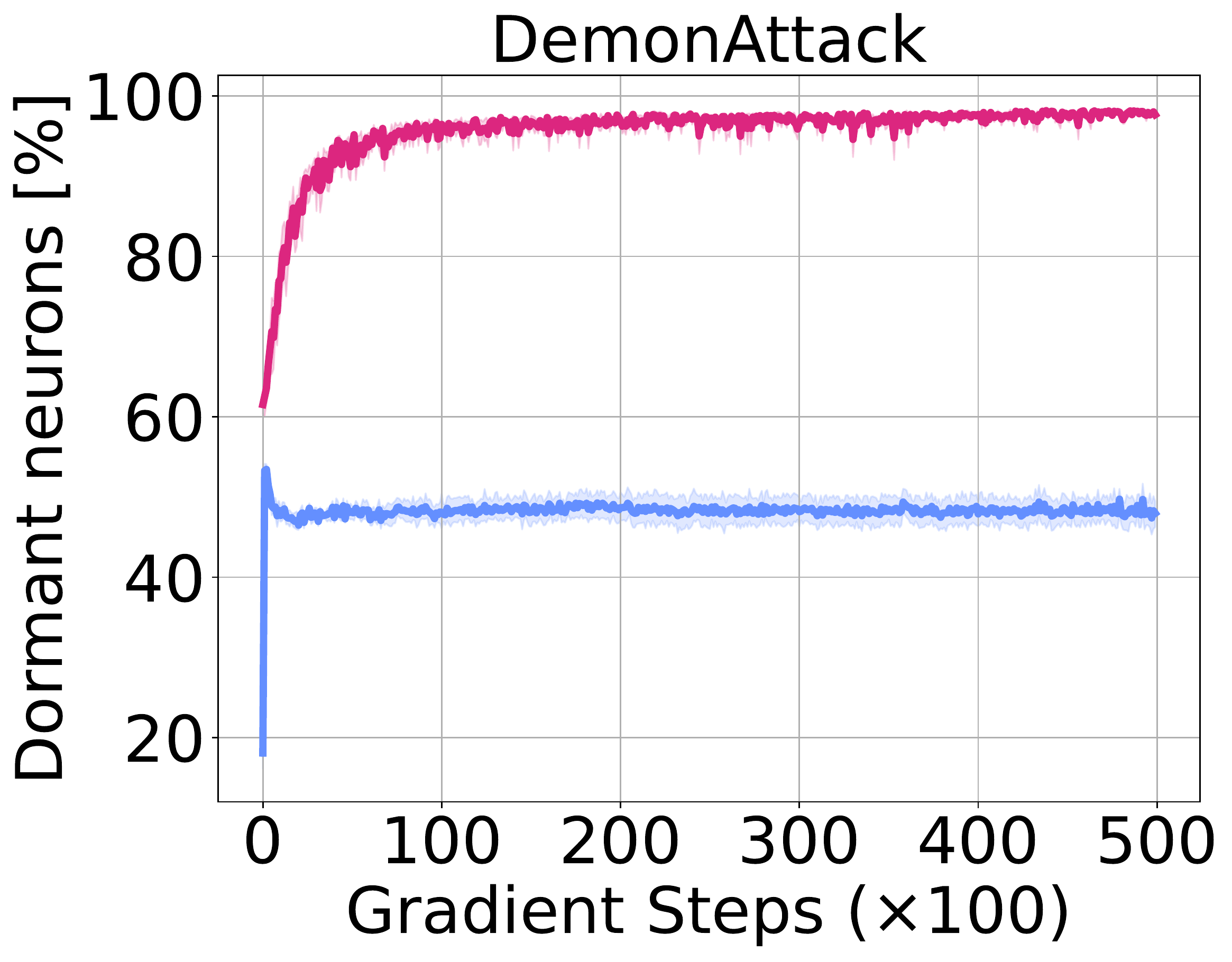}
}
\caption{A pretrained network that exhibits dormant neurons has less ability than a randomly initialized network to fit a fixed target. Results are averaged over 5 seeds.}
\label{fig:regression}
\end{center}
\vskip -0.2in
\end{figure}

\paragraph{Dormant neurons make learning new tasks more difficult.}
We directly examine the effect of dormant neurons on an RL network's ability to learn new tasks. To do so, we train a DQN agent with a replay ratio of 1 (this agent exhibits a high level of dormant neurons as observed in \autoref{fig:dormant_in_high_RR}). Next we fine-tune this network by distilling it towards a well performing DQN agent's network, using a traditional regression loss and compare this with a randomly initialized agent trained using the same loss. In \autoref{fig:regression} we see that the pre-trained network, which starts with a high level of dormant neurons, shows degrading performance throughout training; in contrast, the randomly initialized baseline is able to continuously improve. Further, while the baseline network maintains a stable level of dormant neurons, the number of dormant neurons in the pre-trained network continues to increase throughout training.

%% file: 4.method.tex
\section{\textit{Re}cycling \textit{Do}rmant Neurons ({\em ReDo})}
\label{sec:method}

Our analyses in Section \ref{sec:analysis}, which demonstrates the existence of the dormant neuron phenomenon in online RL, suggests these dormant neurons may have a role to play in the diminished expressivity highlighted by \citet{kumar2020implicit} and \citet{lyle2021understanding}. To account for this, we propose to {\bf re}cycle {\bf do}rmant neurons periodically during training ({\em ReDo}).

The main idea of {\em ReDo}, outlined in Algorithm~\ref{alg:redo}, is rather simple: during regular training, periodically check in all layers whether any neurons are $\tau$-dormant; for these, reinitialize their incoming weights and zero out the outgoing weights. The incoming weights are initialized using the original weight distribution. Note that if $\tau$ is $0$, we are effectively leaving the network's output unchanged; if $\tau$ is small, the output of the network is only slightly changed.

\begin{figure}[!t]
\vskip 0.2in
\begin{center}
{
\includegraphics[width=0.48\columnwidth]{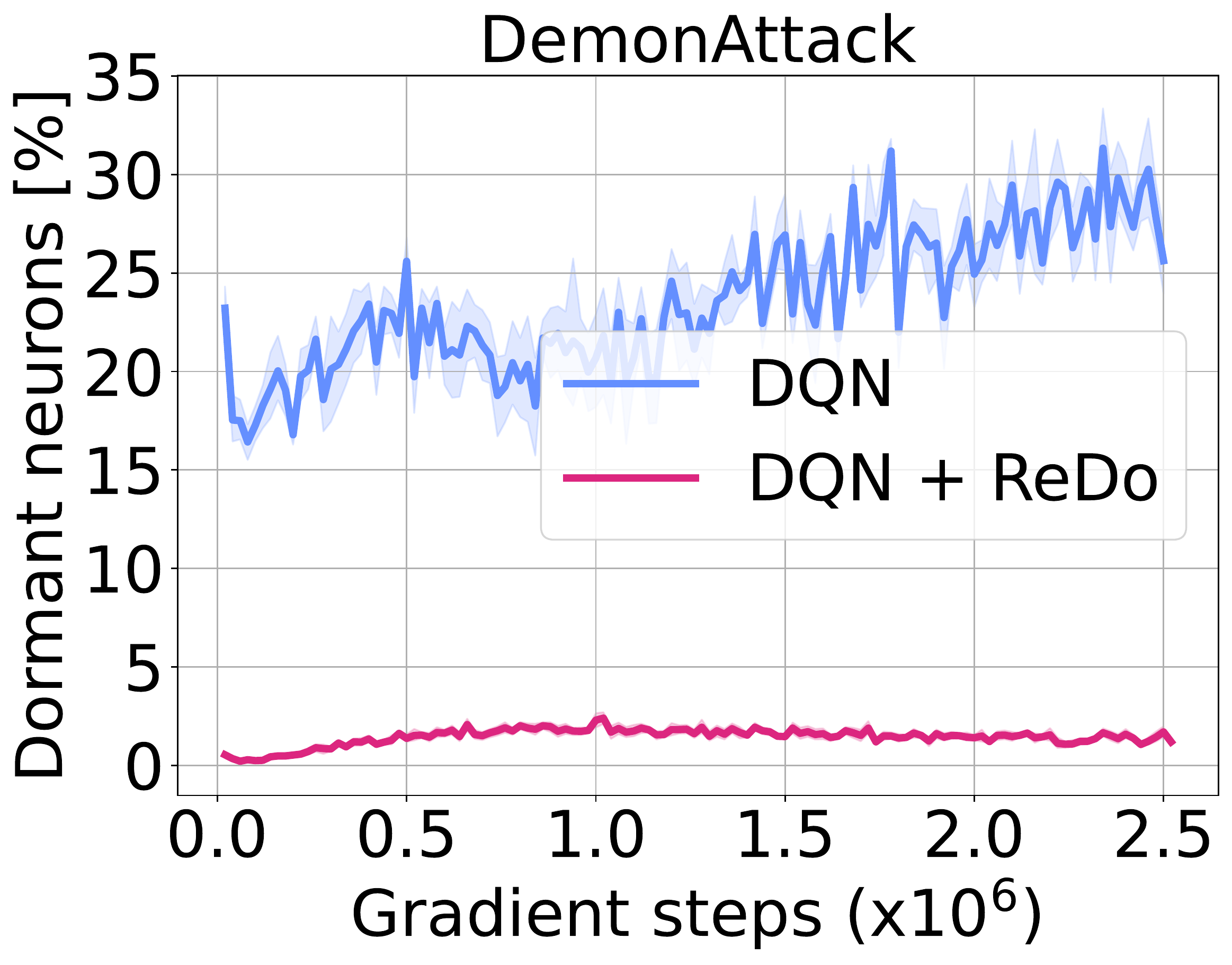}\hfill
\includegraphics[width=0.48\columnwidth]{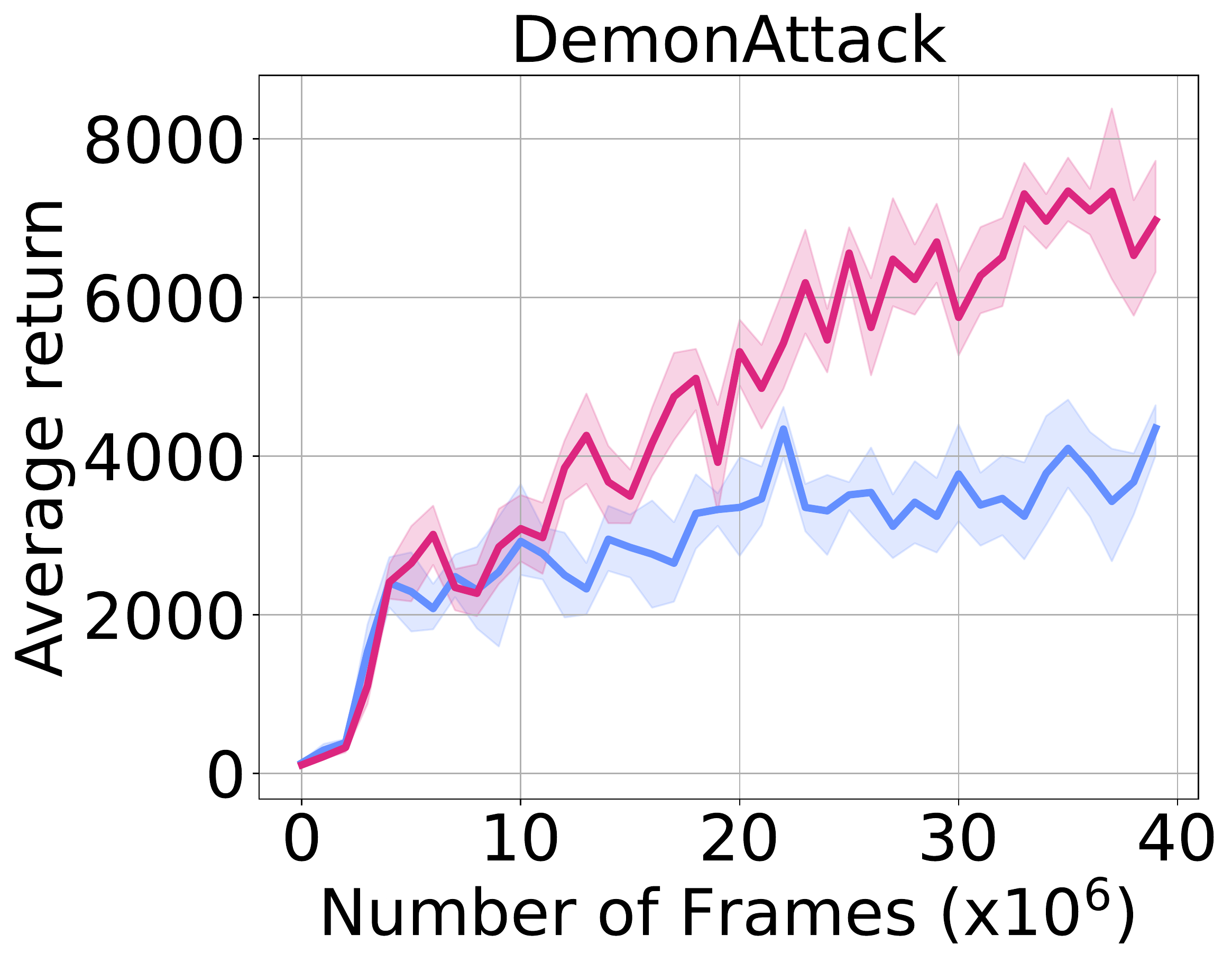} 
}
\caption{Evaluation of {\em ReDo}'s effectiveness (with $\tau=0.025$) in reducing dormant neurons (left) and improving performance (right) on DQN (with $RR=0.25$).}
\label{fig:default_setting}
\end{center}
\vskip -0.2in
\end{figure}

Figure \ref{fig:default_setting} showcases the effectiveness of {\em ReDo} in dramatically reducing the number of dormant neurons, which also results in improved agent performance. Before diving into a deeper empirical evaluation of our method in Section \ref{sec:experiments}, we discuss some algorithmic alternatives we considered when designing {\em ReDo}.

\paragraph{Alternate recycling strategies.} We considered other recycling strategies, such as scaling the incoming connections using the mean of the norm of non-dormant neurons. However, this strategy performed similarly to using initial weight distribution. Similarly, alternative initialization strategies like initializing outgoing connections randomly resulted in similar or worse returns. Results of these investigations are shared in Appendix \ref{app:recycling_strategies}.  

\paragraph{Are ReLUs to blame?} RL networks typically use ReLU activations, which can saturate at zero outputs, and hence zero gradients. 
To investigate whether the issue is specific to the use of ReLUs, in Appendix \ref{appendix:effect_activation_fn} we measured the number of dormant neurons and resulting performance when using a different activation function. We observed that there is a mild decrease in the number of dormant neurons, but the phenomenon is still present.

\begin{algorithm}[t]
   \caption{{\em ReDo}}
   \label{alg:redo}
\begin{algorithmic}
   \STATE {\bfseries Input:} Network parameters $\theta$, threshold $\tau$, training steps $T$, frequency $F$ 
   \FOR{$t=1$ to {\bfseries to} $T$}
   \STATE Update $\theta$ with regular RL loss
   \IF{$t \mod F == 0$}
   \FOR{each neuron $i$}
   \IF{$s^\ell_i \leq \tau$}
   \STATE Reinitialize input weights of neuron $i$
   \STATE Set outgoing weights of neuron $i$ to $0$
   \ENDIF
   \ENDFOR
   \ENDIF
   \ENDFOR
\end{algorithmic}
\end{algorithm}

%% file: 5.results.tex
\section{Empirical Evaluations}
\label{sec:experiments}
\begin{figure*}[ht]
\vskip 0.2in
\begin{center}
{
\includegraphics[width=0.49\columnwidth]{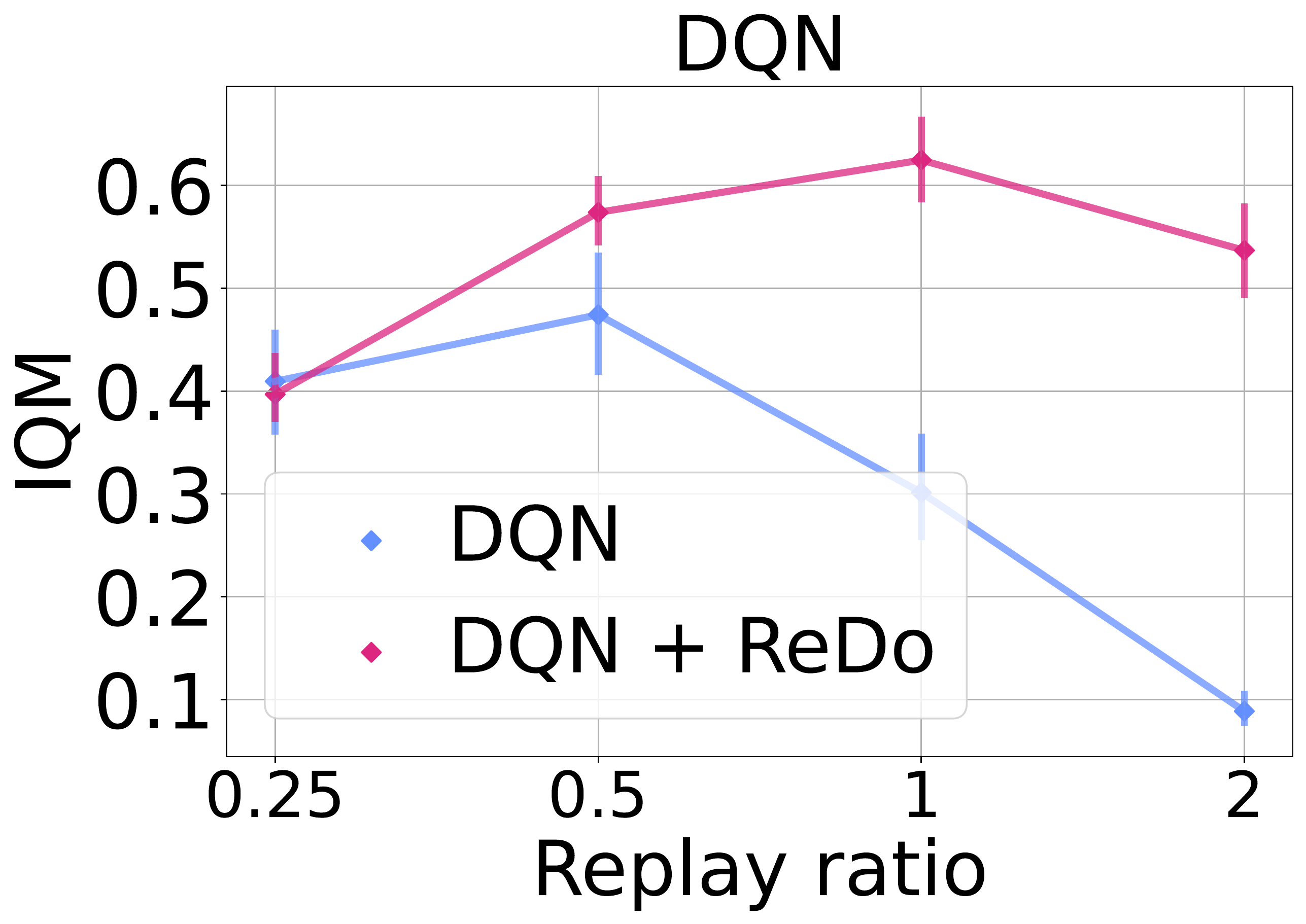}
\includegraphics[width=0.49\columnwidth]{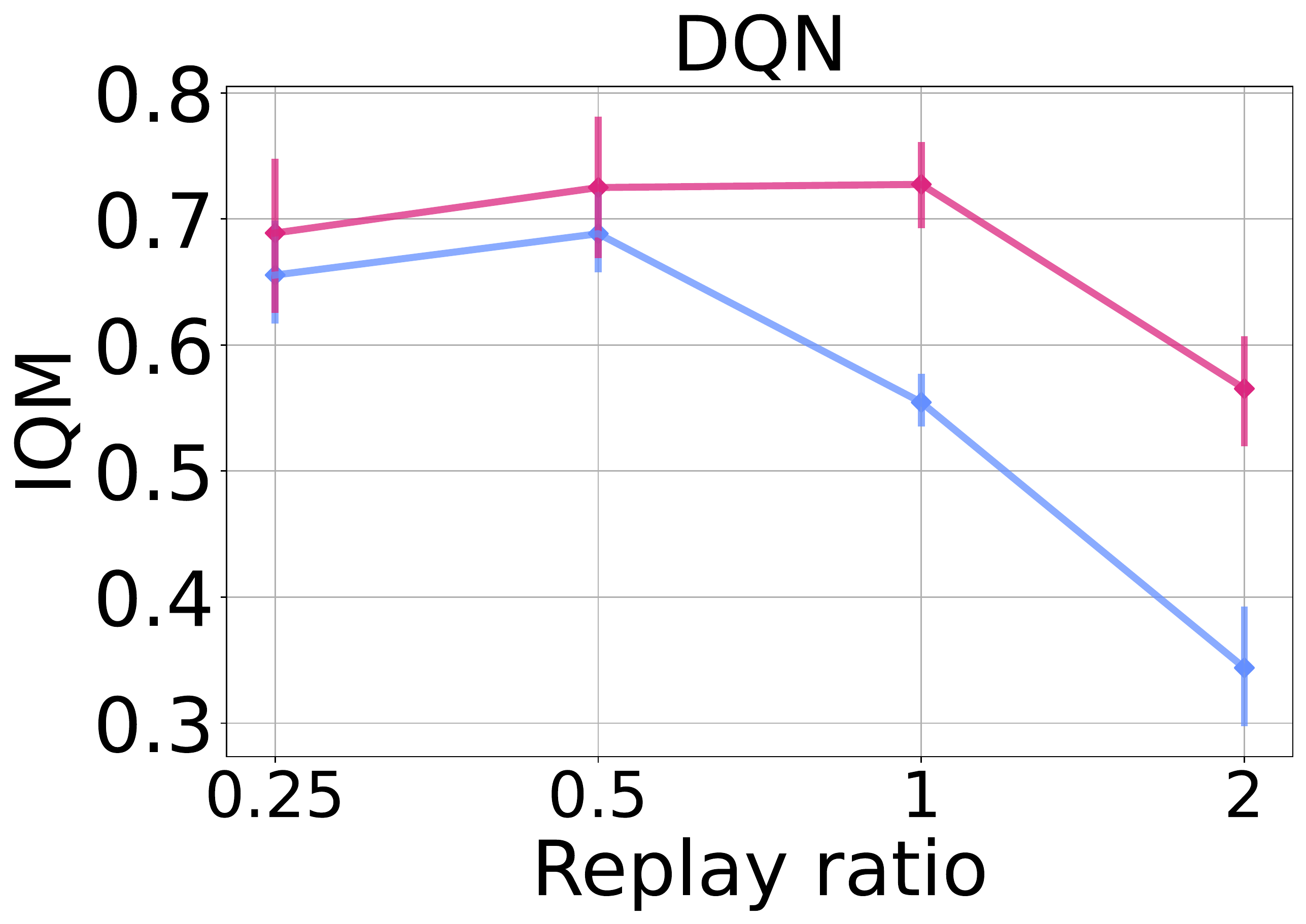}
\includegraphics[width=0.49\columnwidth]{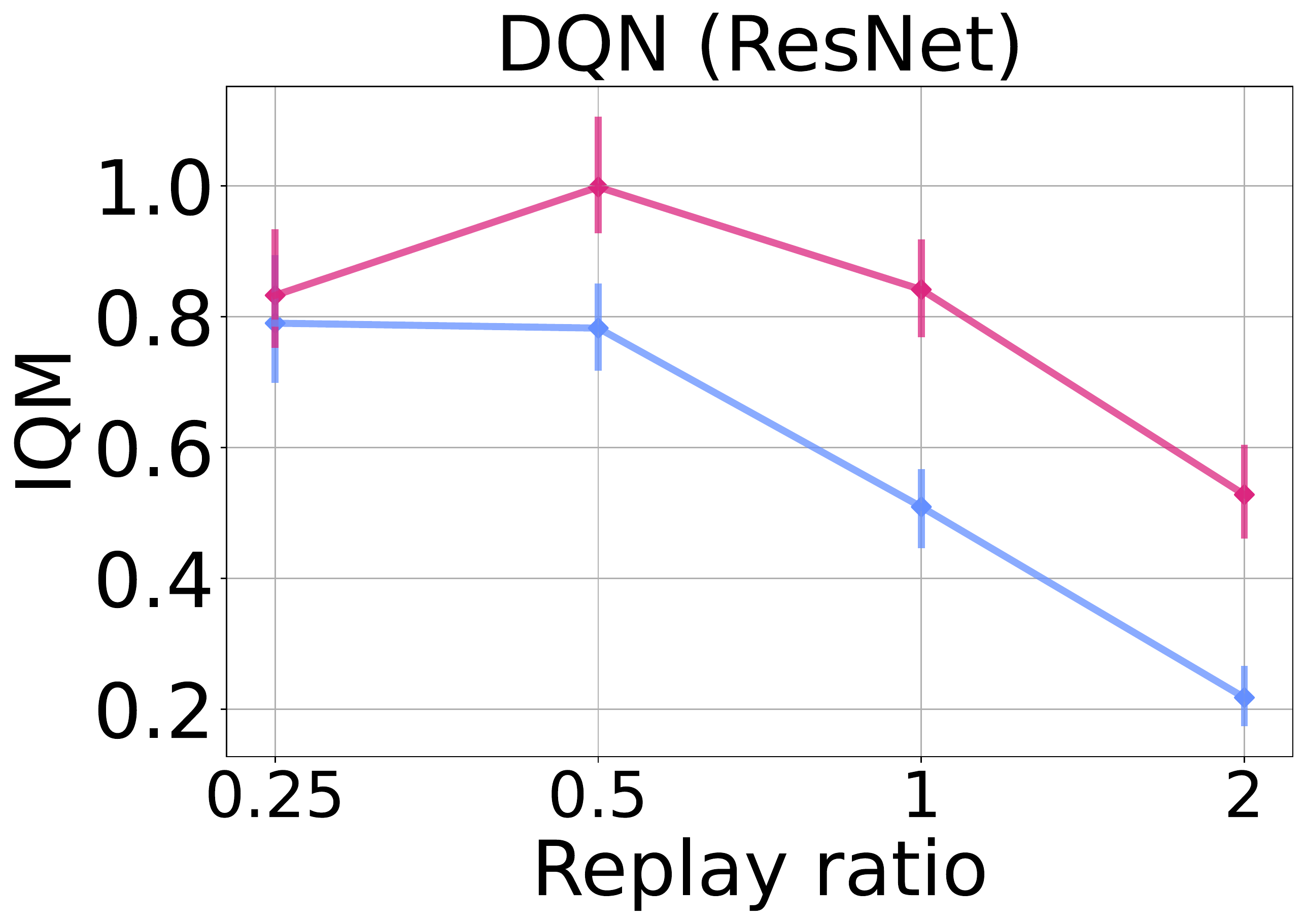}
\includegraphics[width=0.49\columnwidth]{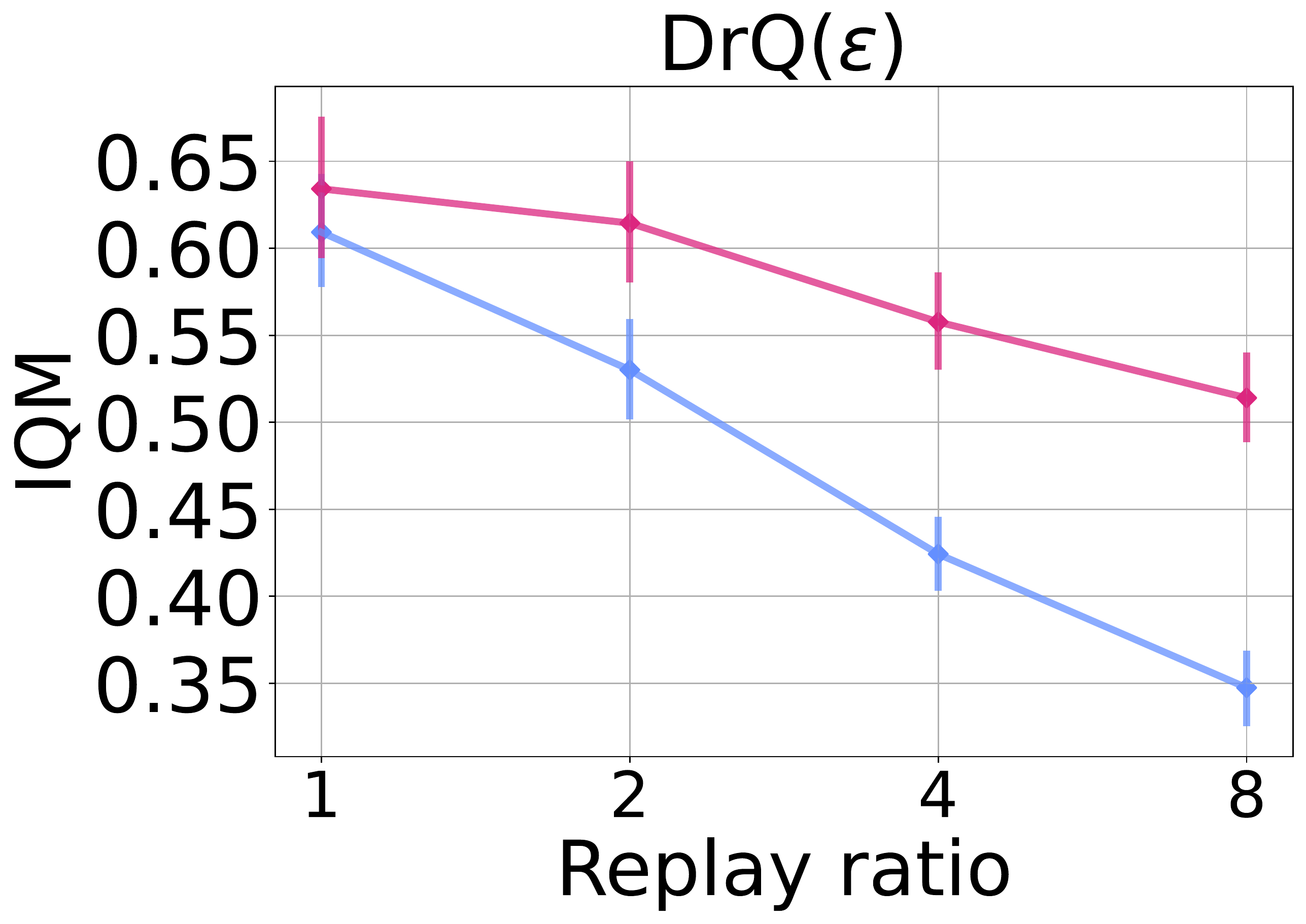}
}
\caption{Evaluating the effect of increased replay ratio with and without {\em ReDo}. From left to right: DQN with default settings, DQN with $n$-step of 3, $DQN$ with the ResNet architecture, and DrQ($\epsilon$). We report results using 5 seeds, while DrQ($\epsilon$) use 10 seeds; error bars report 95\% confidence intervals.}
\label{fig:IQM_RR}
\end{center}
\vskip -0.3in
\end{figure*}

\paragraph{Agents, architectures, and environments.} We evaluate DQN on 17 games from the Arcade Learning Environment \citep{bellemare2013arcade} (as used in \cite{kumar2020implicit,kumar2021dr3} to study the loss of network expressivity). We study two different architectures: the default CNN used by \citet{mnih2015human}, and the ResNet architecture used by the IMPALA agent \citep{espeholt2018impala}. 

Additionally, we evaluate DrQ($\epsilon$) \cite{yarats2021image,agarwal2021deep} on the 26 games used in the Atari 100K benchmark \citep{kaiser2019model}, and SAC \citep{haarnoja2018soft} on four MuJoCo environments \citep{todorov2012mujoco}. 

\paragraph{Implementation details.} All our experiments and implementations were conducted using the Dopamine framework \citep{castro18dopamine}\footnote{Code is available at\\  \url{https://github.com/google/dopamine/tree/master/dopamine/labs/redo}}. For agents trained with {\em ReDo}, we use a threshold of $\tau=0.1$, unless otherwise noted, as we found this gave a better performance than using a threshold of $0$ or $0.025$. %
When aggregating results across multiple games, we report the Interquantile Mean (IQM), recommended by \citet{agarwal2021deep} as a more statistically reliable alternative to median or mean, using 5 independent seeds for each DQN experiment, 10 for the DrQ and SAC experiments, and reporting 95\% stratified bootstrap confidence intervals.

\subsection{Consequences for Sample Efficiency}
Motivated by our finding that higher replay ratios exacerbate dormant neurons and lead to poor performance (\autoref{fig:dormant_in_high_RR}), we investigate whether {\em ReDo} can help mitigate these. To do so, we report the IQM for four replay ratio values: $0.25$ (default for DQN), $0.5$, $1$, and $2$ when training with and without {\em ReDo}. Since increasing the replay ratio increases the training time and cost, we train DQN for 10M frames, as opposed to the regular 200M. As the leftmost plot in \autoref{fig:IQM_RR} demonstrates, {\em ReDo} is able to avoid the performance collapse when increasing replay ratios, and even to benefit from the higher replay ratios when trained with {\em ReDo}.   

\paragraph{Impact on multi-step learning.} In the center-left plot of \autoref{fig:IQM_RR} we added $n$-step returns with a value of $n=3$ \citep{sutton2018reinforcement}. While this change results in a general improvement in DQN's performance, it still suffers from performance collapse with higher replay ratios; {\em ReDo} mitigates this and improves performance across all values.

\paragraph{Varying architectures.} To evaluate {\em ReDo}'s impact on different network architectures, in the center-right plot of \autoref{fig:IQM_RR} we replace the default CNN architecture used by DQN with the ResNet architecture used by the IMPALA agent \cite{espeholt2018impala}. We see a similar trend: {\em ReDo} enables the agent to make better use of higher replay ratios, resulting in improved performance.

\paragraph{Varying agents.} We evaluate on a sample-efficient value-based agent DrQ($\epsilon$) \cite{yarats2021image,agarwal2021deep}) on the Atari 100K benchmark in the rightmost plot of \autoref{fig:IQM_RR}. In this setting, we train for 400K steps, where we can see the effect of dormant neurons on performance, and study the following replay ratio values: $1$ (default), $2$, $4$, $8$. Once again, we observe {\em ReDo}'s effectiveness in improving performance at higher replay ratios.

In the rest of this section, we do further analyses to understand the improved performance of {\em ReDo} and how it fares against related methods. We perform this study on a DQN agent trained with a replay ratio of 1 using the default CNN architecture.

\subsection{Learning Rate Scaling}
\begin{figure}[t]
\vskip 0.2in
\begin{center}
\includegraphics[width=0.42\columnwidth]{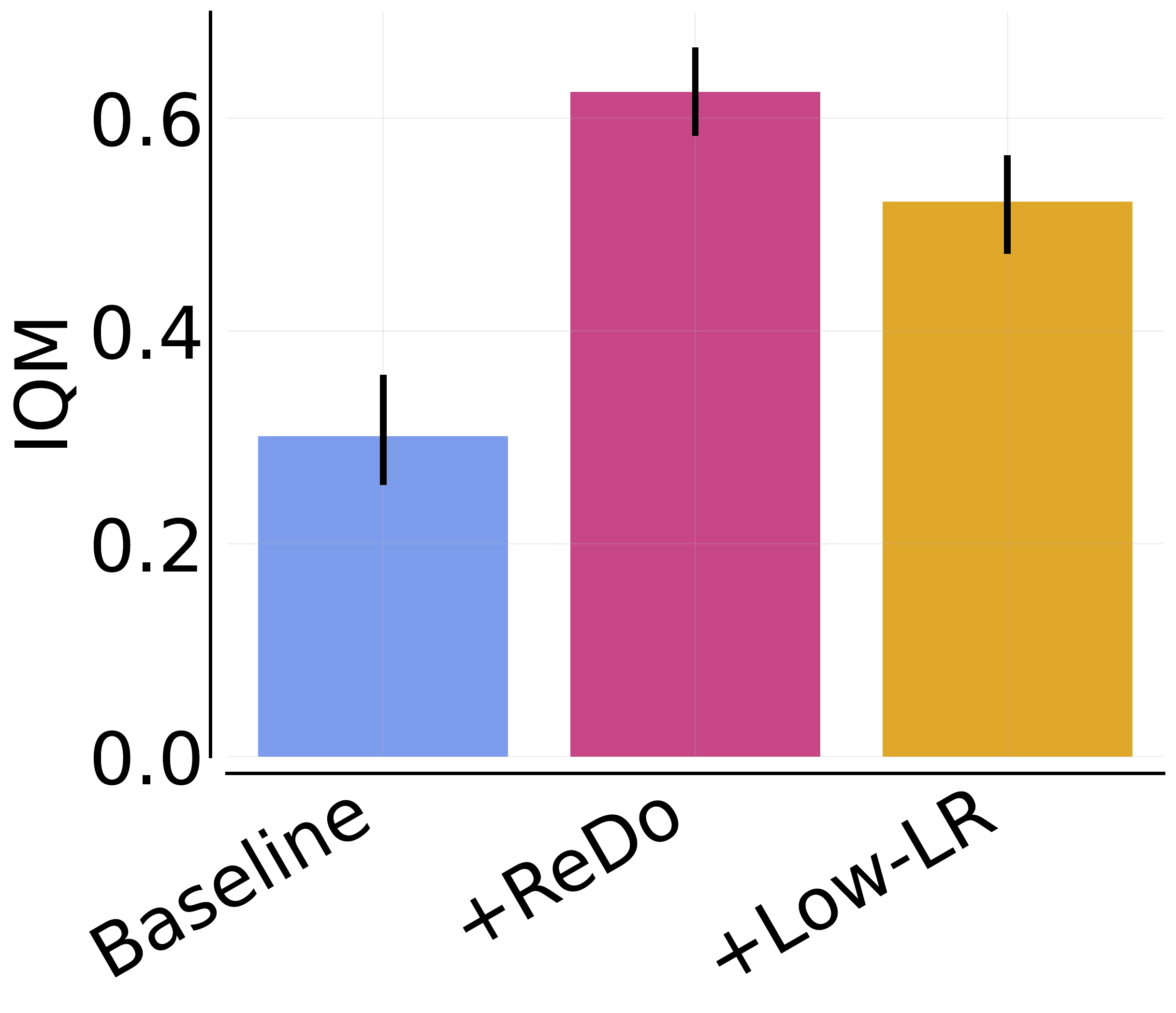}
\includegraphics[width=0.55\columnwidth]{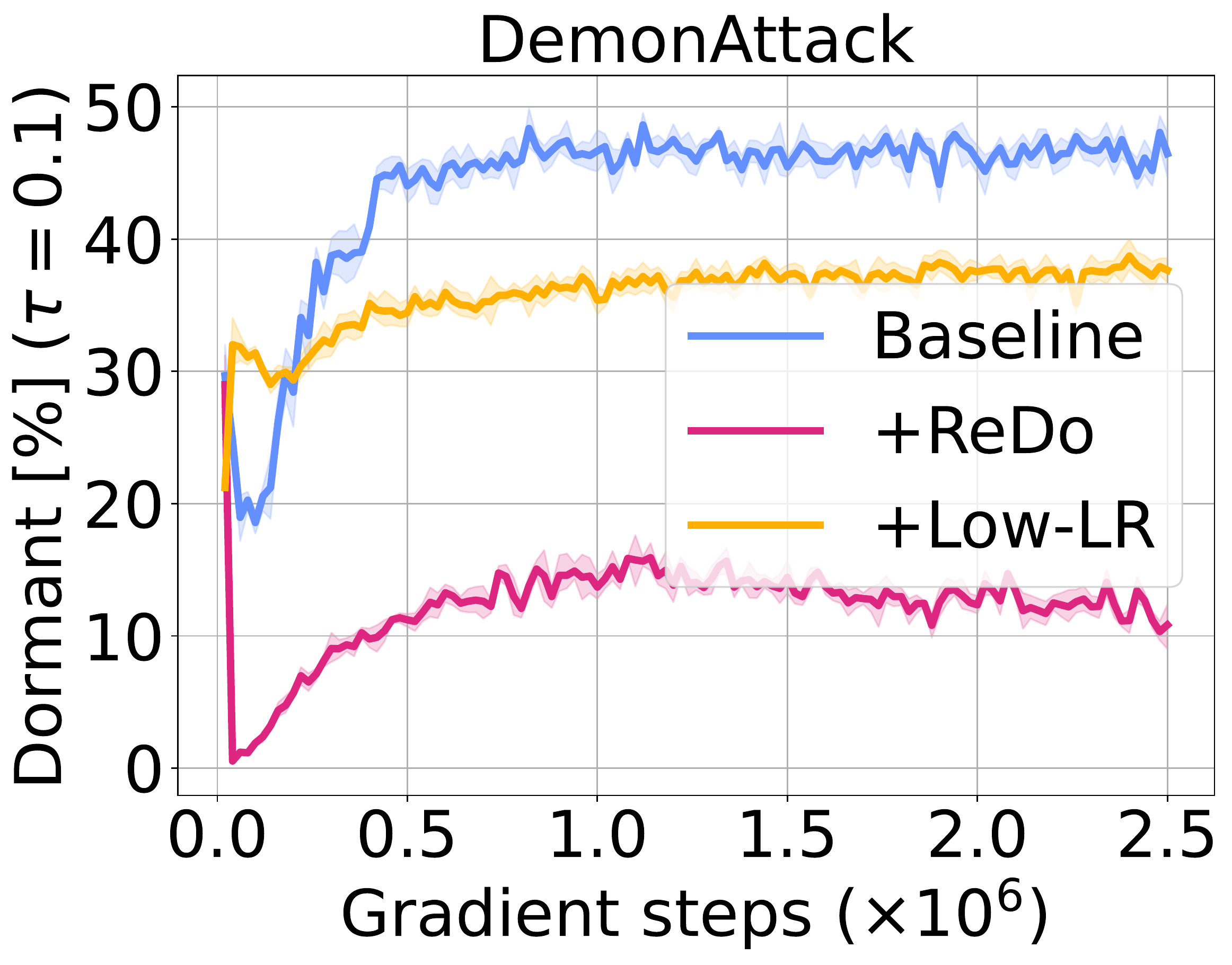}
\caption{Effect of reduced learning rate in high replay ratio setting. Scaling learning rate helps, but does not solve the dormant neuron problem. Aggregated results across 17 games (left) and the percentage of dormant neurons during training on DemonAttack (right).}
\label{fig:lr_scale}
\end{center}
\vskip -0.2in
\end{figure}

An important point to consider is that the default learning rate may not be optimal for higher replay ratios. Intuitively, performing more gradient updates would suggest a {\em reduced} learning rate would be more beneficial. To evaluate this, we decrease the learning rate by a factor of four when using a replay ratio of $1$ (four times the default value). \autoref{fig:lr_scale} confirms that a lower learning rate reduces the number of dormant neurons and improves performance. However, percentage of dormant neurons is still high and using {\em ReDo} with a high replay ratio and the default learning rate obtains the best performance. 

\subsection{Is Over-parameterization Enough?}
\begin{figure}[t]
\vskip 0.2in
\begin{center}
\centerline{
\includegraphics[width=0.54\columnwidth]{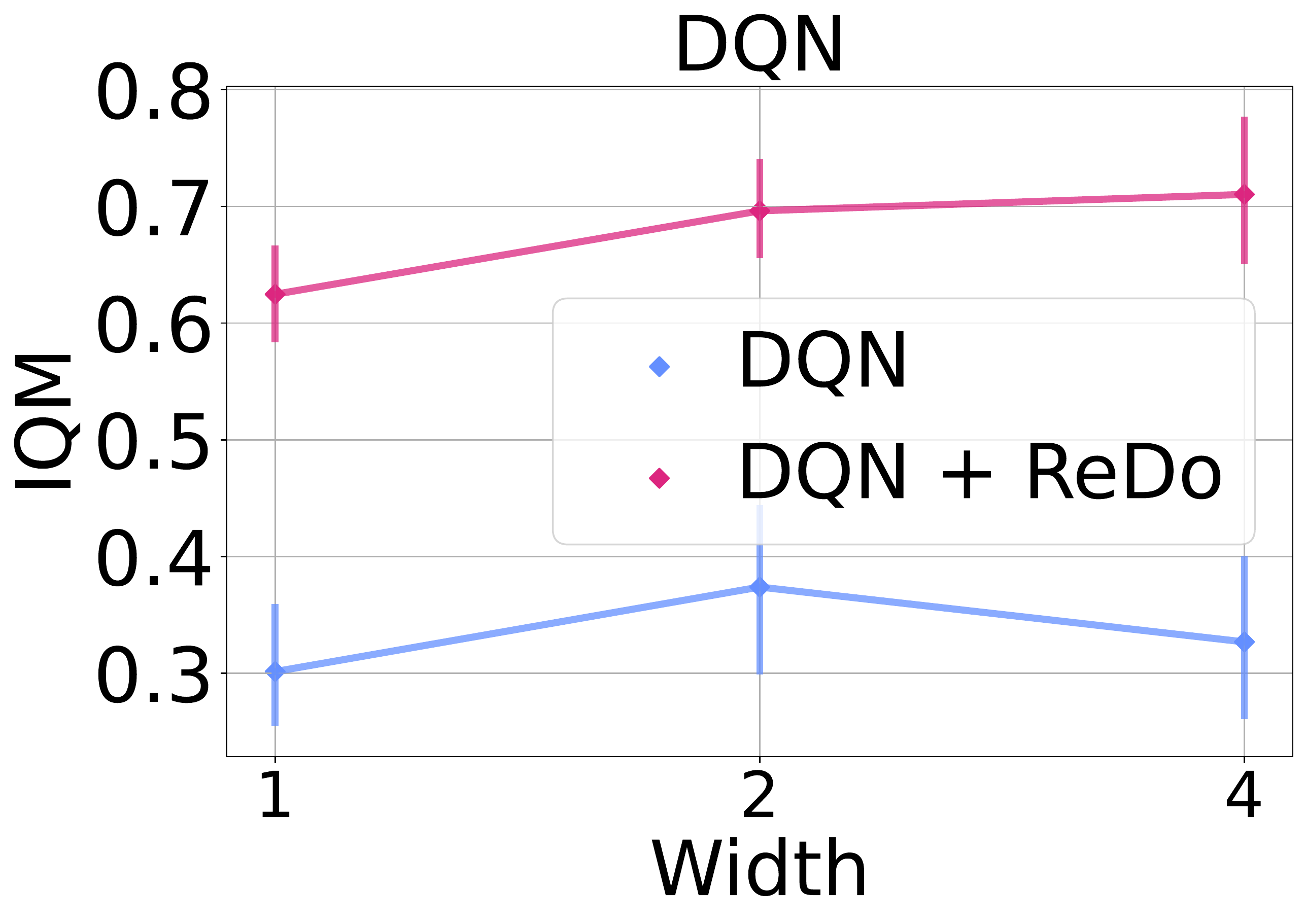}\hspace{0.2cm}
\includegraphics[width=0.48\columnwidth]{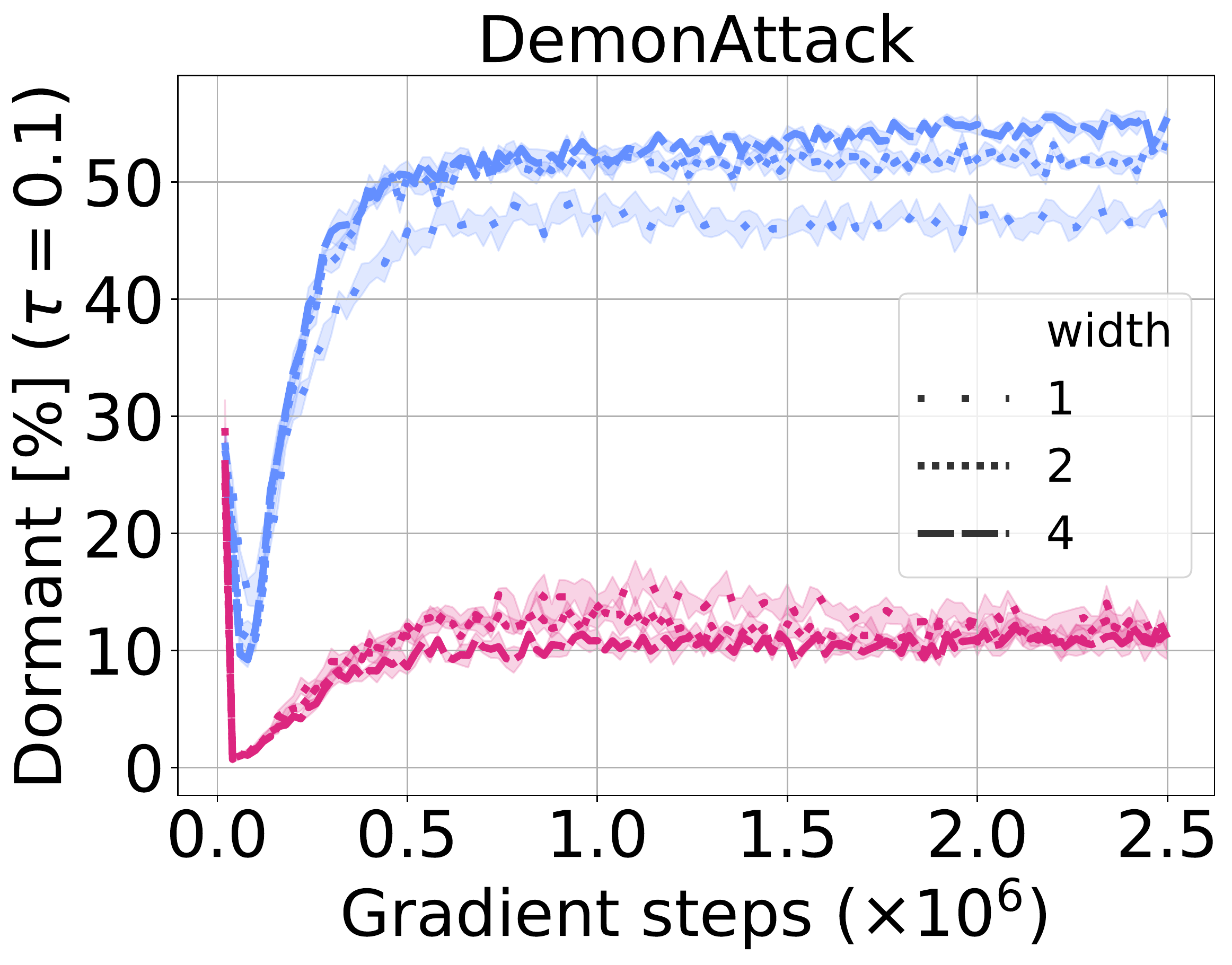}
}
\caption{Performance of DQN trained with $RR = 1 $ using different network width. Increasing the width of the network slightly improves the performance. Yet, the performance gain does not reach the gain obtained by {\em ReDo}. {\em ReDo} improves the performance across different network sizes.}
\label{fig:IQM_RR1_width}
\end{center}
\vskip -0.2in
\end{figure}

\citet{lyle2021understanding} and \citet{fu2019diagnosing} suggest sufficiently over-parameterized networks can fit new targets over time; this raises the question of whether over-parameterization can help address the dormant neuron phenomenon. To investigate this, we increase the size of the DQN network by doubling and quadrupling the width of its layers (both the convolutional and fully connected). %
The left plot in \autoref{fig:IQM_RR1_width} shows that larger networks have at most a mild positive effect on the performance of DQN, and the resulting performance is still far inferior to that obtained when using {\em ReDo} with the default width. Furthermore, training with {\em ReDo} seems to improve as the network size increases, suggesting that the agent is able to better exploit network parameters, compared to when training without {\em ReDo}.

An interesting finding in the right plot in \autoref{fig:IQM_RR1_width} is that the percentage of dormant neurons is similar across the varying widths. As expected, the use of {\em ReDo} dramatically reduces this number for all values. This finding is somewhat at odds with that from
\citet{sankararaman2020impact}. They demonstrated that, in supervised learning settings, increasing the width decreases the gradient confusion and leads to faster training. If this observation would also hold in RL, we would expect to see the percentage of dormant neurons decrease in larger models.

\begin{figure}[t]
\begin{center}
\includegraphics[width=0.45\columnwidth]{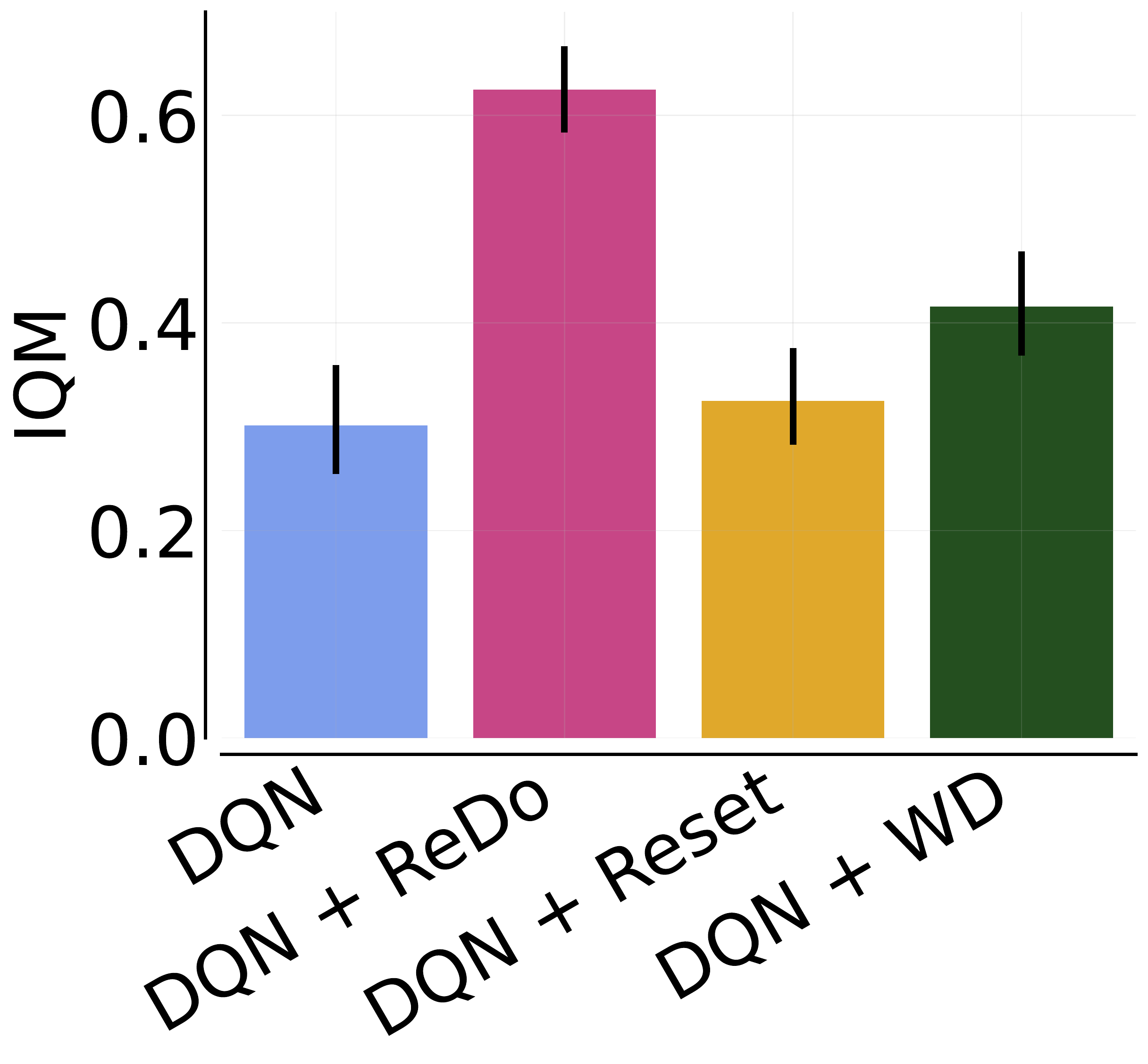}\hfill
\includegraphics[width=0.53\columnwidth]{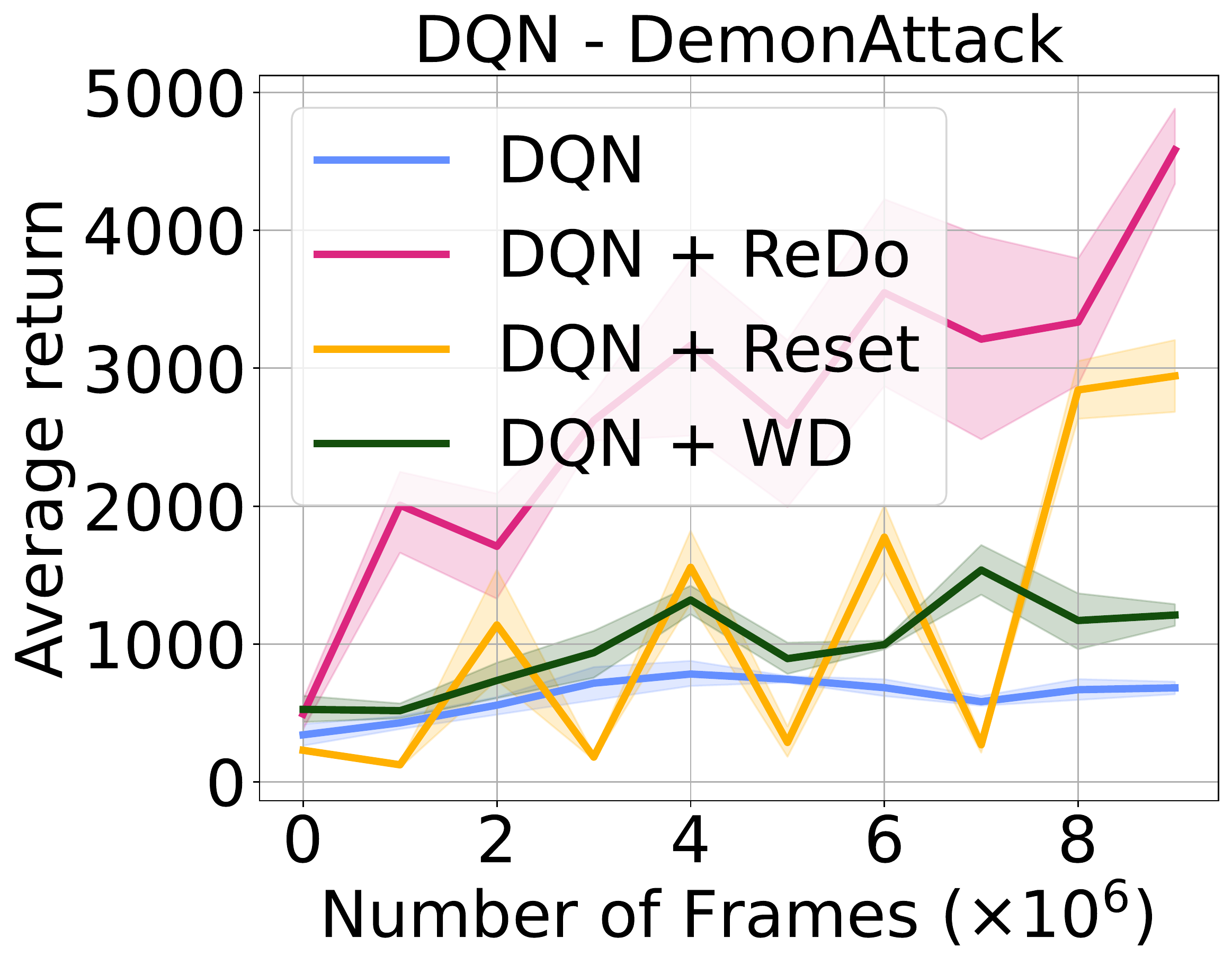}
\caption{Comparison of the performance for {\em ReDo} and two different regularization methods (Reset \cite{nikishin2022primacy} and weight decay (WD)) when integrated with training DQN agents. Aggregated results across 17 games (left) and the learning curve on DemonAttack (right). }
\label{fig:IQM_RR1_baselines}
\end{center}
\vskip -0.2in
\end{figure}

\subsection{Comparison with Related Methods}
\citet{nikishin2022primacy} also observed performance collapse when increasing the replay ratio, but attributed this to overfitting to early samples (an effect they refer to as the ``primacy bias''). To mitigate this, they proposed periodically resetting the network, which can be seen as a form of regularization. We compare the performance of {\em ReDo} against theirs, which periodically resets only the penultimate layer for Atari environments. Additionally, we compare to adding weight decay, as this is a simpler, but related, form of regularization.  It is worth highlighting that \citet{nikishin2022primacy} also found high values of replay ratio to be more amenable to their method. As \autoref{fig:IQM_RR1_baselines} illustrates, weight decay is comparable to periodic resets, but {\em ReDo} is superior to both. 

\begin{figure}[t]
\vskip 0.2in
\begin{center}
{
\includegraphics[width=0.48\columnwidth]{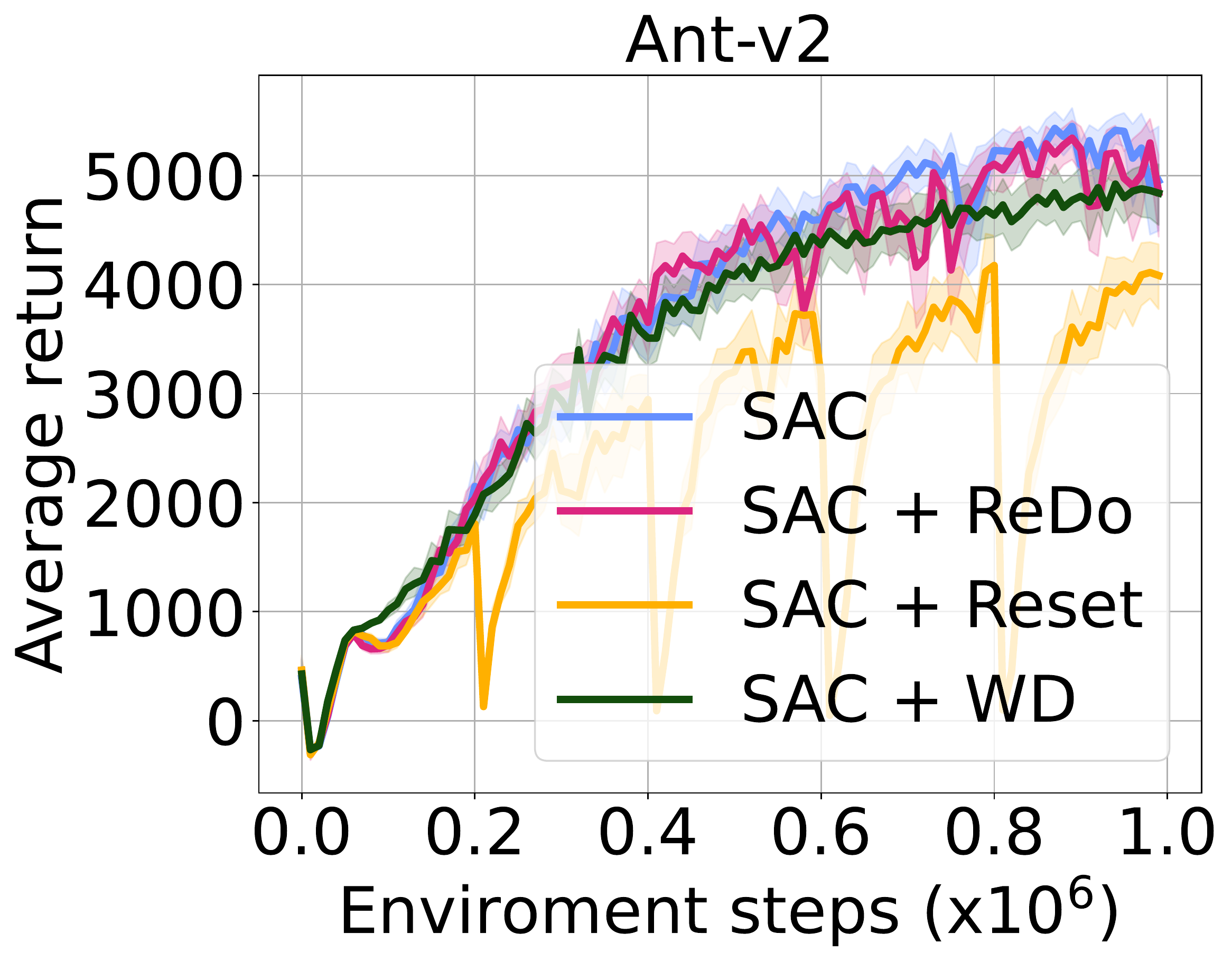}\hfill
\includegraphics[width=0.48\columnwidth]{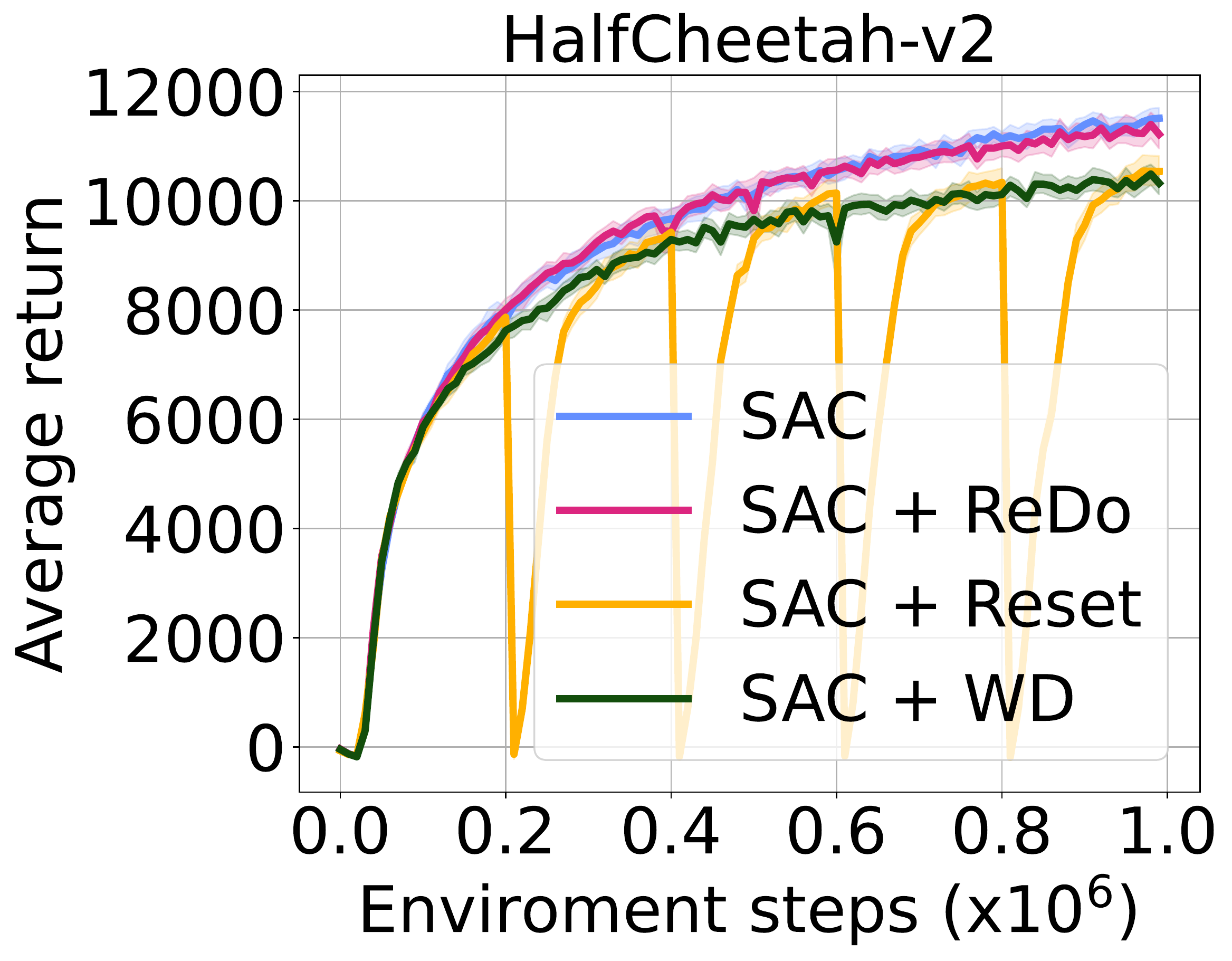}
}

\caption{Comparison of the performance of SAC agents with {\em ReDo} and two different regularization methods (Reset \cite{nikishin2022primacy} and weight decay (WD)). See \autoref{fig:appendix:redo_reset_mujoco} for other environments.}
\label{fig:redo_reset_mujoco}
\end{center}
\vskip -0.2in
\end{figure}

We continue our comparison with resets and weight decay on two MuJoCo environments with the SAC agent \citep{haarnoja2018soft}. As \autoref{fig:redo_reset_mujoco} shows, {\em ReDo} is the only method that does not suffer a performance degradation. The results on other environments can be seen in Appendix \ref{appendix:dormant_phenomenon_more_domains}.

\subsection{Neuron Selection Strategies}
Finally, we compare our strategy for selecting the neurons that will be recycled (Section \ref{sec:analysis}) against two alternatives: (1) {\bf Random:} neurons are selected randomly, and (2) {\bf Inverse {\em ReDo}:} neurons with the {\em highest} scores according to \autoref{eqn:neuronScore} are selected. To ensure a fair comparison, the number of recycled neurons is a fixed percentage for all methods, occurring every 1000 steps. The percentage of neurons to recycle follows a cosine schedule starting at 0.1 and ending at 0. As \autoref{fig:different_scores} shows, recycling active or random neurons hinders learning and causes performance collapse.

%% file: 6.related_work.tex
\section{Related Work}

\paragraph{Function approximators in RL.} The use of over-parameterized neural networks as function approximators was instrumental to some of the successes in RL, such as achieving superhuman performance on Atari 2600 games \cite{mnih2015human} and continuous control \cite{lillicrap2016continuous}. Recent works observe a change in the network's capacity over the course of training, which affects the agent's performance. \citet{kumar2020implicit, kumar2021dr3} show that the expressivity of the network decreases gradually due to bootstrapping. \citet{gulcehre2022empirical} investigate the sources of expressivity loss in offline RL and observe that underparamterization emerges with prolonged training. \citet{lyle2021understanding} demonstrate that RL agents lose their ability to fit new target functions over time, due to the non-stationary in the targets. Similar observations have been found, referred to as plasticity loss, in the continual learning setting where the data distribution is changing over time \cite{DBLP:journals/corr/abs-2106-00042,dohare2021continual}. These observations call for better understanding how RL learning dynamics affect the capacity of their neural networks.

\begin{figure}[t]
\vskip 0.2in
\begin{center}
\includegraphics[width=0.49\columnwidth]{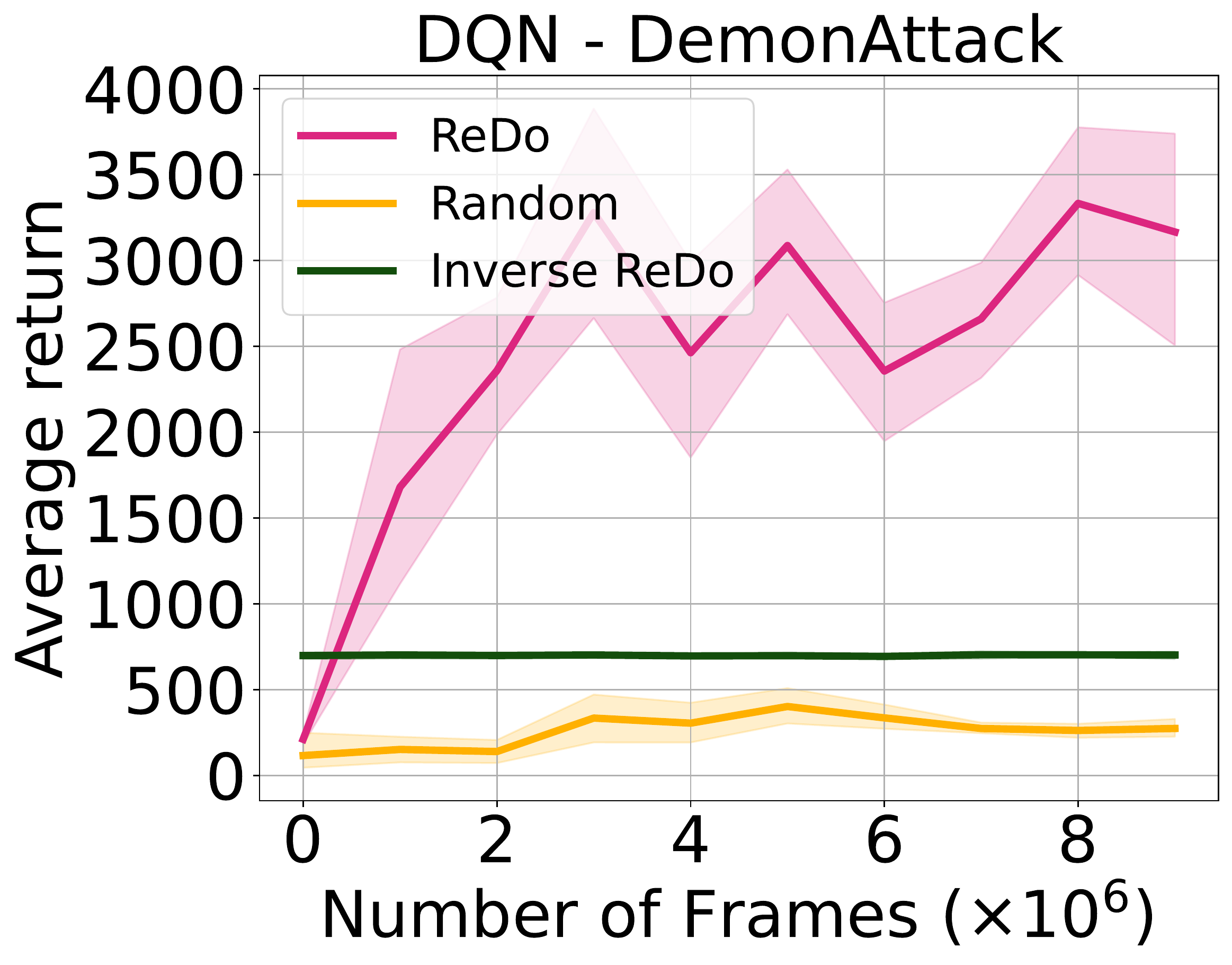}
\includegraphics[width=0.49\columnwidth]{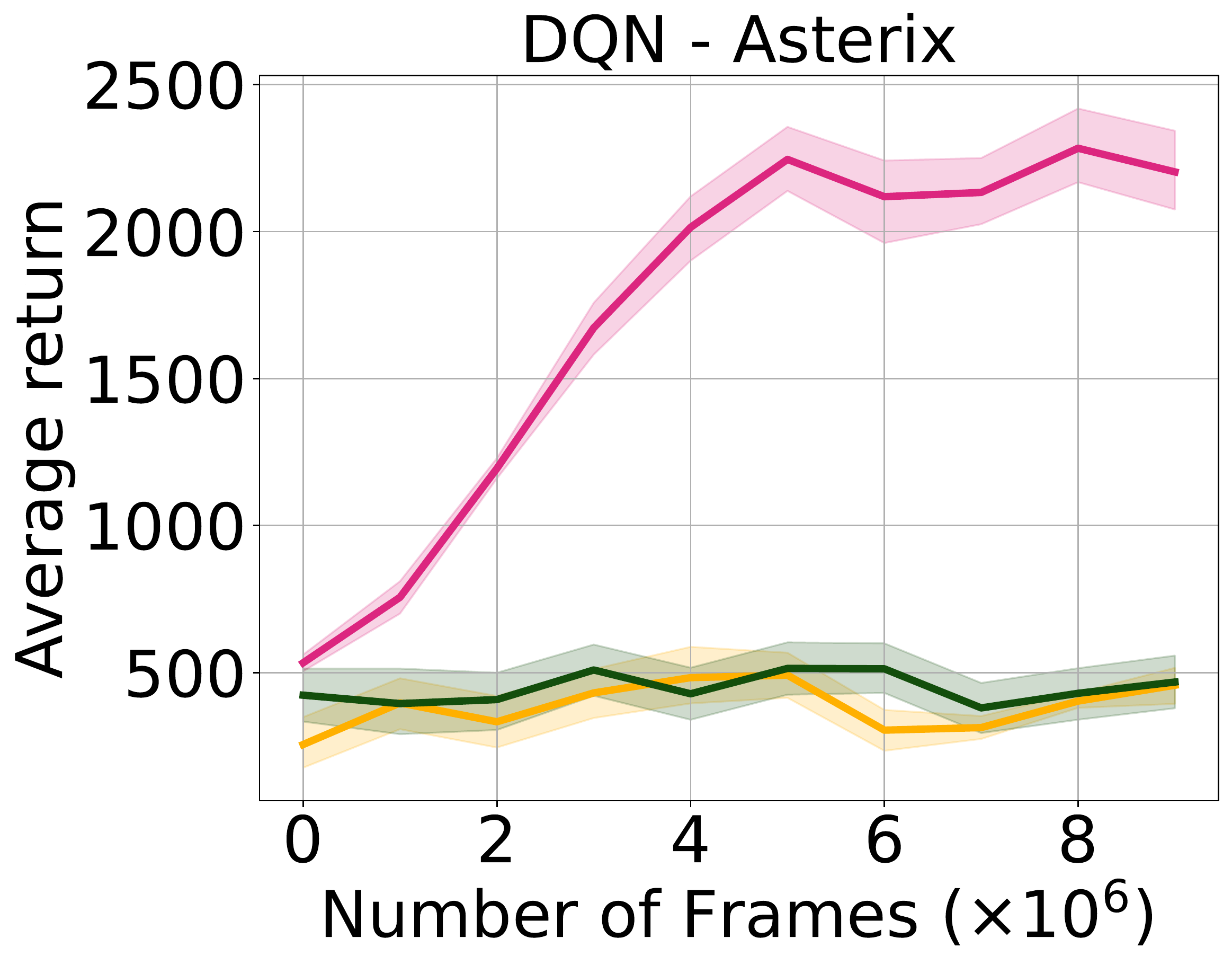}
\caption{Comparison of different strategies for selecting the neurons that will be recycled. Recycling neurons with the highest score (Inverse {\em ReDo}) or random neurons causes performance collapse.}
\label{fig:different_scores}
\end{center}
\vskip -0.1in
\end{figure}

There is a recent line of work investigating network topologies by using sparse neural networks in online \cite{graesser2022state,sokar2022dynamic,tan2022rlx2} and offline RL \cite{arnob2021single}. They show up to 90\% of the network's weights can be removed with minimal loss in performance. This suggests that RL agents are not using the capacity of the network to its full potential. %

\paragraph{Generalization in RL.} RL agents are prone to overfitting, whether it is to training environments, reducing their ability to generalize to unseen environments \cite{kirk2021survey}, or to early training samples, which degrades later training performance \cite{fu2019diagnosing,nikishin2022primacy}. Techniques such as regularization \cite{hiraoka2021dropout,NEURIPS2020_5a751d6a}, ensembles \cite{chen2020randomized}, or data augmentation \cite{fan2021secant,janner2019trust,NEURIPS2021_1e0f65eb} have been adopted to account for overfitting.

Another line of work addresses generalization via \textit{re-initializing} a subset or all of the weights of a neural network during training. This technique is mainly explored in supervised learning \cite{taha2021knowledge,zhou2021fortuitous,alabdulmohsin2021impact,zaidi2022does}, transfer learning  \cite{li2020rifle}, and online learning \cite{ash2020warm}. A few recent works have explored this for RL: \citet{igl2020transient} periodically reset an agent’s full network and then performs distillation from the pre-reset network. \citet{nikishin2022primacy} (already discussed in \autoref{fig:IQM_RR1_baselines}) periodically resets the last layers of an agent’s network. 
Despite its performance gains, fully resetting some or all layers can lead to the agent ``forgetting'' prior learned knowledge. The authors account for this by using a sufficiently large replay buffer, so as to never discard any observed experience; this, however, makes it difficult to scale to environments with more environment interactions. Further, recovering performance after each reset requires many gradient updates. Similar to our approach, \citet{dohare2021continual} adapt the stochastic gradient descent by resetting the smallest utility features for {\em continual learning}. We compare their utility metric to the one used by {\em ReDo} in Appendix \ref{appendix:cbp} and observe similar or worse performance. 

\paragraph{Neural network growing.} A related research direction is to {\em prune} and {\em grow} the architecture of a neural network. On the growing front, \citet{evci2021gradmax} and \citet{dai2019nest} proposed gradient-based strategies to grow new neurons in dense and sparse networks, respectively. \citet{yoon2018lifelong} and \citet{wu2019splitting} proposed methods to split existing neurons. \citet{zhou2012online} adds new neurons and merges similar features for online learning.

%% file: 7.conclusion.tex
\section{Discussion and Conclusion}
In this work we identified the {\em dormant neuron phenomenon} whereby, during training, an RL agent's neural network exhibits an increase in the number of neurons with little-or-no activation. We demonstrated that this phenomenon is present across a variety of algorithms and domains, and provided evidence that it does result in reduced expressivity and inability to adapt to new tasks.  
 
Interestingly, studies in neuroscience have found similar types of dormant neurons (precursors) in the adult brain of several mammalian species, including humans \cite{benedetti2022would}, albeit with different dynamics. Certain brain neurons {\em start off} as dormant during embryonic development, and progressively awaken with age, eventually becoming mature and functionally integrated as excitatory neurons  \cite{rotheneichner2018cellular,benedetti2020functional,benedetti2022would}. Contrastingly, the dormant neurons we investigate here emerge over time and exacerbate with more gradient updates. 

To overcome this issue, we proposed a simple method ({\em ReDo}) to maintain network utilization throughout training by periodic recycling of dormant neurons. The simplicity of {\em ReDo} allows for easy integration with existing RL algorithms. Our experiments suggest that this can lead to improved performance. Indeed, the results in \autoref{fig:IQM_RR} and \ref{fig:IQM_RR1_width} suggest that {\em ReDo} can be an important component in being able to successfully scale RL networks in a sample-efficient manner.

\paragraph{Limitations and future work.} Although the simple approach of recycling neurons we introduced yielded good results, it is possible that better approaches exist. For example, {\em ReDo} reduces dormant neurons significantly but it doesn't completely eliminate them. Further research on initialization and optimization of the recycled capacity can address this and lead to improved performance. Additionally, the dormancy threshold is a hyperparameter that requires tuning; having an adaptive threshold over the course of training could improve performance even further. Finally, further investigation into the relationship between the task’s complexity, network capacity, and the dormant neuron phenomenon would provide a more comprehensive understanding.

Similarly to the findings of \citet{graesser2022state}, this work suggests there are important gains to be had by investigating the network architectures and topologies used for deep reinforcement learning. Moreover, the observed network's behavior during training (i.e. the change in the network capacity utilization), which differs from supervised learning, indicates a need to explore optimization techniques specific to reinforcement learning due to its unique learning dynamics.

%% file: acknowledgements.tex
\section*{Acknowledgements}
We would like to thank Max Schwarzer, Karolina Dziugaite, Marc G. Bellemare, Johan S. Obando-Ceron, Laura Graesser, Sara Hooker and Evgenii Nikishin, as well as the rest of the Brain Montreal team for their feedback on this work.  We would also like to thank the Python community \cite{van1995python, 4160250}
for developing tools that enabled this work, including NumPy \cite{harris2020array}, Matplotlib \cite{hunter2007matplotlib} and JAX \cite{bradbury2018jax}.

%% file: 8.appendix.tex
\section*{Author Contributions}
\begin{itemize}
    \item Ghada: Led the work, worked on project direction and plan, participated in discussions, wrote most of the code, ran most of the experiments, led the writing, and wrote the draft of the paper.
    \item Rishabh: Advised on project direction and participated in project discussions, ran an offline RL experiment, worked on the plots and helped with paper writing.
    \item Pablo: Worked on project direction and plan, participated in discussions throughout the project, helped with reviewing code, ran some experiments, worked substantially on paper writing, supervised Ghada.
    \item Utku: Proposed project direction and the initial project plan, reviewed and open-sourced the code, ran part of the experiments, worked on the plots and helped with paper writing, supervised Ghada.
\end{itemize}
\section{Experimental Details}
\label{appendix:experimental_details}
\begin{table}[t]
\caption{Common Hyper-parameters for DQN and DrQ($\epsilon$).}
\label{table:common_hyperparameters}
\vskip 0.15in
\begin{center}
\begin{tabular}{lr}
\toprule
Parameter & Value \\
\midrule
Optimizer & Adam \cite{KingmaB14}\\
Optimizer: $\epsilon$ & $1.5 \times 10^{-4}$ \\
Training $\epsilon$ & 0.01 \\
Evaluation $\epsilon$ & 0.001 \\
Discount factor & 0.99 \\
Replay buffer size & $10^6$ \\
Minibatch size & 32 \\
Q network: channels & 32, 64, 64 \\
Q-network: filter size & 8 $\times$ 8, 4 $\times$ 4, 3 $\times$ 3 \\
Q-network: stride & 4, 2, 1 \\ 
Q-network: hidden units &  512 \\
\hline
Recycling period & 1000 \\
$\tau$-Dormant  & 0.025 for default setting, 0.1 otherwise\\
Minibatch size for estimating neurons score & 64\\
\bottomrule
\end{tabular}
\end{center}
\vskip -0.1in
\end{table}

\begin{table}[t]
\caption{Hyper-parameters for DQN.}
\label{table:DQN_hyperparameters}
\vskip 0.15in
\begin{center}
\begin{tabular}{lr}
\toprule
Parameter & Value \\
\midrule
Optimizer: Learning rate & $6.25 \times 10^{-5}$ \\
Initial collect steps & 20000 \\
$n$-step & 1 \\
Training iterations & Default setting: 40, otherwise: 10  \\
Training environment steps per iteration & 250K \\
(Updates per environment step, Target network update period)  & (0.25, 8000) \\
 & (0.5, 4000) \\ 
 & (1, 2000) \\ 
 & (2, 1000) \\ 
\bottomrule
\end{tabular}
\end{center}
\vskip -0.1in
\end{table}

\begin{table}[t]
\caption{Hyper-parameters for DrQ($\epsilon$).}
\label{table:DrQ_hyperparameters}
\vskip 0.15in
\begin{center}
\begin{tabular}{lr}
\toprule
Parameter & Value \\
\midrule
Optimizer: Learning rate &  $1 \times 10^{-4}$ \\
Initial collect steps & 1600 \\
$n$-step & 10 \\
Training iterations & 40 \\
Training environment steps per iteration & 10K \\
Updates per environment step & 1, 2, 4, 8 \\
\bottomrule
\end{tabular}
\end{center}
\vskip -0.1in
\end{table}

\begin{table}[t]
\caption{Hyper-parameters for SAC.}
\label{table:SAC_hyperparameters}
\vskip 0.15in
\begin{center}
\begin{tabular}{lr}
\toprule
Parameter & Value \\
\midrule
Initial collect steps & 10000 \\
Discount factor & 0.99 \\
Training environment steps & $10^6$ \\
Replay buffer size & $10^6$ \\
Updates per environment step (Replay Ratio) & 1, 2, 4, 8 \\ 
Target network update period & 1\\
target smoothing coefficient $\tau$ & 0.005\\
Optimizer & Adam \cite{KingmaB14}\\
Optimizer: Learning rate &  $3 \times 10^{-4}$\\
Minibatch size & 256 \\
Actor/Critic: Hidden layers & 2 \\
Actor/Critic: Hidden units & 256 \\
\hline
Recycling period & 200000 \\
$\tau$-Dormant &  0\\
Minibatch size for estimating neurons score & 256\\
\bottomrule
\end{tabular}
\end{center}
\vskip -0.1in
\end{table}

\paragraph{Discrete control tasks.} We evaluate DQN \cite{mnih2015human} on 17 games from the Arcade Learning Environment \cite{bellemare2013arcade}: Asterix, Demon Attack, Seaquest,  Wizard of Wor, Bream Reader, Road Runner, James Bond, Qbert, Breakout, Enduro, Space Invaders, Pong, Zaxxon, Yars’ Revenge,  Ms. Pacman, Double Dunk, Ice Hockey. This set is used by previous works \cite{kumar2020implicit,kumar2021dr3} to study the {\em implicit under-parameterization} phenomenon in offline RL. For hyper-parameter tuning, we used five games (Asterix, Demon Attack, Seaquest, Breakout, Beam Rider). We evaluate DrQ($\epsilon$) on the 26 games of Atari 100K \cite{kaiser2019model}. We used the best hyper-parameters found for DQN in training DrQ($\epsilon$). 

\paragraph{Continuous control tasks.} We evaluate SAC \cite{haarnoja2018soft} on four environments from MuJoCo suite \cite{todorov2012mujoco}: HalfCheetah-v2, Hopper-v2, Walker2d-v2, Ant-v2.

\paragraph{Code.} For discrete control tasks, we build on the implementation of DQN and DrQ provided in Dopamine \cite{castro18dopamine}, including the architectures used for agents. The hyper-parameters are provided in Tables \ref{table:common_hyperparameters},  \ref{table:DQN_hyperparameters}, and \ref{table:DrQ_hyperparameters}.  For continuous control, we build on the SAC implementation in TF-Agents \cite{TFAgents} and the codebase of \cite{graesser2022state}. The hyper-parameters are provided in \autoref{table:SAC_hyperparameters}.  

\paragraph{Evaluation.} We follow the recommendation from \cite{agarwal2021deep} to report reliable aggregated results across games using the interquartile mean (IQM). IQM is the calculated mean after discarding the
bottom and top 25\% of normalized scores aggregated from multiple runs and games. 

\paragraph{Baselines.} For weight decay, we searched over the grid [$10^{-6}, 10^{-5}, 10^{-4}, 10^{-3}$]. The best found value is $10^{-5}$. For reset \cite{nikishin2022primacy}, we consider re-initializing the last layer for Atari games (same as the original paper). They use a reset period of $2 \times 10^4$ in for Atari 100k \cite{kaiser2019model}, which corresponds to having 5 restarts in a training run. Since we run longer experiments, we searched over the grid [$5 \times 10^4, 1 \times 10^5, 2.5 \times 10^5, 5 \times 10^5$] gradient steps for the reset period which corresponds to having 50, 25, 10 and 5 restarts per training (10M frames, replay ratio 1). The best found period is $1 \times 10^5$. For SAC, we reset agent's networks entirely every $2 \times 10^5$ environment steps, following the original paper.

\paragraph{Replay ratio.} For DQN, we evaluate replay ratio values: \{0.25 (default), 0.5, 1, 2\}. Following \cite{van2019use}, we scale the target update period based on the value of the replay ratio as shown in \autoref{table:DQN_hyperparameters}. For DrQ($\epsilon$), we evaluate the values: \{1 (default), 2, 4, 8\}.

\paragraph{{\em ReDo} hyper-parameters.} We did the hyper-parameter search for DQN trained with $RR = 1$ using the nature CNN architecture. We searched over the grids [1000, 10000, 100000] and [0, 0.01, 0.1] for the recycling period and $\tau$-dormant, respectively. We apply the best values found to all other settings of DQN, including the ResNet architecture and DrQ($\epsilon$), as reported in \autoref{table:common_hyperparameters}. 

\paragraph{Dormant neurons in supervised learning.} Here we provide the experimental details of the supervised learning analysis illustrated in Section \ref{sec:analysis}. We train a convolutional neural network on CIFAR-10 \cite{krizhevsky2009learning} using stochastic gradient descent and cross-entropy loss. We select 10000 samples from the dataset to reduce the computational cost. We analyze the dormant neurons in two supervised learning settings: (1) training a network with \textit{fixed targets}, the standard single-task supervised learning, where we train a network using the inputs and labels of CIFAR-10 for 100 epochs, and (2) training a network with \textit{non-stationary targets}, where we shuffle the labels every 20 epochs to generate new targets. \autoref{table:CIFAR_hyperparameters} provides the details of the network architecture and training hyper-parameters.

\begin{table}[t]
\caption{Hyperparameters for CIFAR-10.}
\label{table:CIFAR_hyperparameters}
\vskip 0.15in
\begin{center}
\begin{tabular}{lr}
\toprule
Parameter & Value \\
\midrule
Optimizer & SGD \\
Minibatch size & 256 \\
Learning rate & 0.01 \\
Momentum & 0.9 \\
\hline
Architecture: & \\
Layer & (channels, kernel size, stride) \\
Convolution & (32, 3, 1)  \\
Convolution & (64, 3, 1) \\
MaxPool & (-, 2, 2) \\
Convolution & (64, 3, 1) \\
MaxPool & (-, 2, 2) \\ 
Dense &  (128, -, -) \\
\bottomrule
\end{tabular}
\end{center}
\vskip -0.1in
\end{table}

\paragraph{Learning ability of networks with dormant neurons.} Here we present the details of the regression experiment provided in Section \ref{sec:analysis}. Inputs and targets for regression come from a DQN agent trained on DemonAttack for 40M frames with the default hyper-parameters. The pre-trained network was trained for 40M frames using a replay ratio of 1.  

\begin{figure*}
\vskip 0.2in
\begin{center}
{
\includegraphics[width=0.19\columnwidth]{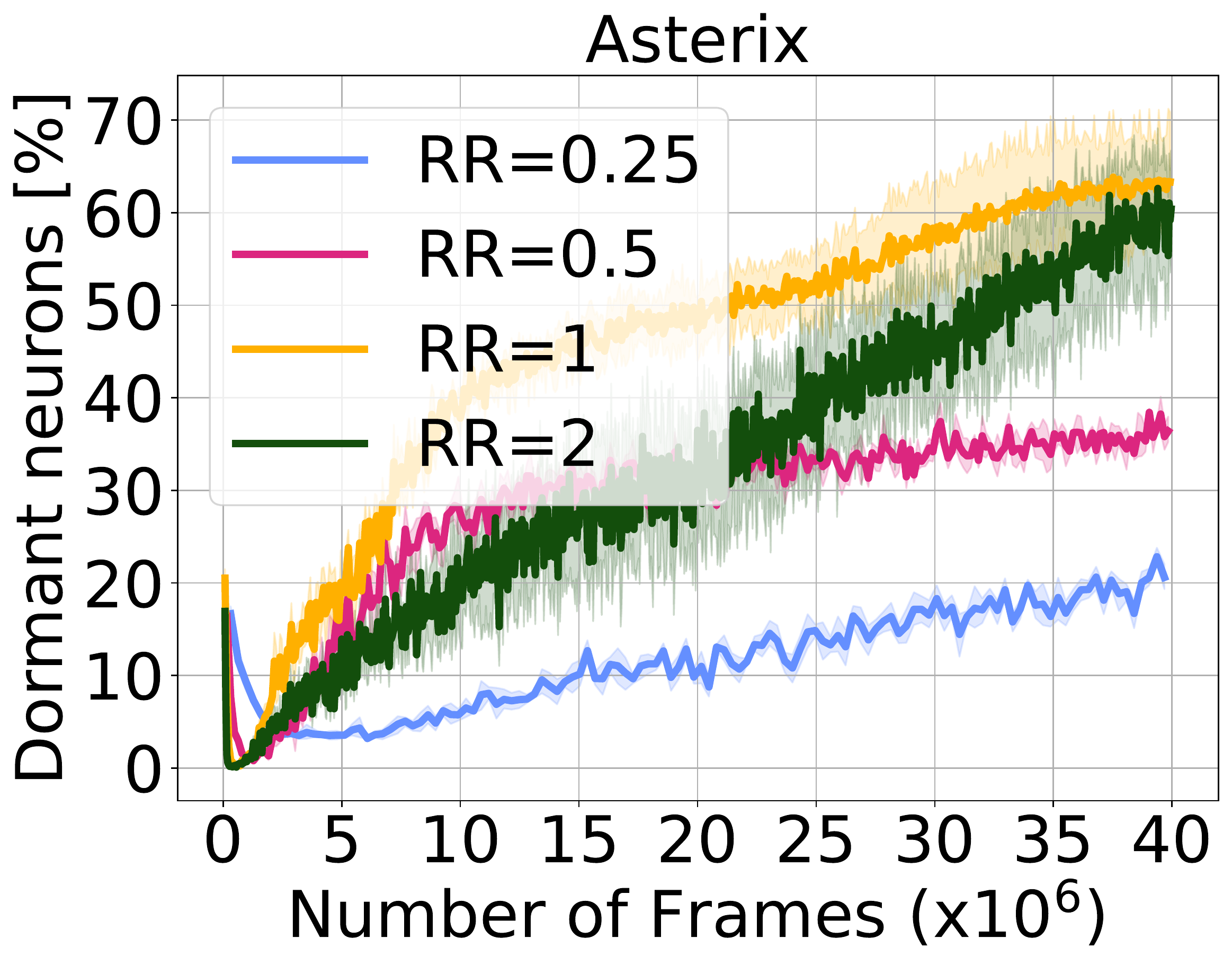}
\includegraphics[width=0.19\columnwidth]{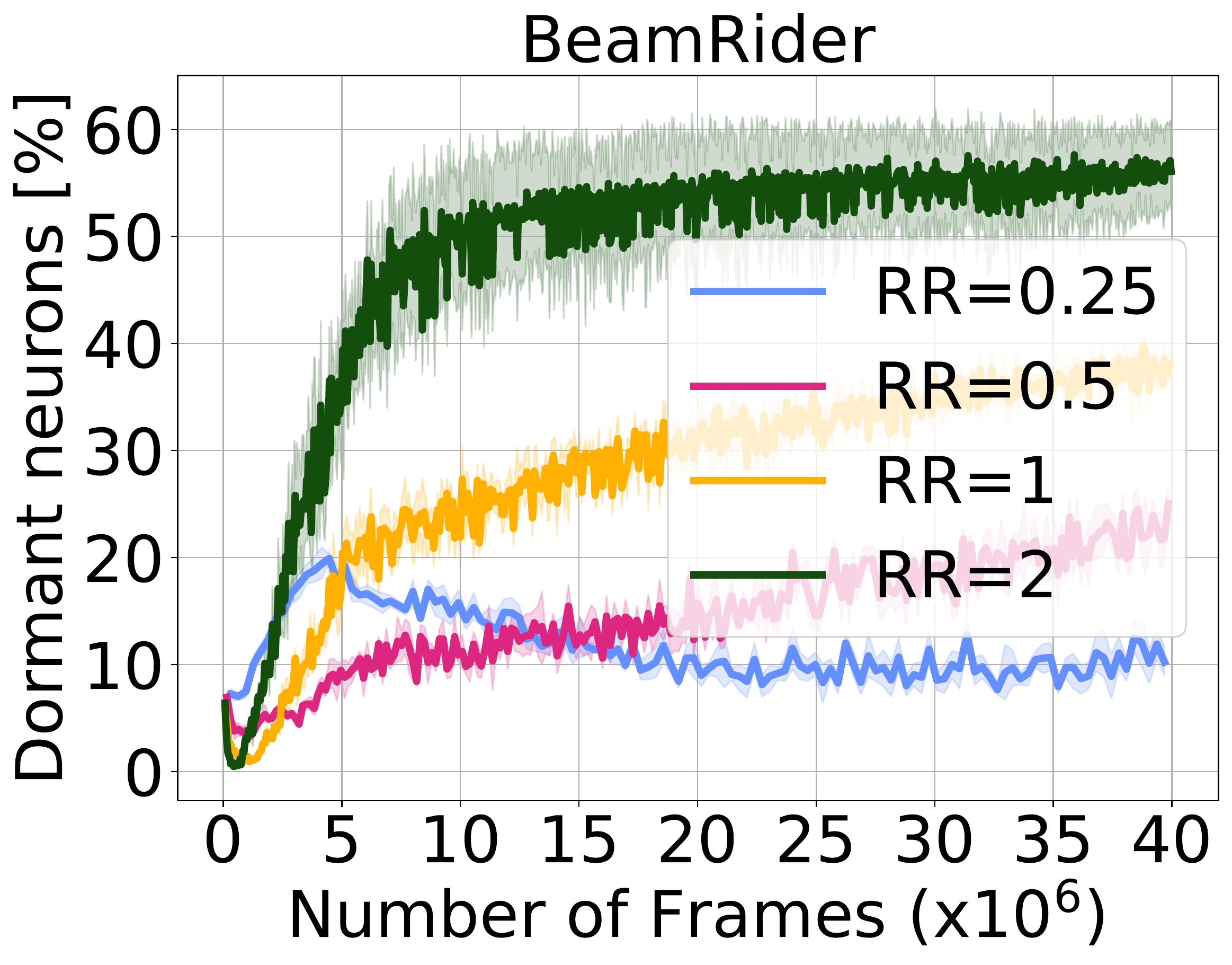}
\includegraphics[width=0.19\columnwidth]{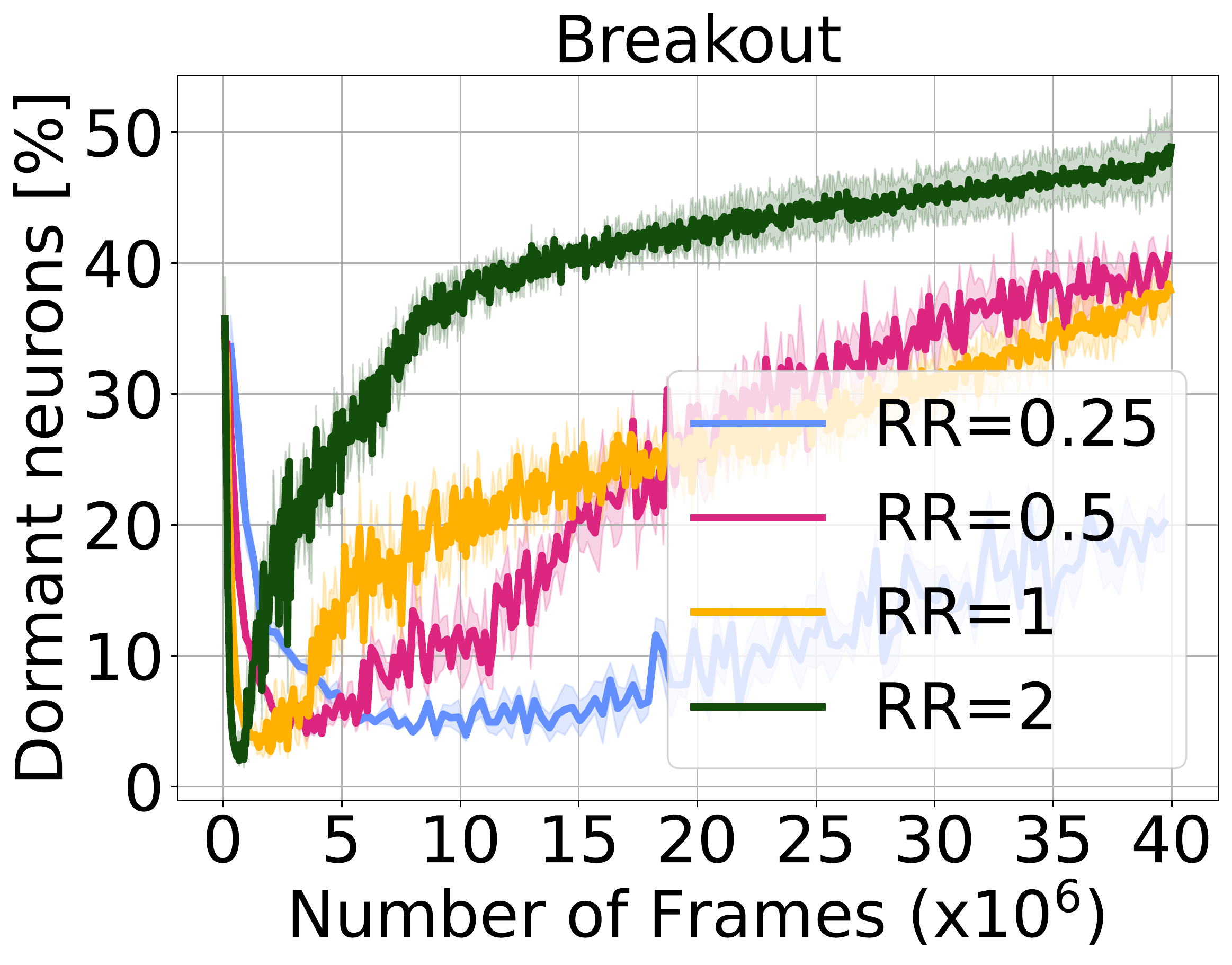}
\includegraphics[width=0.19\columnwidth]{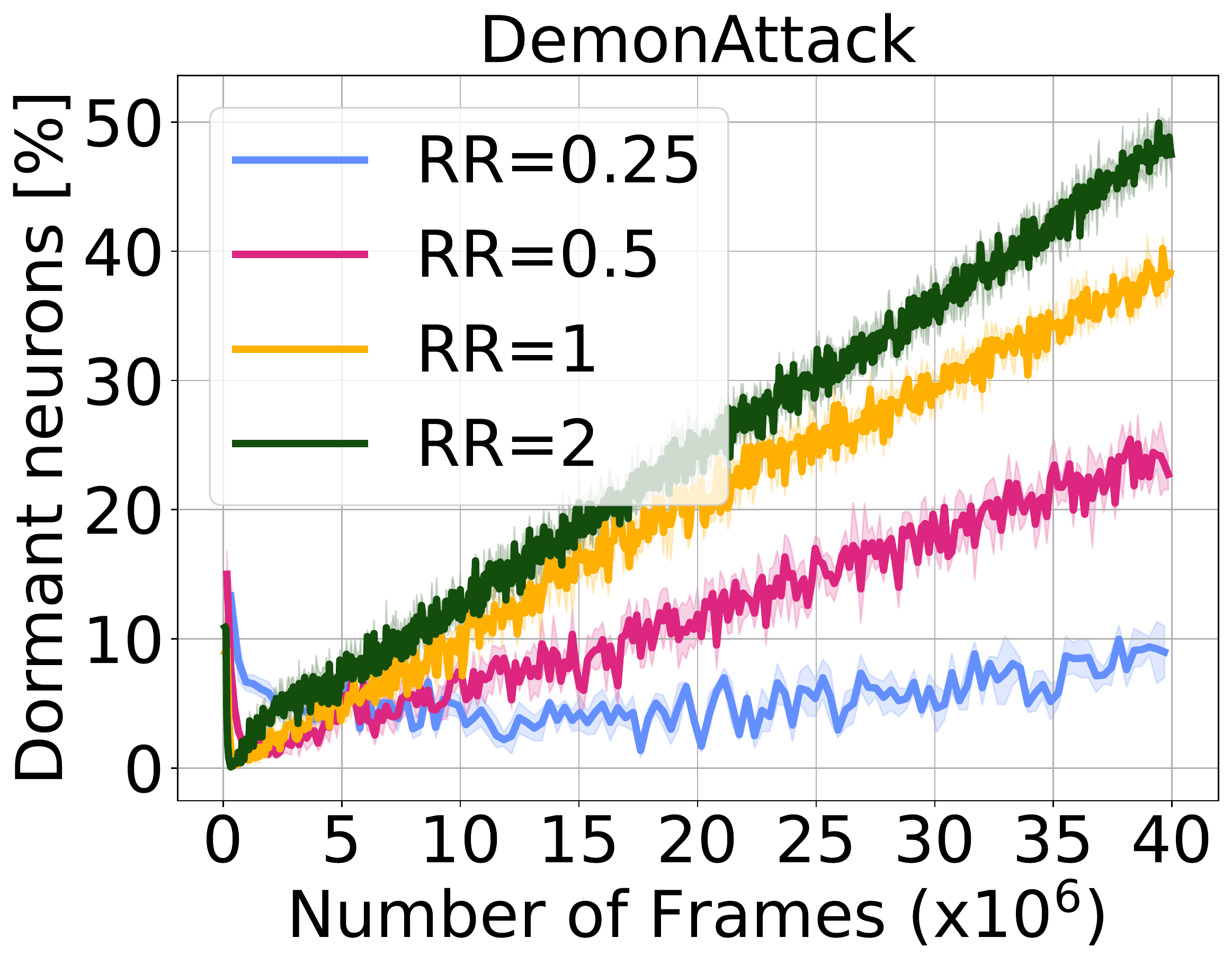}
\includegraphics[width=0.19\columnwidth]{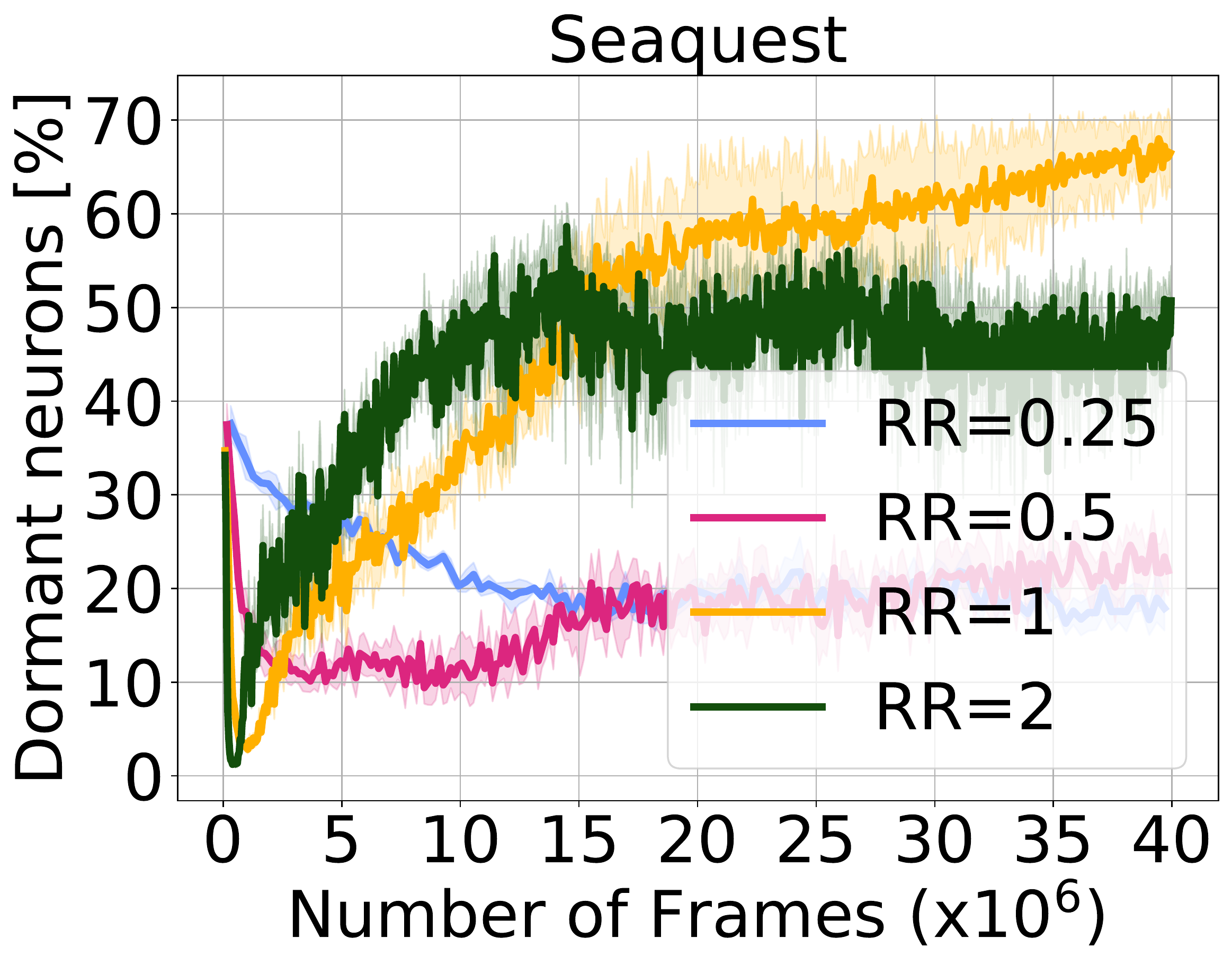}
\includegraphics[width=0.19\columnwidth]{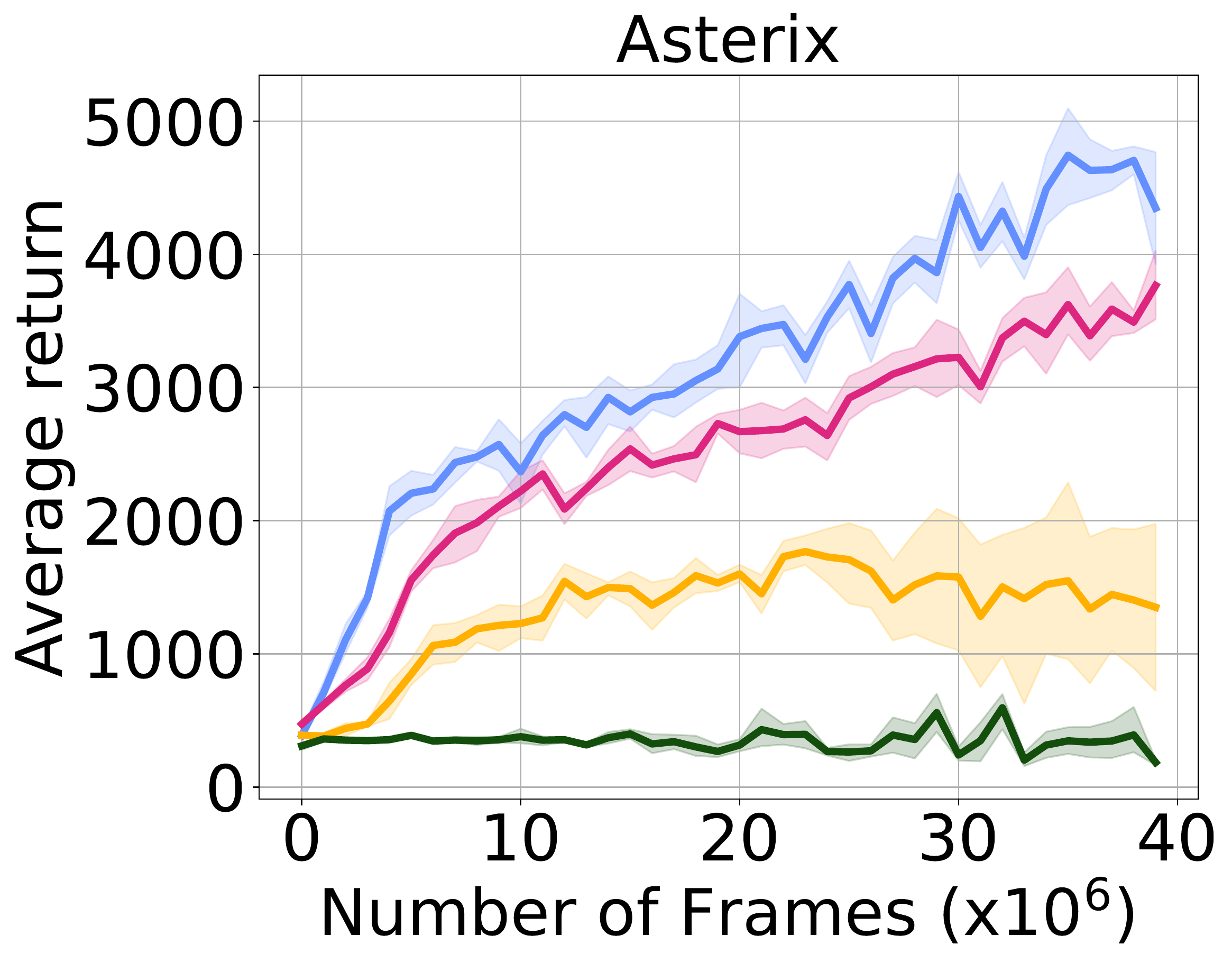} 
\includegraphics[width=0.19\columnwidth]{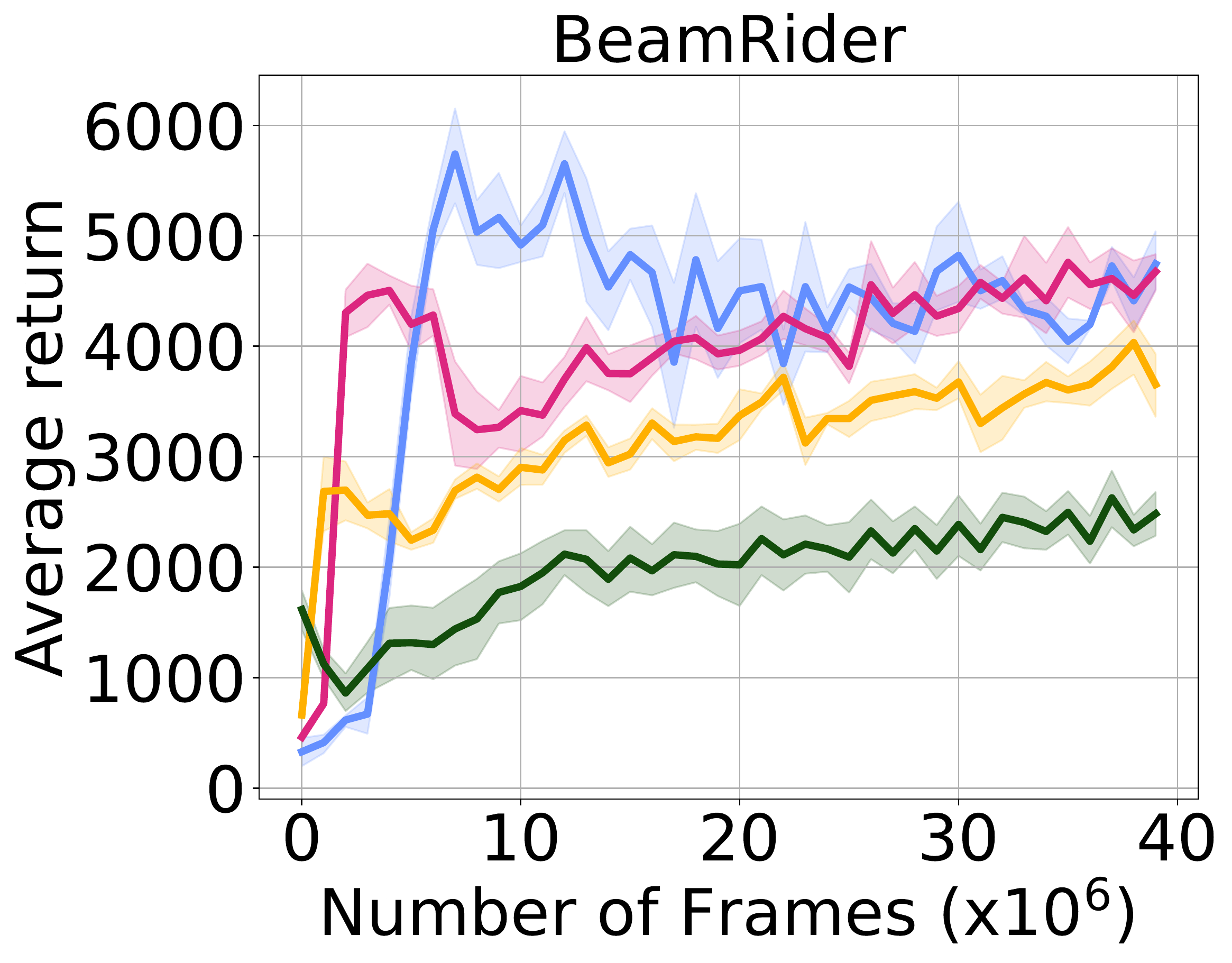} 
\includegraphics[width=0.19\columnwidth]{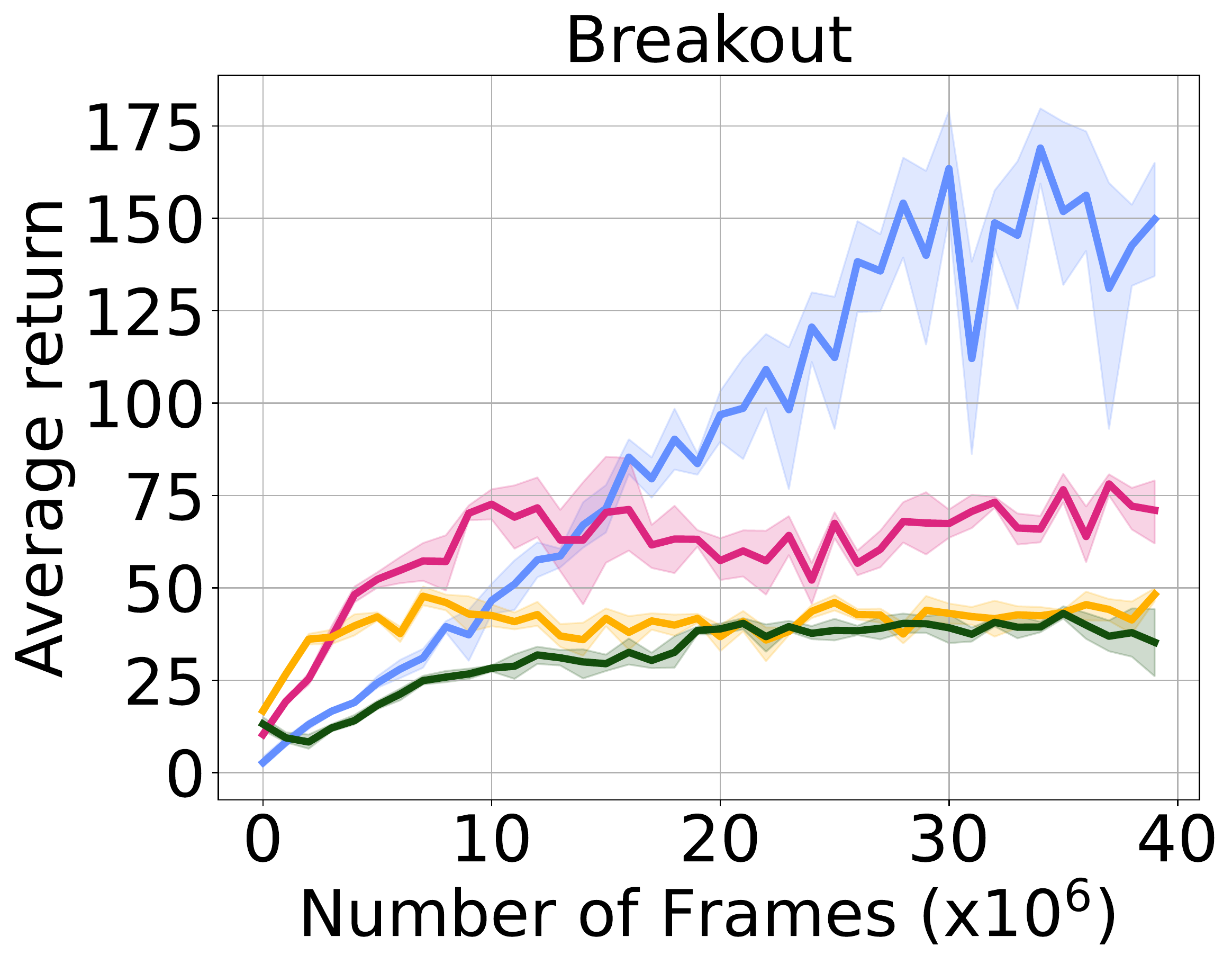} 
\includegraphics[width=0.19\columnwidth]{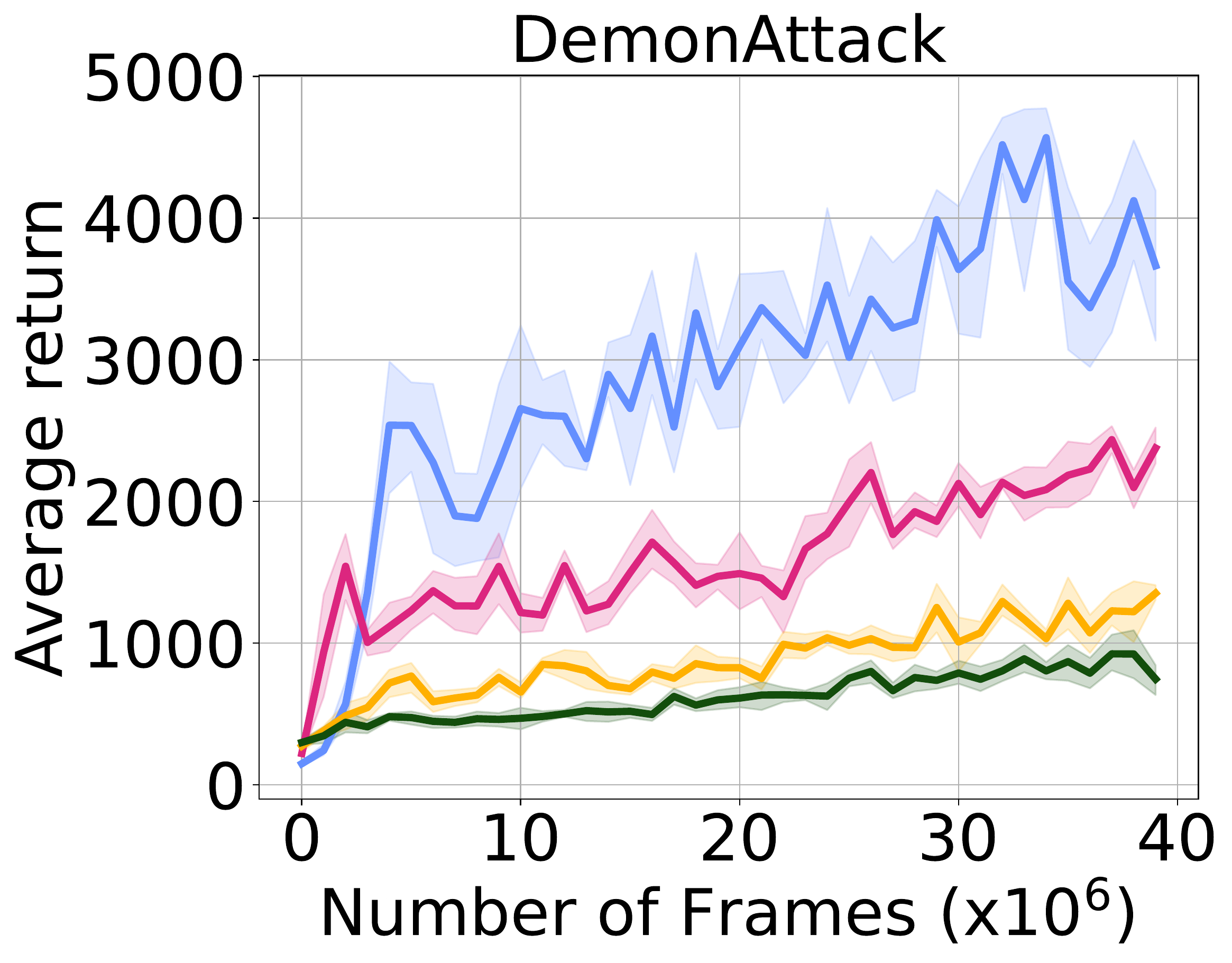}
\includegraphics[width=0.19\columnwidth]{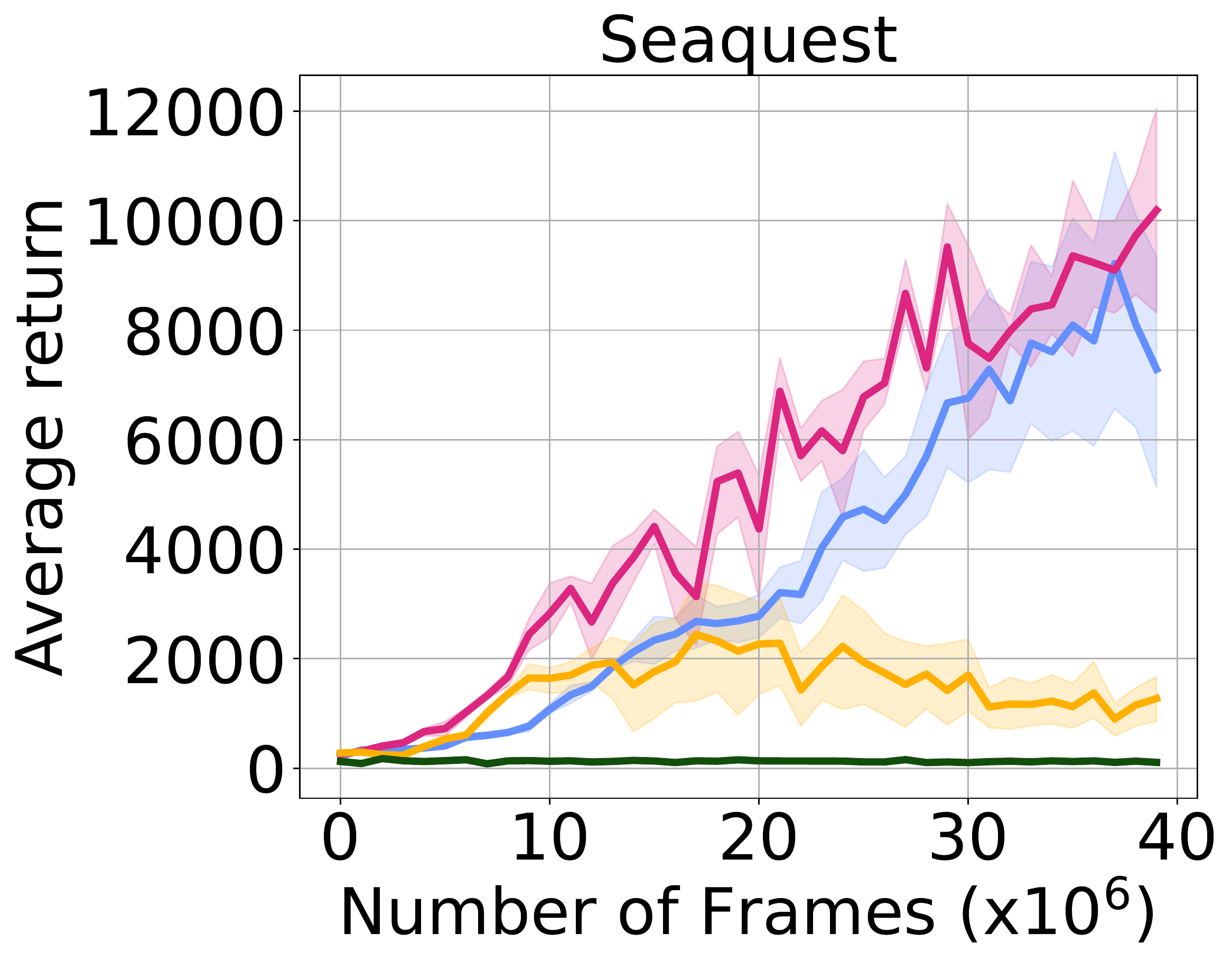}
}
\caption{Effect of replay ratio in the number of dormant neurons for DQN on Atari environments  (experiments presented in \autoref{fig:dormant_in_high_RR}).}
\label{appendix:fig:rr_sweep}
\end{center}
\vskip -0.2in
\end{figure*}

\section{The Dormant Neuron Phenomenon in Different Domains}
\label{appendix:dormant_phenomenon_more_domains}
In this appendix, we demonstrate the dormant neuron phenomenon on DrQ($\epsilon$) \cite{yarats2021image} on the Atari 100K benchmark \cite{kaiser2019model} as well as on additional games from the Arcade Learning Environment on DQN. Additionally, we show the phenomenon on continuous control tasks and analyze the role of dormant neurons in performance. We consider SAC \cite{haarnoja2018soft} trained on MuJoCo environments \cite{todorov2012mujoco}. Same as our analyses in Section \ref{sec:analysis}, we consider $\tau = 0$ to illustrate the phenomenon.  

\begin{figure*}
\vskip 0.2in
\begin{center}
{
\includegraphics[width=0.19\columnwidth]{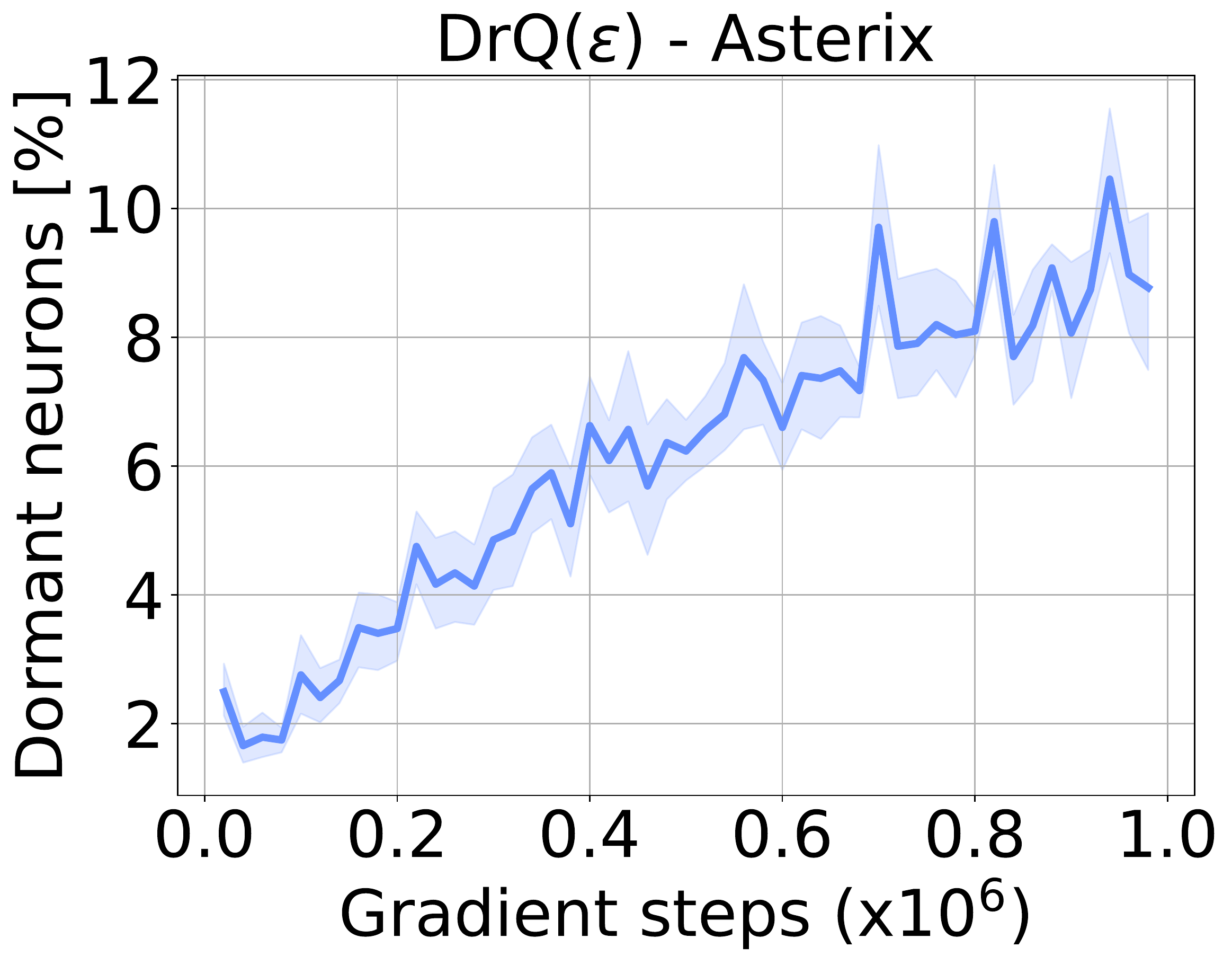}
\includegraphics[width=0.19\columnwidth]{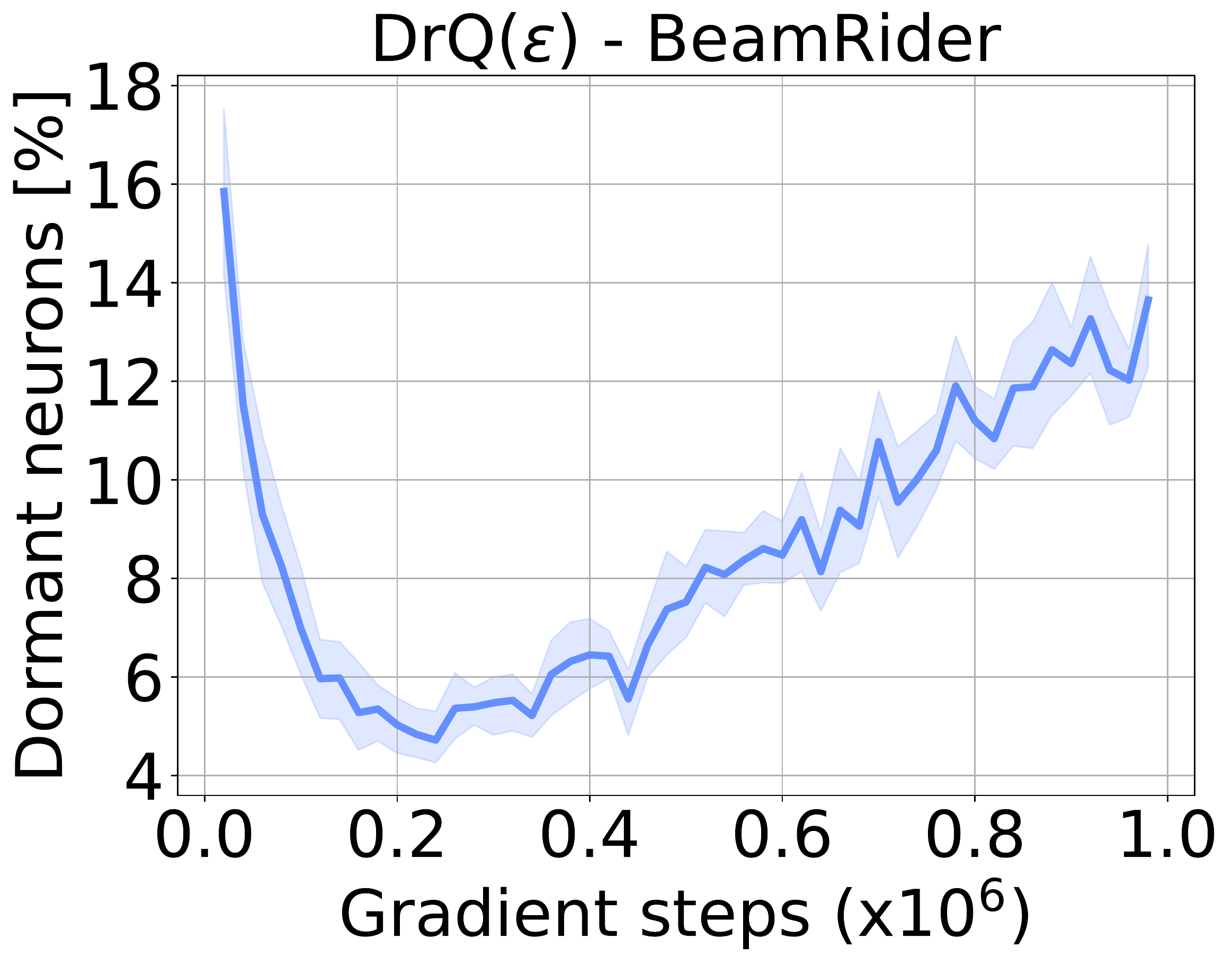}
\includegraphics[width=0.19\columnwidth]{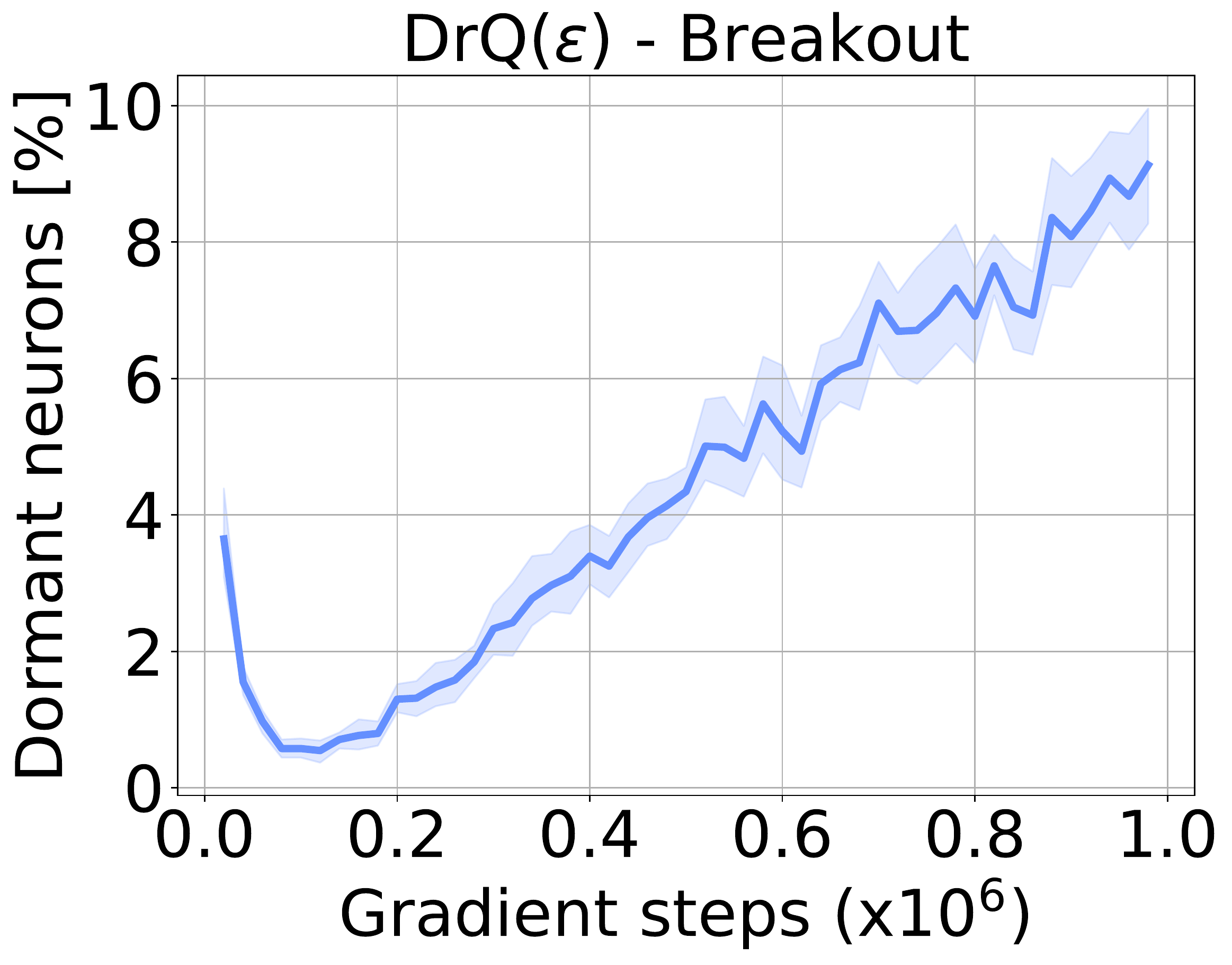}
\includegraphics[width=0.19\columnwidth]{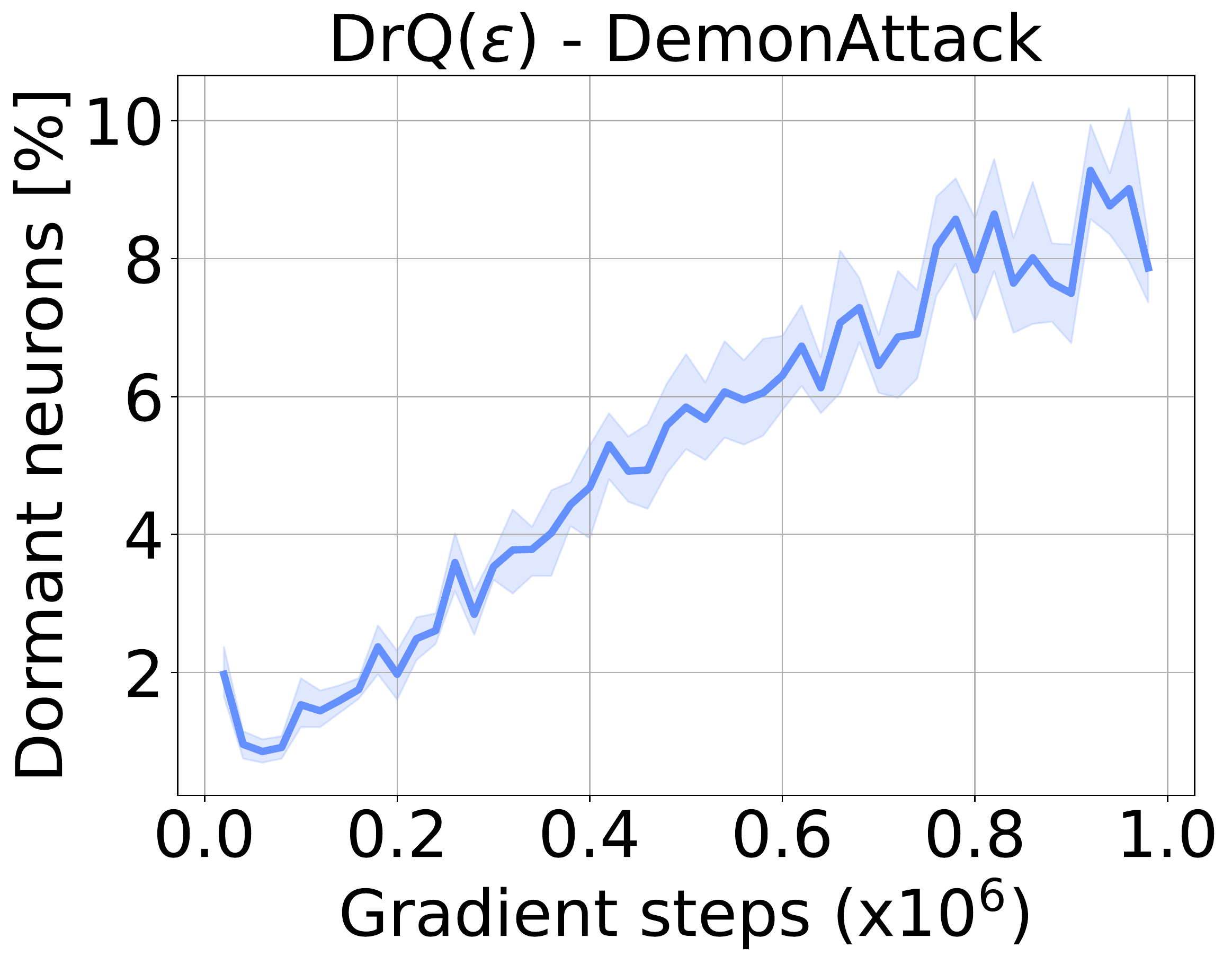} 
\includegraphics[width=0.19\columnwidth]{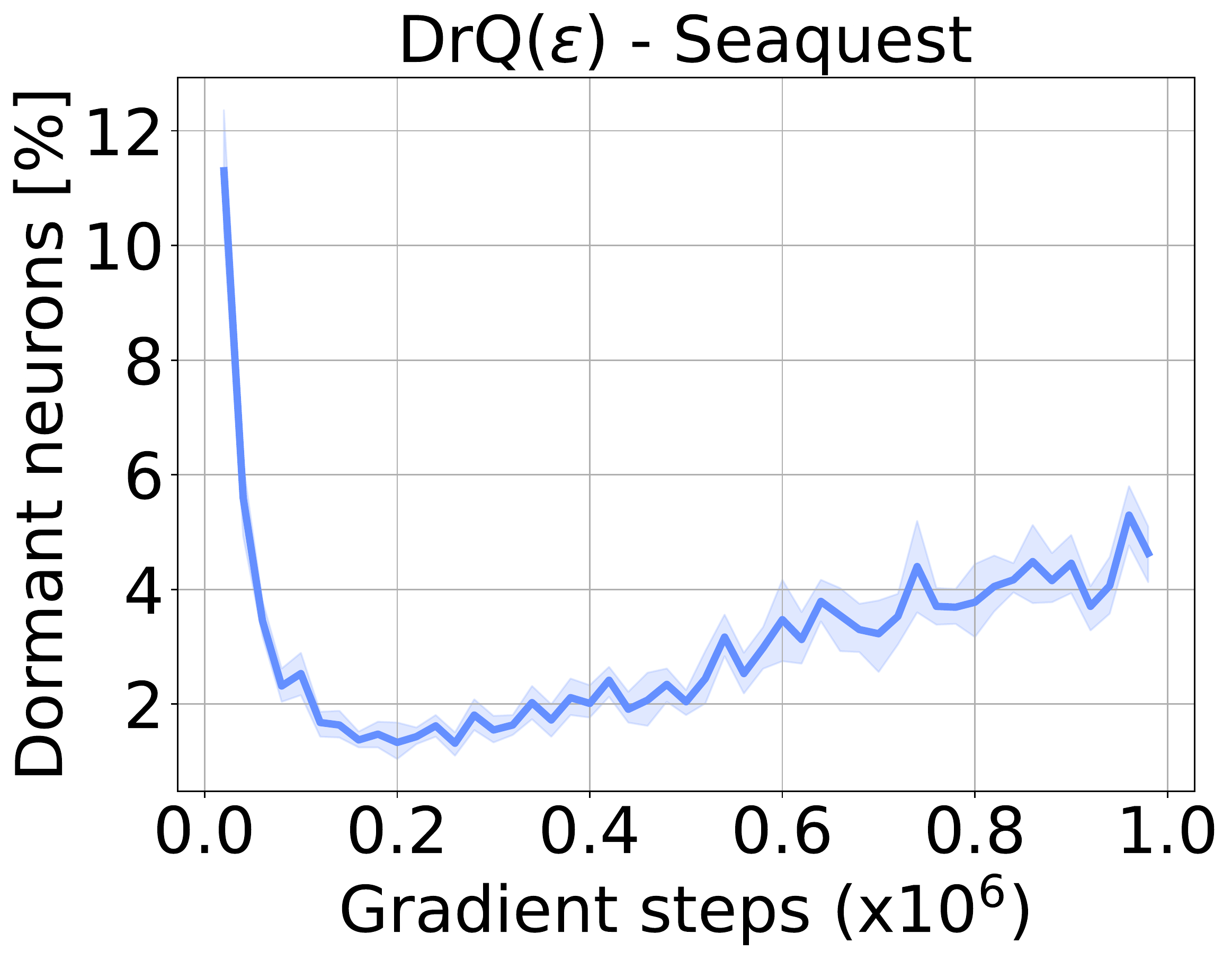}
}
\caption{The dormant neuron phenomenon becomes apparent as the number of training steps increases during the training of DrQ($\epsilon$) with the default replay ratio on Atrai 100K.}
\label{appendix:fig:drq_standard}
\end{center}
\vskip -0.2in
\end{figure*}

\begin{figure}[ht]
\vskip 0.2in
\begin{center}
{
\includegraphics[width=0.22\columnwidth]{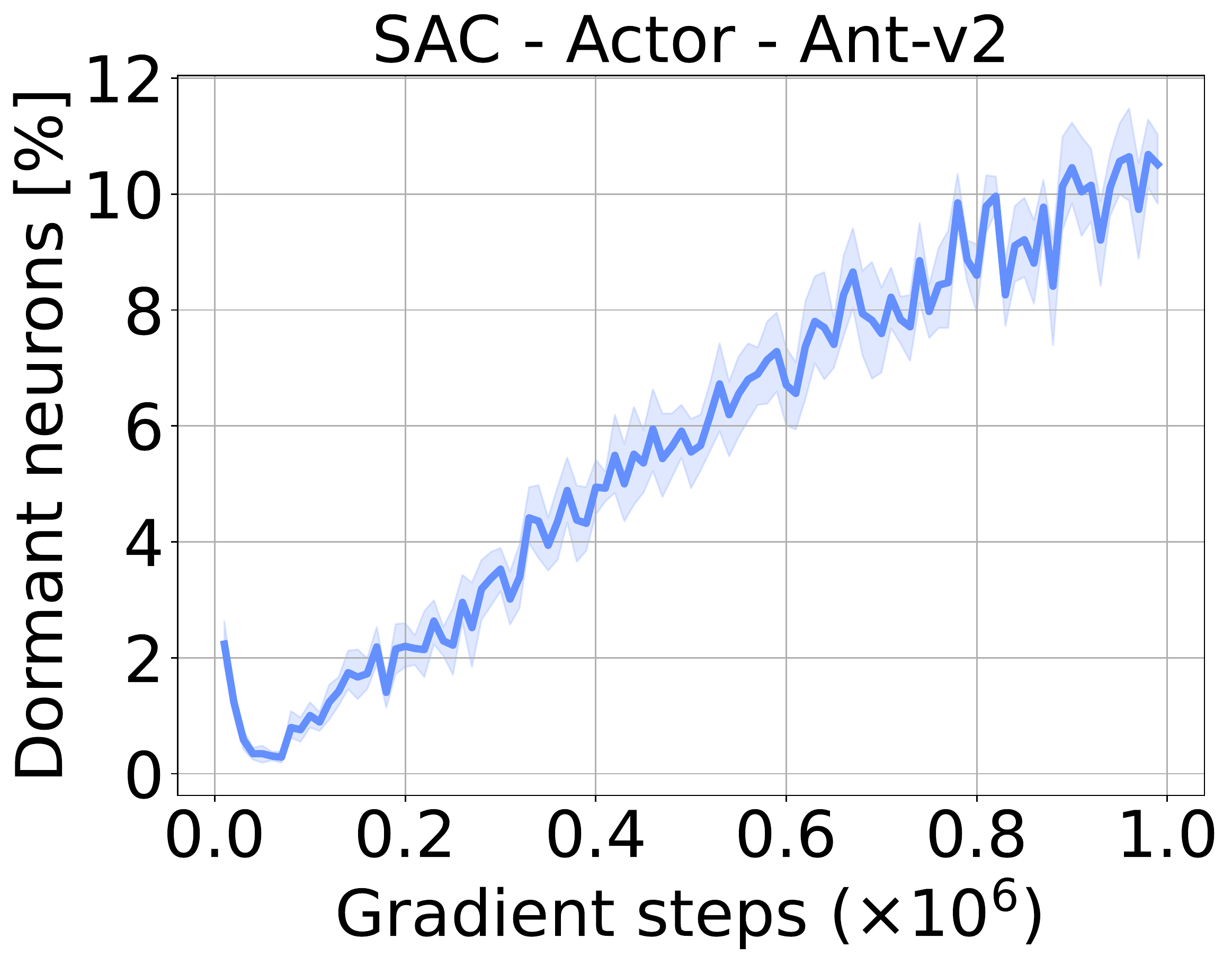}
\includegraphics[width=0.22\columnwidth]{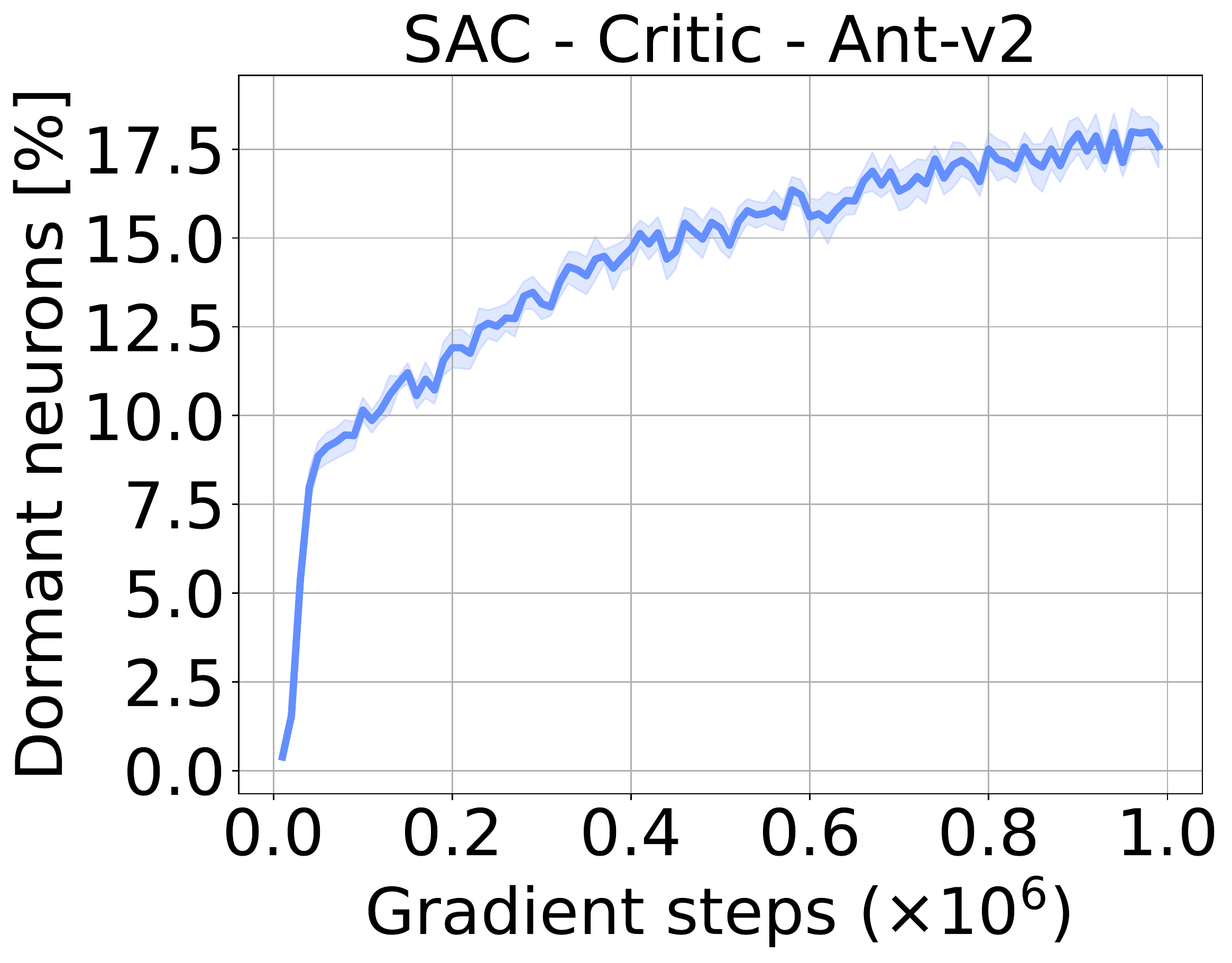}
\hspace{0.1cm}
\includegraphics[width=0.22\columnwidth]{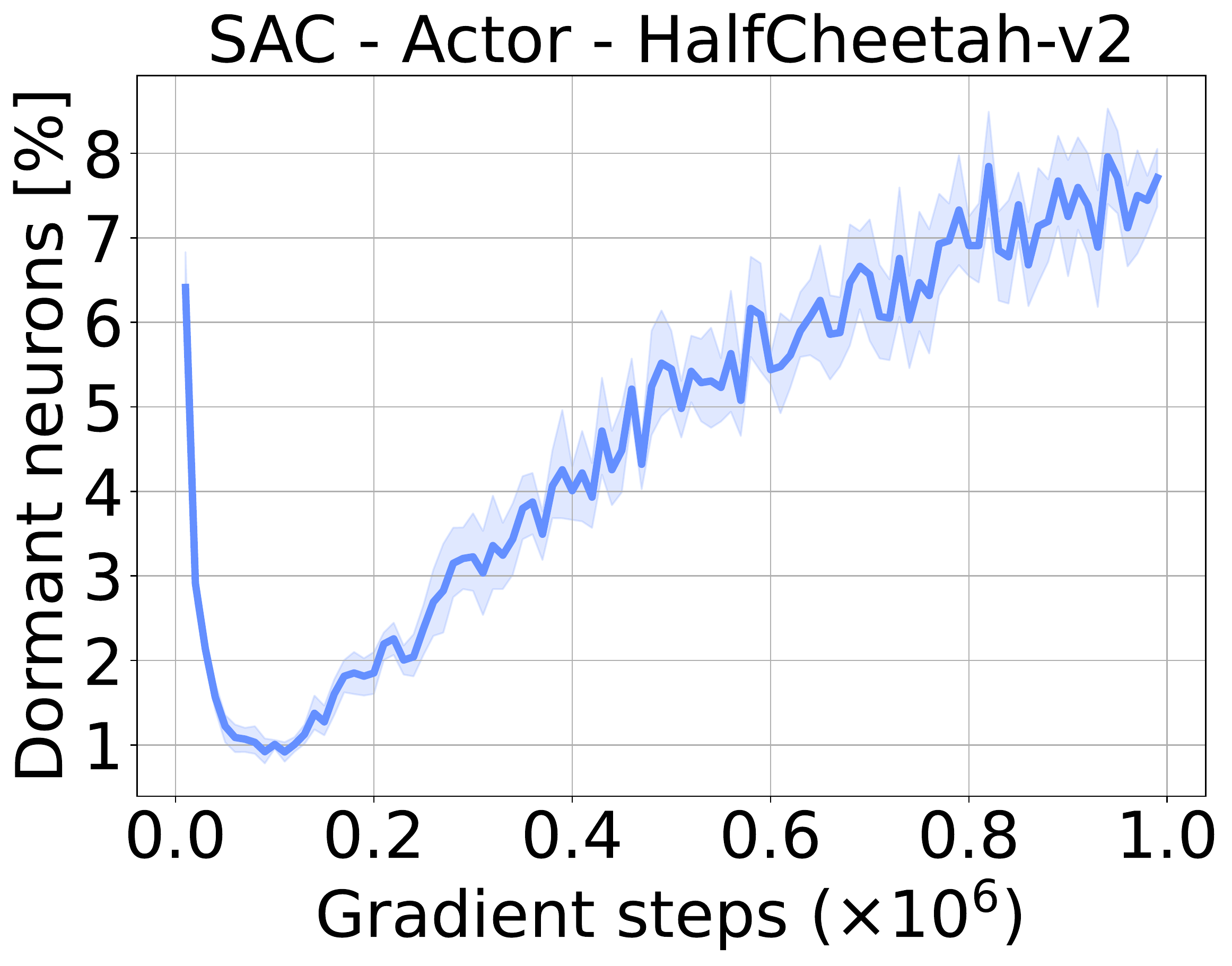}
\includegraphics[width=0.22\columnwidth]{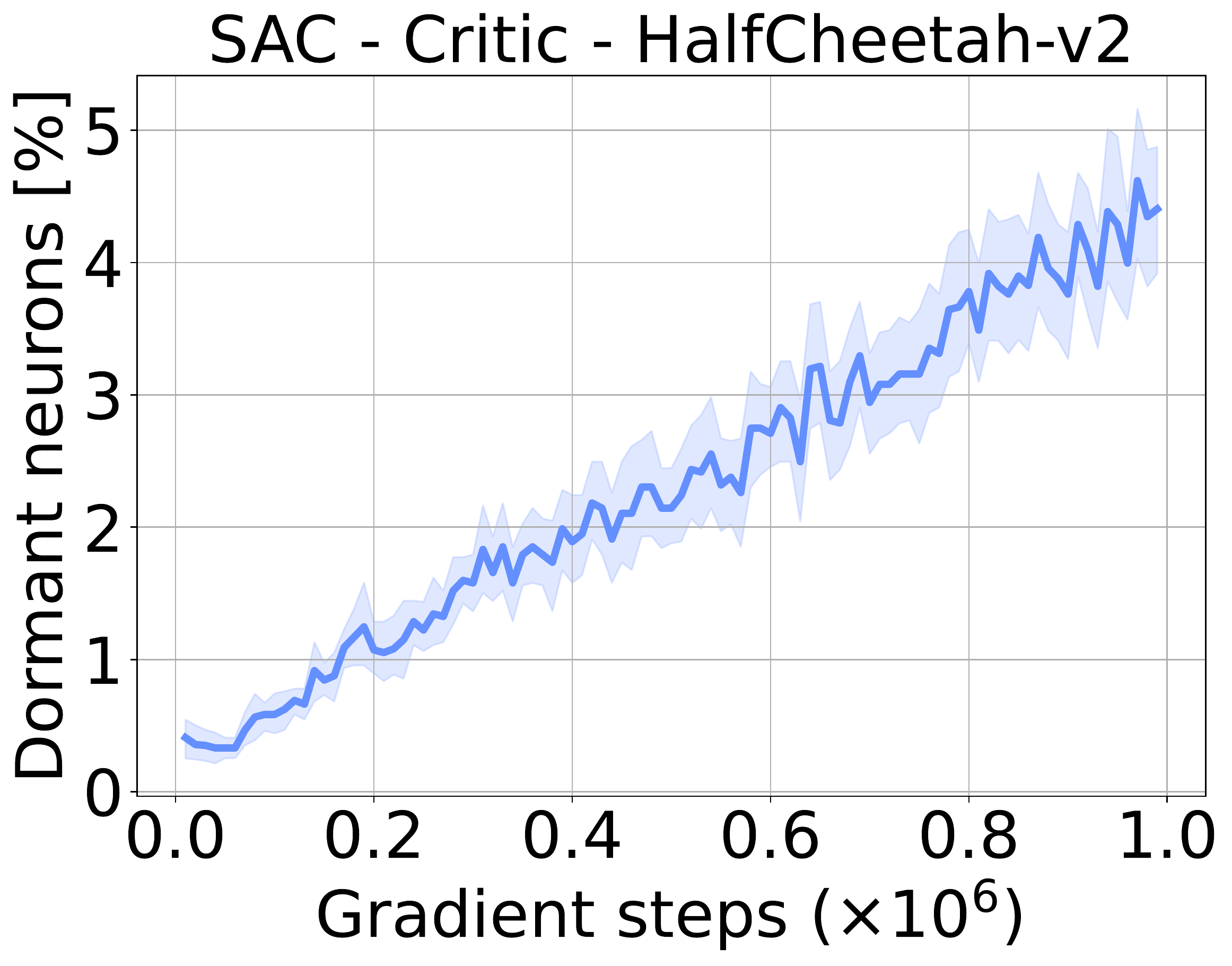}
}
\caption{The number of dormant neurons increases over time during the training of SAC on MuJoCo environments. }
\label{fig:dormant_neurons_mujoco}
\end{center}
\vskip -0.2in
\end{figure}

\begin{figure}[b]
\vskip 0.2in
\begin{center}
{
\includegraphics[width=0.27\columnwidth]{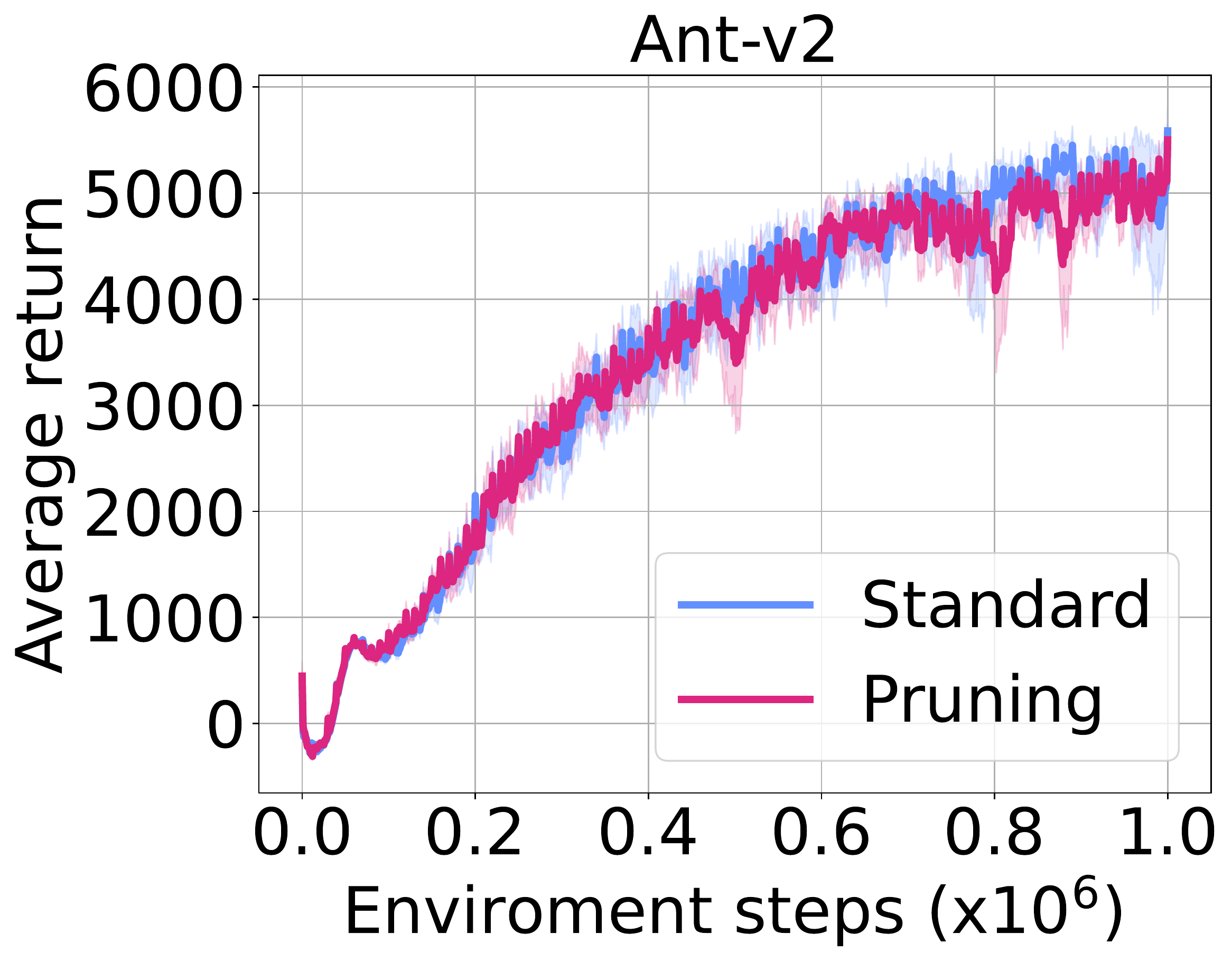}
\hspace{0.5cm}
\includegraphics[width=0.27\columnwidth]{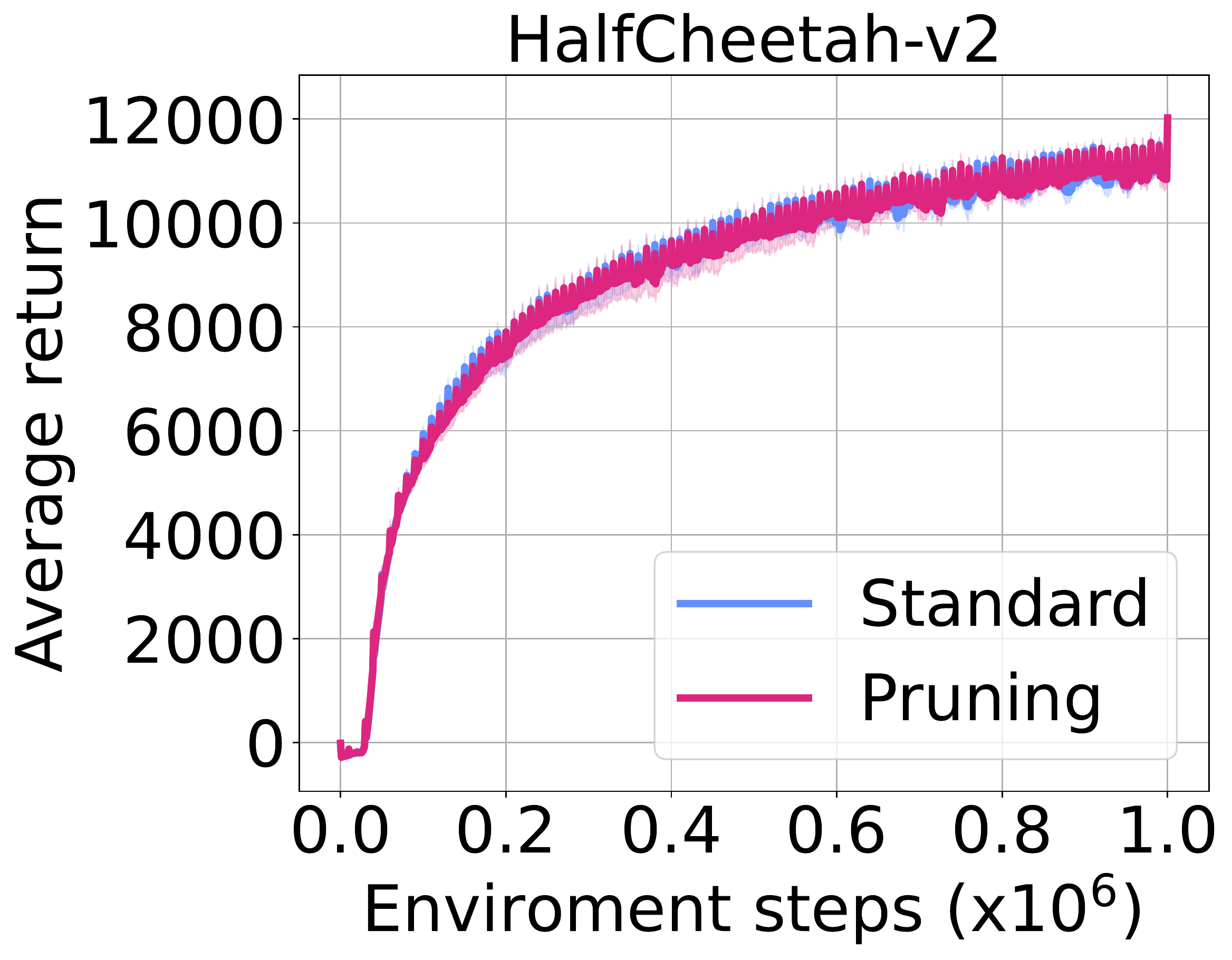}
}
\caption{Pruning dormant neurons during the training of SAC on MuJoCo environments does not affect the performance.}
\label{fig:pruning_mujoco}
\end{center}
\vskip -0.2in
\end{figure}

\begin{figure}[ht]
\vskip 0.2in
\begin{center}
{
\includegraphics[width=0.22\columnwidth]{figures/results/AverageReturn_Ant-v2_Mujoco_standard.pdf}
\includegraphics[width=0.22\columnwidth]{figures/results/AverageReturn_HalfCheetah-v2_Mujoco_standard.pdf}
\includegraphics[width=0.22\columnwidth]{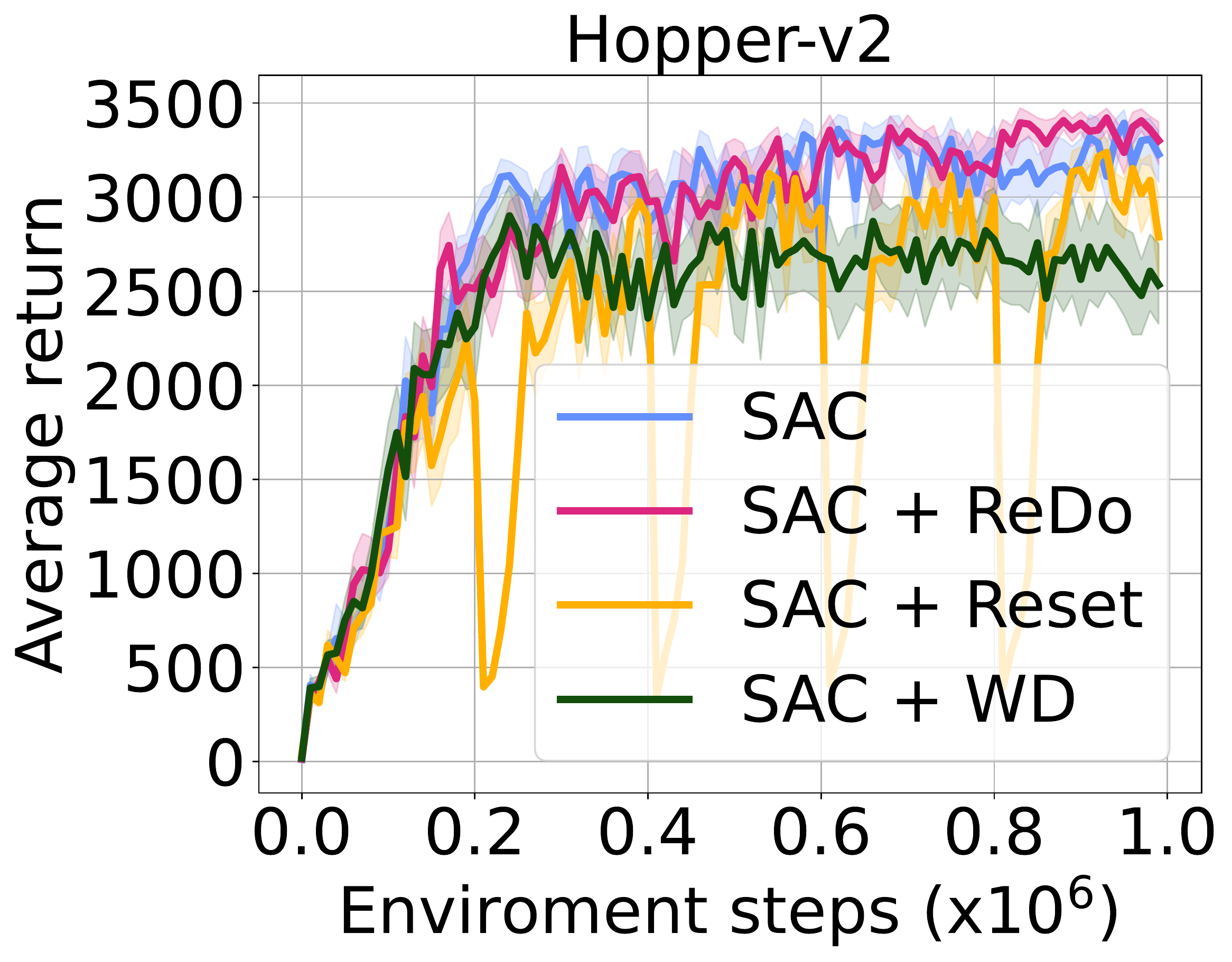}
\includegraphics[width=0.22\columnwidth]{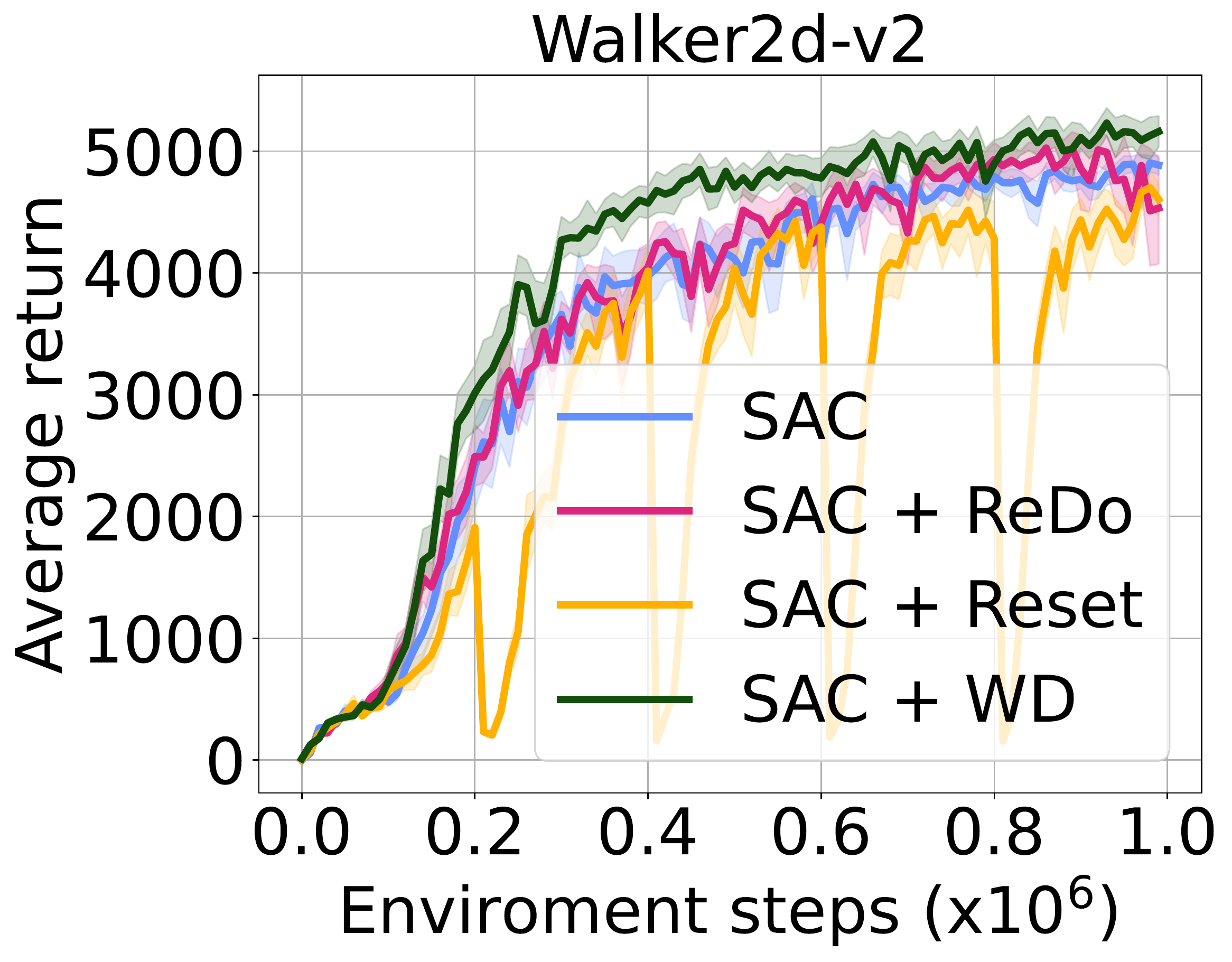}
}
\caption{Comparison of the performance of SAC agents with {\em ReDo} and two different regularization methods.}
\label{fig:appendix:redo_reset_mujoco}
\end{center}
\vskip -0.2in
\end{figure}

\autoref{appendix:fig:rr_sweep} shows that across games, the number of dormant neurons consistently increases with higher values for the replay ratio on DQN. The increase in dormant neurons correlates with the performance drop observed in this regime. We then investigate the phenomenon on a modern valued-based algorithm DrQ($\epsilon$). As we see in \autoref{appendix:fig:drq_standard}, the phenomenon emerges as the number of training steps increases.  

\autoref{fig:dormant_neurons_mujoco} shows that the phenomenon is also present in continuous control tasks. An agent exhibits an increasing number of dormant neurons in the actor and critic networks during the training of SAC on MuJoco environments. To analyze the effect of these neurons on performance, we prune dormant neurons every 200K steps. \autoref{fig:pruning_mujoco} shows that the performance is not affected by pruning these neurons; indicating their little contribution to the learning process.      
Next, we investigate the effect of {\em ReDo} and the studied baselines (Reset \cite{nikishin2022primacy} and weight decay (WD)) in this domain. \autoref{fig:appendix:redo_reset_mujoco} shows that {\em ReDo} maintains the performance of the agents while other methods cause a performance drop in most cases. We hypothesize that {\em ReDo} does not provide gains here as the state space is considerably low and the typically used network is sufficiently over-parameterized. 

To investigate this, we decrease the size of the actor and critic networks by halving or quartering the width of their layers. We performed these experiments on the complex environment Ant-v2 using 5 seeds. \autoref{table:SAC_different_width} shows the final average return in each case. We observe that when the network size is smaller, there are some gains from recycling the dormant capacity. Further analyses of the relation between task complexity and network capacity would provide a more comprehensive understanding.     

\begin{table}[t]
\caption{Performance of SAC on Ant-v2 using using half and a quarter of the width of the actor and critic networks.}
\label{table:SAC_different_width}
\vskip 0.15in
\begin{center}
\begin{tabular}{lcc}
\toprule
Width & SAC & SAC+ReDo \\
\midrule
0.25 & 2016.18 $\pm$ 102 & \textbf{2114.52} $\pm$ 212 \\
0.5 &  3964.04 $\pm$ 953 & \textbf{4471.61} $\pm$ 648 \\
\bottomrule
\end{tabular}
\end{center}
\vskip -0.1in
\end{table}

\section{Recycling Dormant Neurons}
\label{appendix:recycling_strategies}
Here we study different strategies for recycling dormant neurons and analyze the design choices of {\em ReDo}. We performed these analyses on DQN agents trained with $RR = 1$ and $\tau = 0.1$ on Atari games. Furthermore, we provide some additional insights into the effect of recycling the dormant capacity on improving the sample efficiency and the expressivity of the network.   

\subsection{Effect of Activation Function}
\label{appendix:effect_activation_fn}
In this section, we attempt to understand the effect of the activation function (ReLU) used in our experiments. The ReLU activation function consists of a linear part (positive domain) with unit gradients and a constant zero part (negative domain) with zero gradients. Once the distribution of pre-activations falls completely into the negative part, it would stay there since the weights of the neuron would get zero gradients. This could be the explanation for the increased number of dormant neurons in our neural networks. If this is the case, one might expect activations with non-zero gradients on the negative side, such as leaky ReLU, to have significantly fewer dormant neurons. 

\begin{figure}[ht]
\vskip 0.2in
\begin{center}
{
\includegraphics[width=0.3\columnwidth]{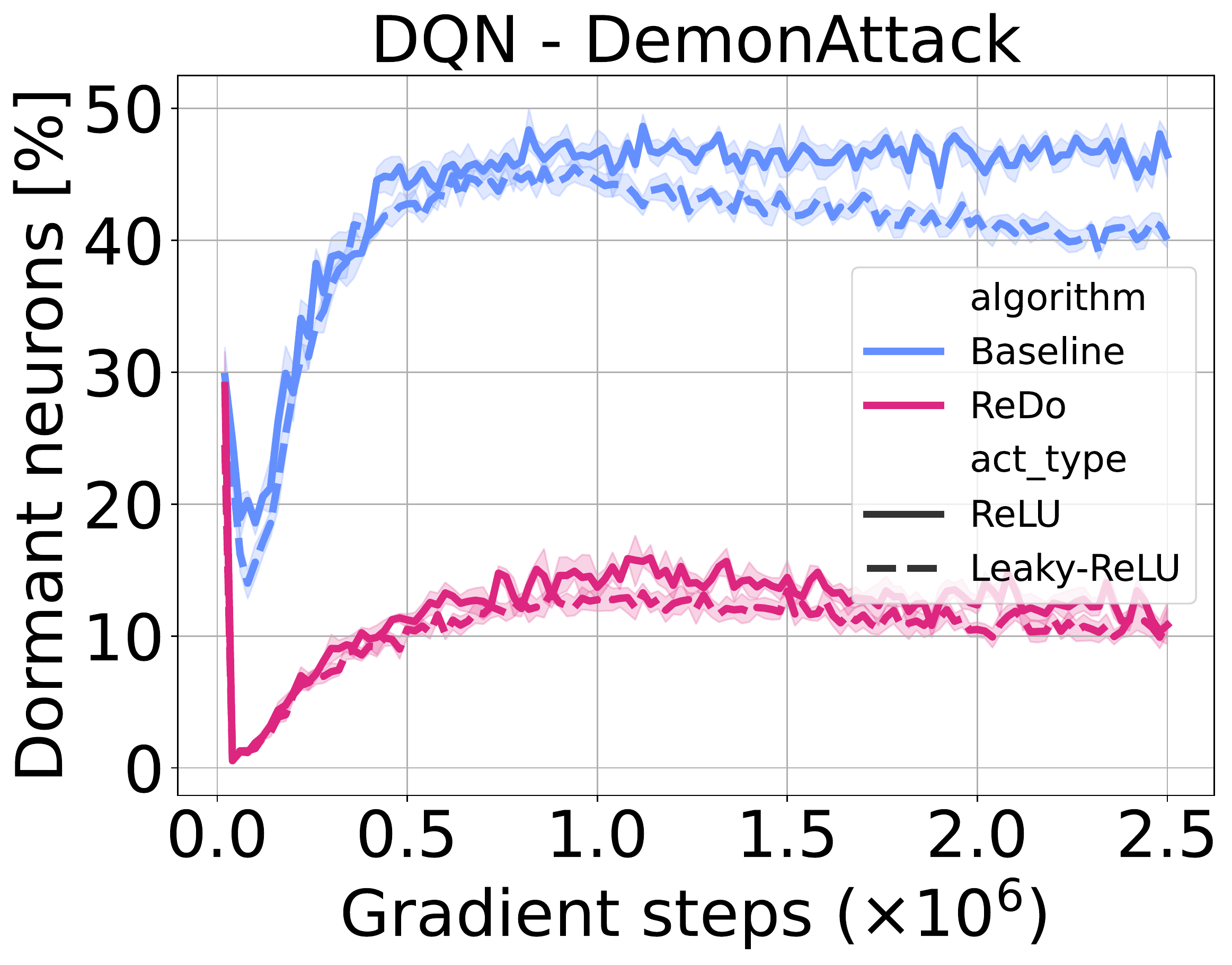}
\includegraphics[width=0.3\columnwidth]{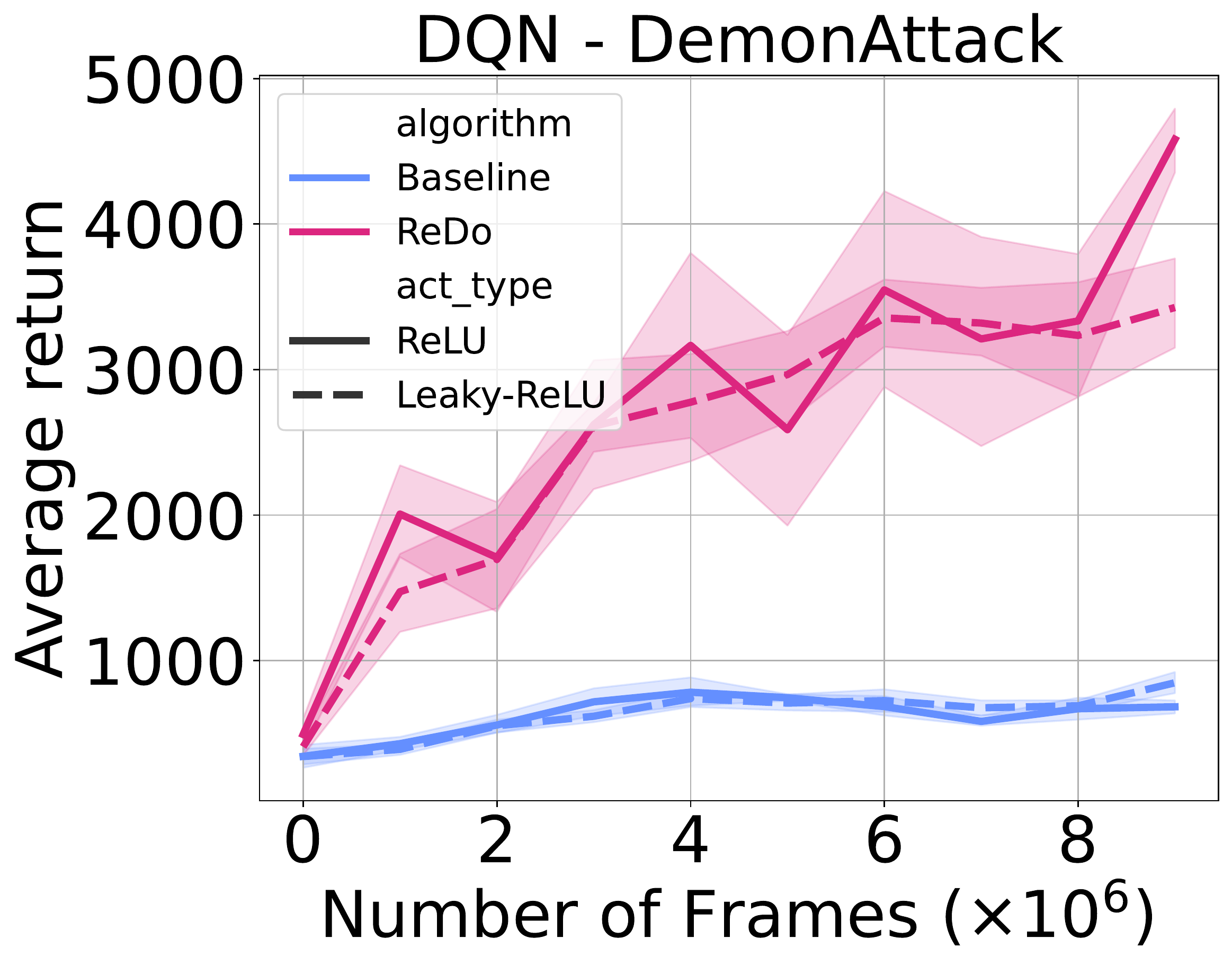}
}
\caption{Training performance and dormant neuron characteristics of networks using leaky ReLU with a negative slope of 0.01 (default value) compared to original networks with ReLU.}
\label{appendix:lrelu}
\end{center}
\vskip -0.2in
\end{figure}

In \autoref{appendix:lrelu}, we compare networks with leaky ReLU to original networks with ReLU activation. As we can see, using leaky ReLU slightly decreases the number of dormant neurons but does not mitigate the issue. {\em ReDo} overcomes the performance drop that occurs during training in the two cases. 

\begin{figure*}
\vskip 0.2in
\begin{center}
{
\includegraphics[width=0.19\columnwidth]{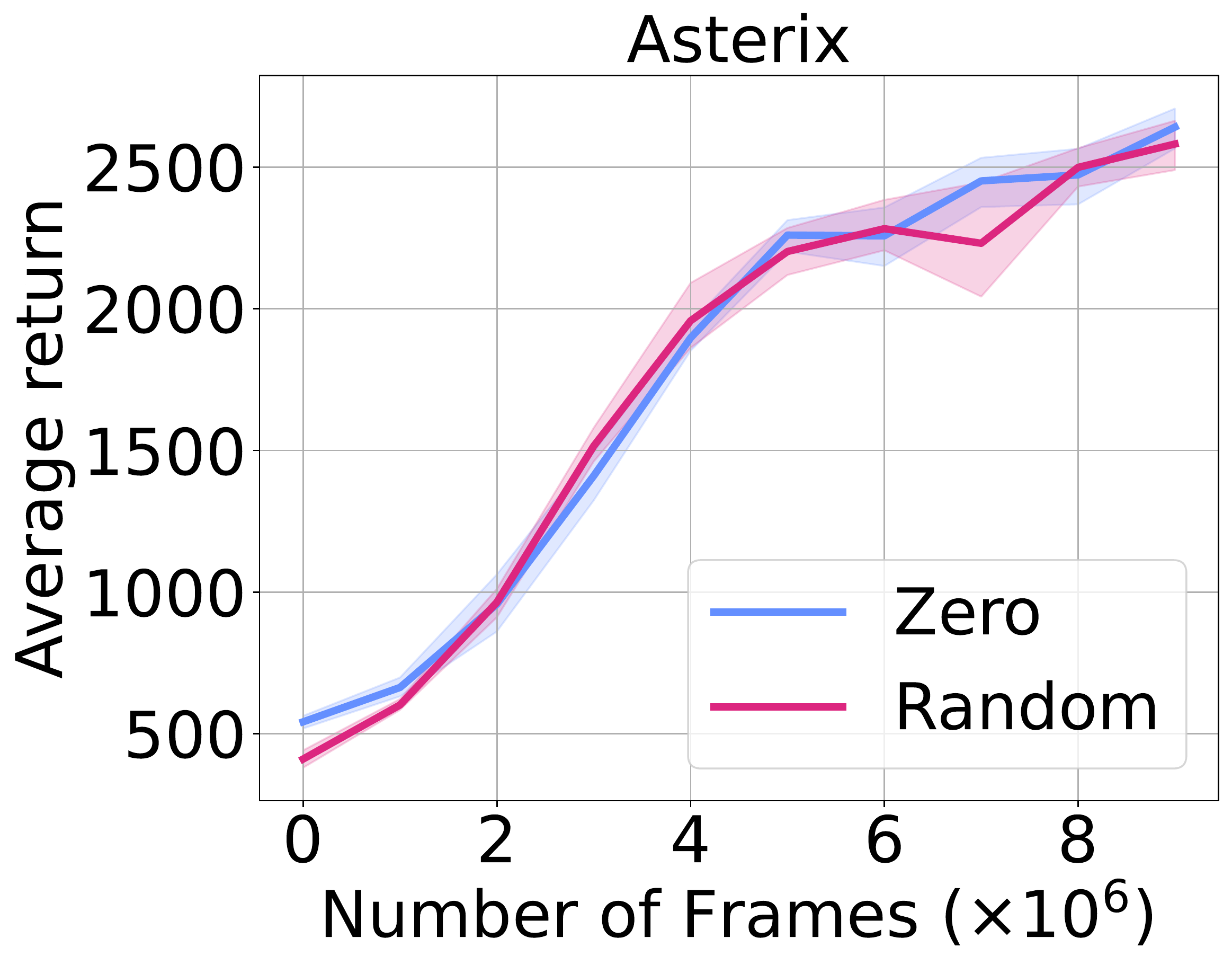}
\includegraphics[width=0.19\columnwidth]{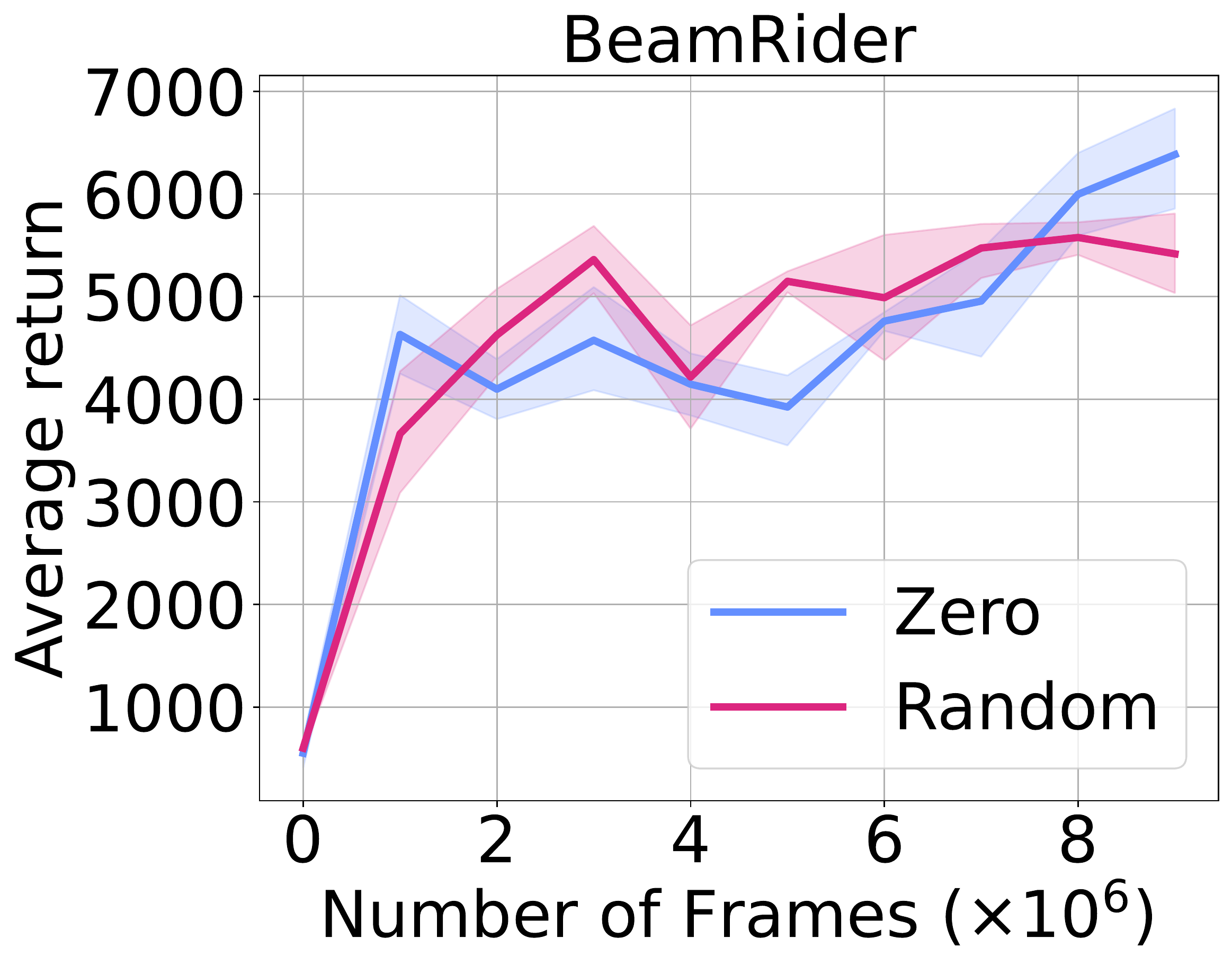}
\includegraphics[width=0.19\columnwidth]{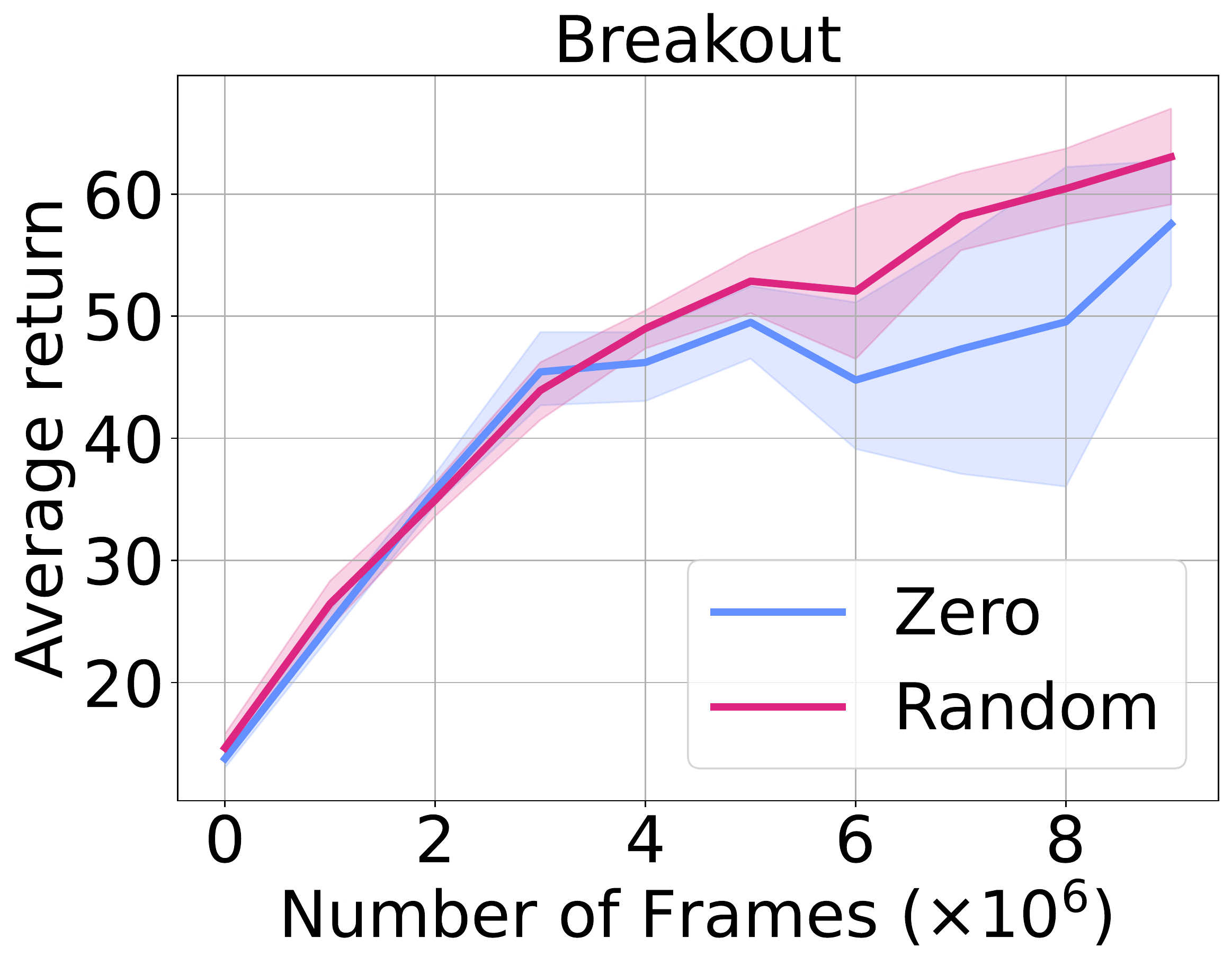}
\includegraphics[width=0.19\columnwidth]{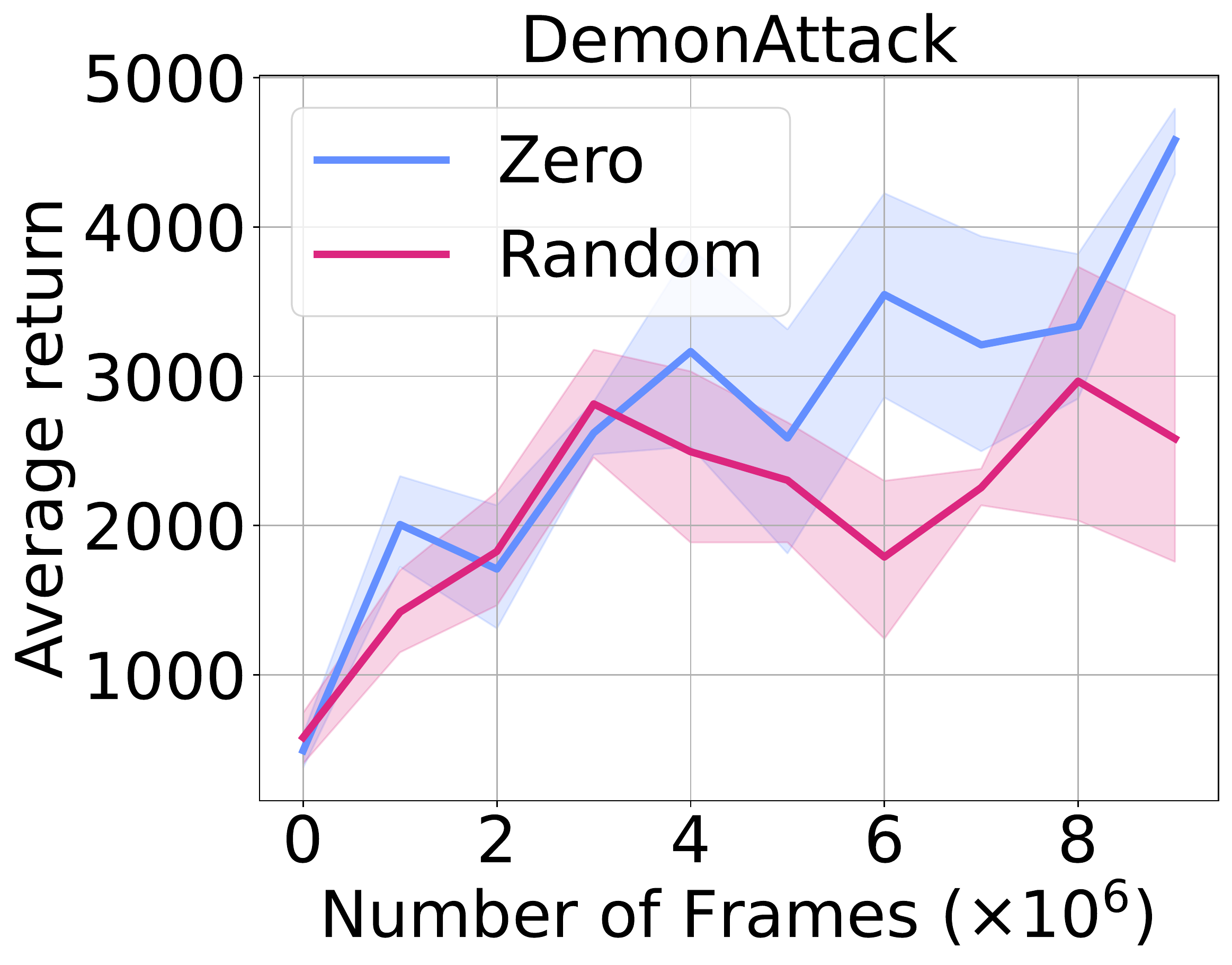}
\includegraphics[width=0.19\columnwidth]{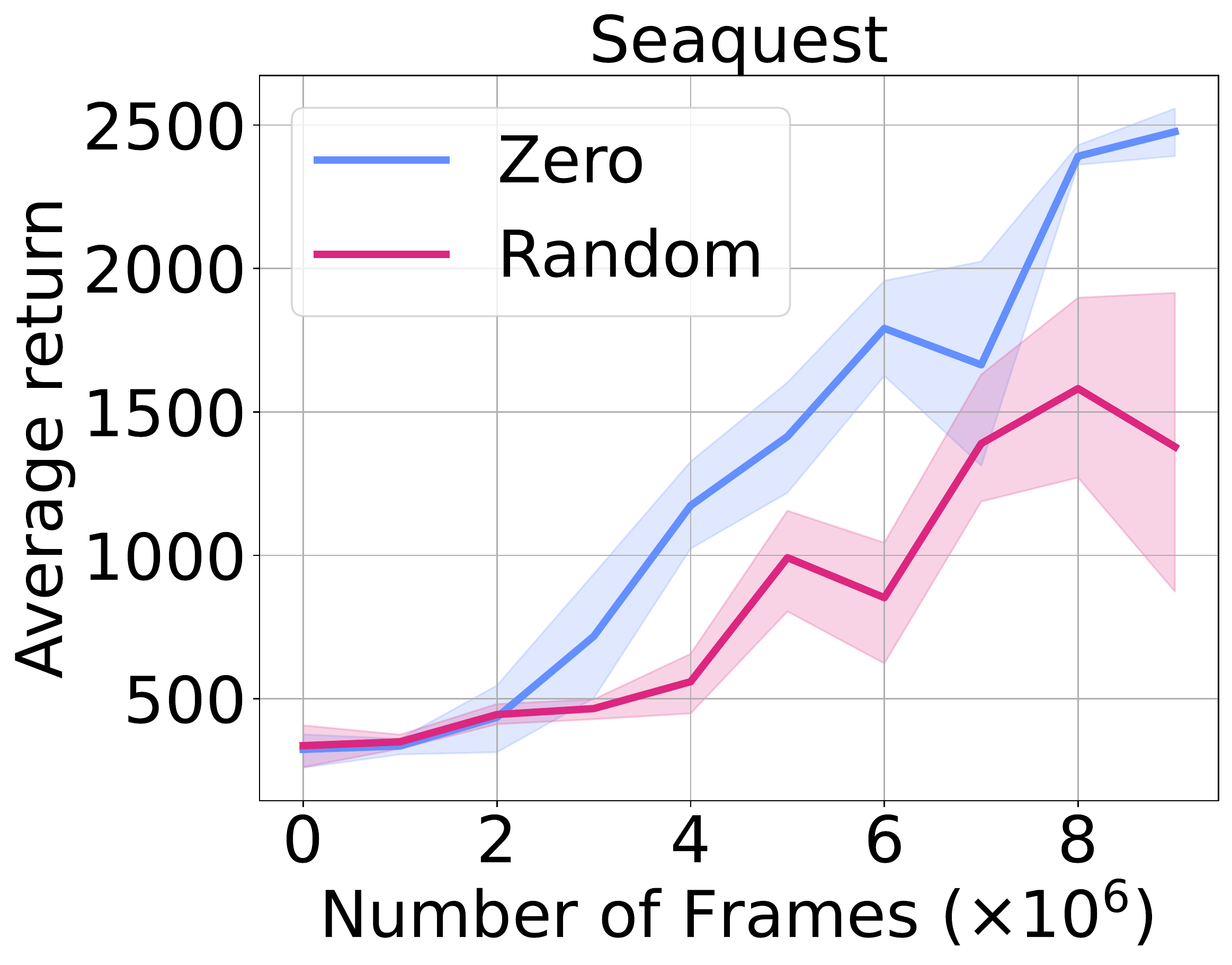} 

}
\caption{Comparison of performance with different strategies of reinitializing the outgoing connections of dormant neurons.}
\label{fig:neuron_init_outgoing}
\end{center}
\vskip -0.2in
\end{figure*}

\begin{figure*}
\vskip 0.2in
\begin{center}
{
\includegraphics[width=0.19\columnwidth]{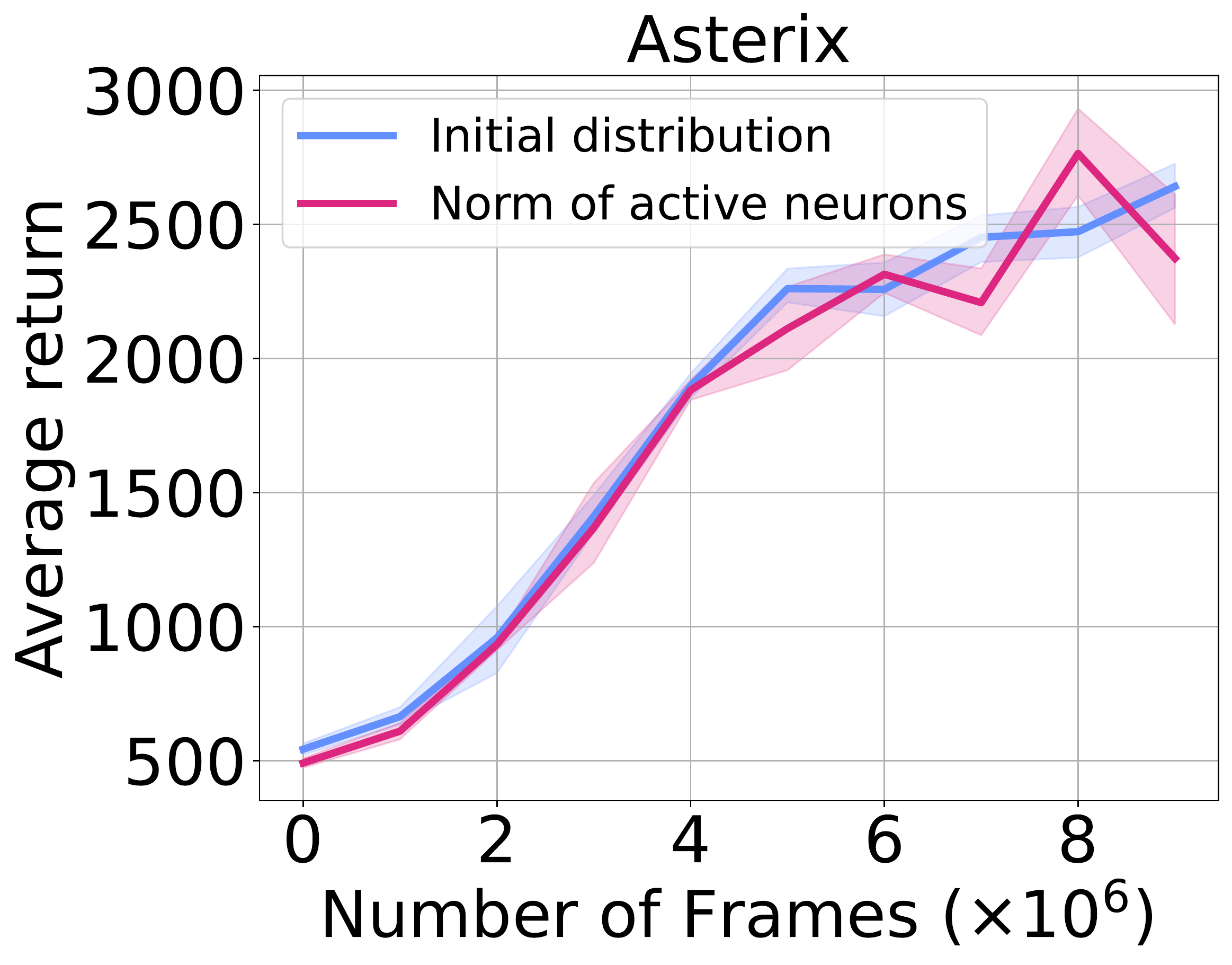}
\includegraphics[width=0.19\columnwidth]{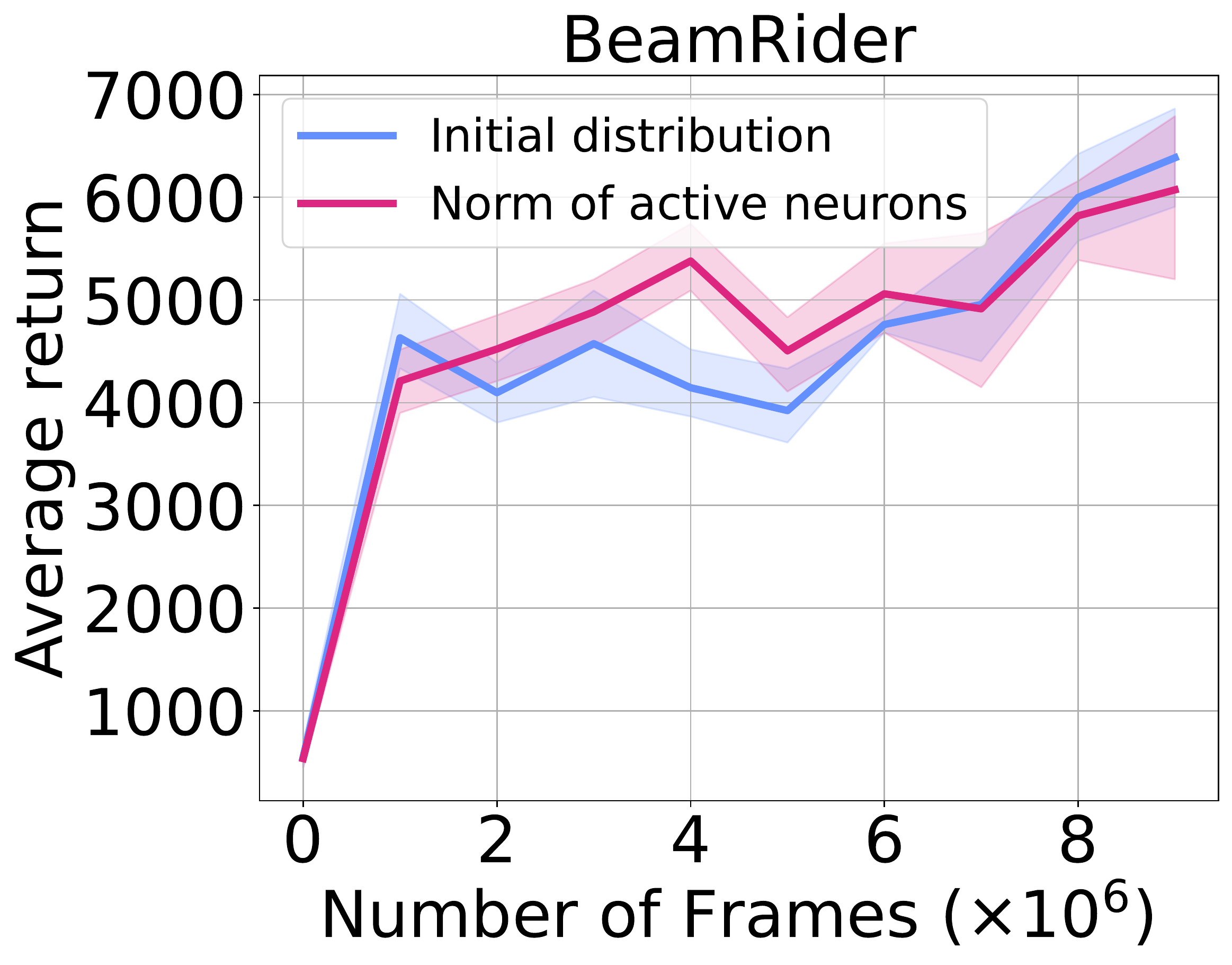}
\includegraphics[width=0.19\columnwidth]{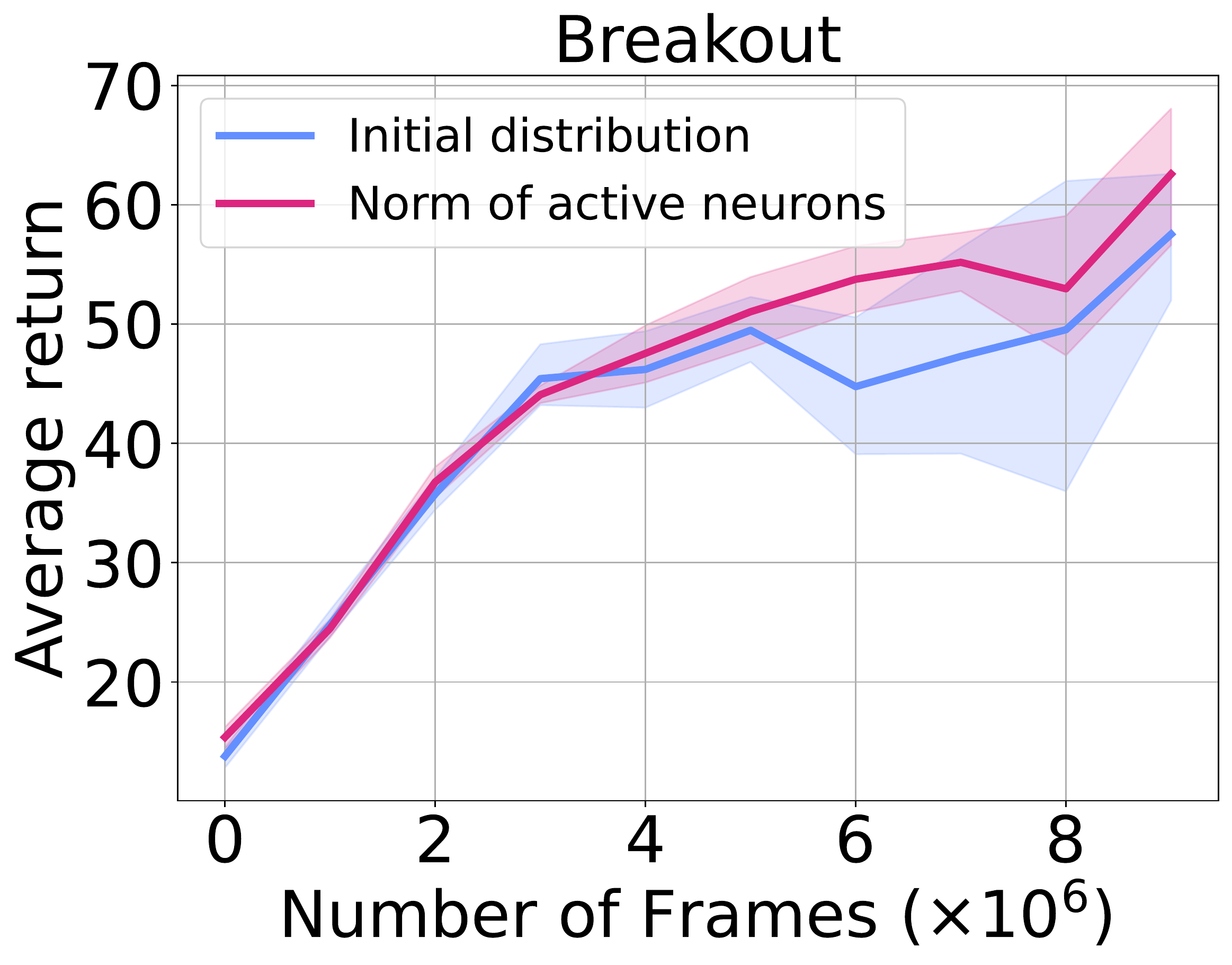}
\includegraphics[width=0.19\columnwidth]{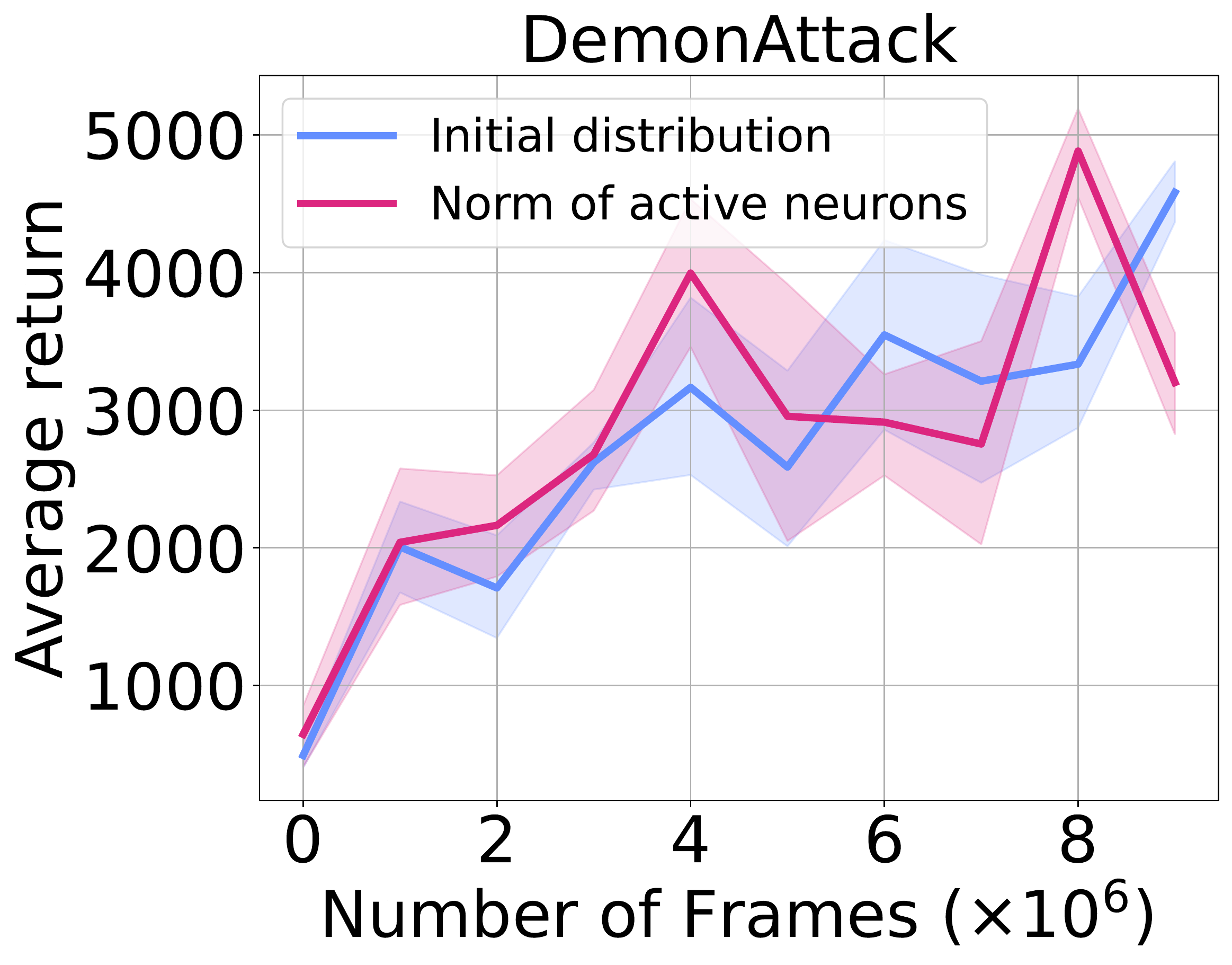}
\includegraphics[width=0.19\columnwidth]{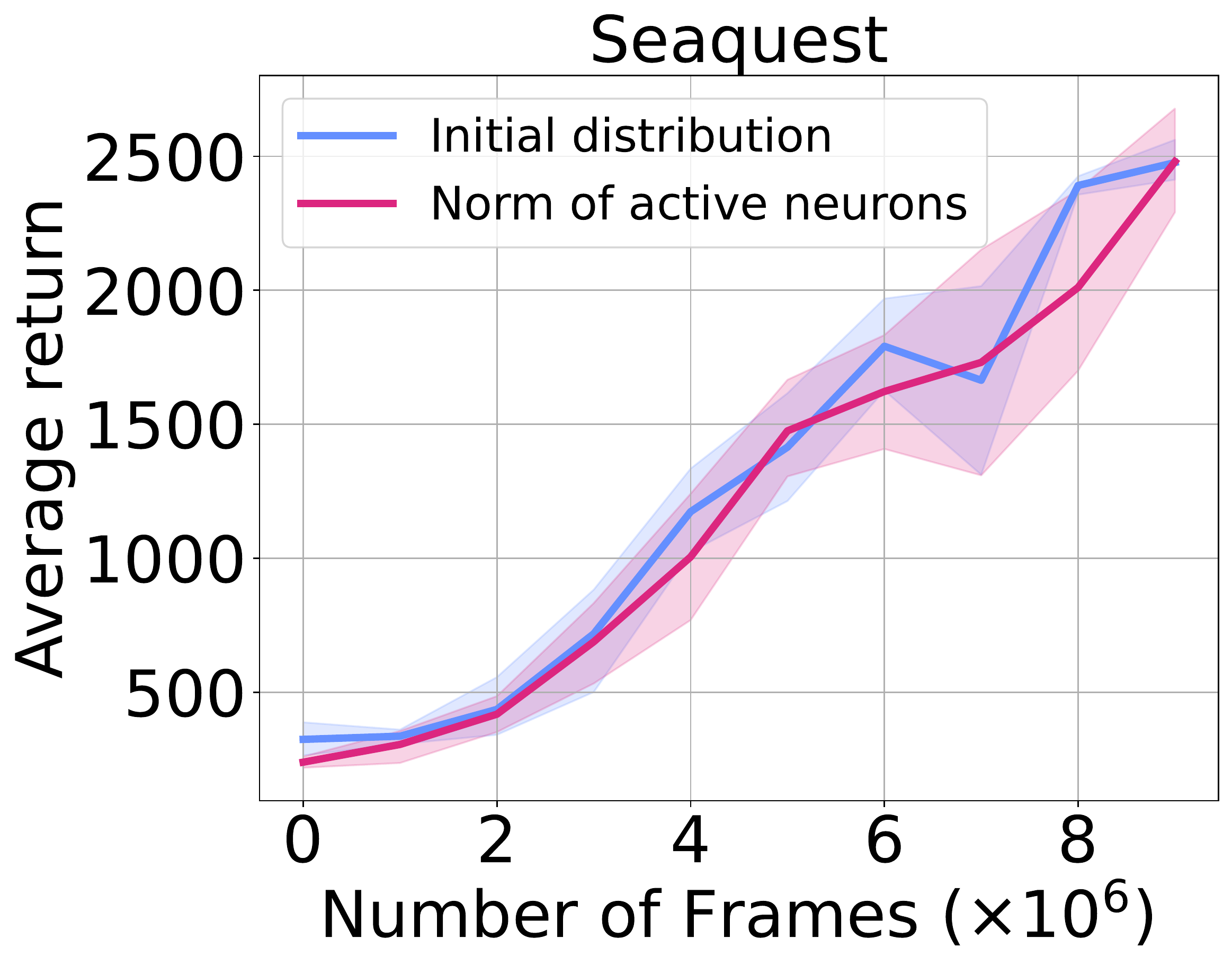} 

}
\caption{Comparison of performance with different strategies of reinitializing the incoming connections of dormant neurons.}
\label{fig:neuron_init_incoming}
\end{center}
\vskip -0.2in
\end{figure*}

\subsection{Recycling Strategies}
\label{app:recycling_strategies}
\paragraph{Outgoing connections.} We investigate the effect of using random weights to reinitialize the outgoing connections of dormant neurons. We compare this strategy against the reinitialization strategy of {\em ReDo} ({\em zero weights}). \autoref{fig:neuron_init_outgoing} shows the performance of DQN on five Atari games. The random initialization of the outgoing connections leads to a lower performance than the zero initialization. This is because the newly added random weights change the output of the network.   

\paragraph{Incoming connections.} Another possible strategy to reinitialize the incoming connections of dormant neurons is to scale their weights with the average norm of non-dormant neurons in the same layer. We observe that this strategy has a similar performance to the random weight initialization strategy, as shown in \autoref{fig:neuron_init_incoming}.   

\subsection{Effect of Batch Size}
\begin{figure*}
\vskip 0.2in
\begin{center}
{
\includegraphics[width=0.19\columnwidth]{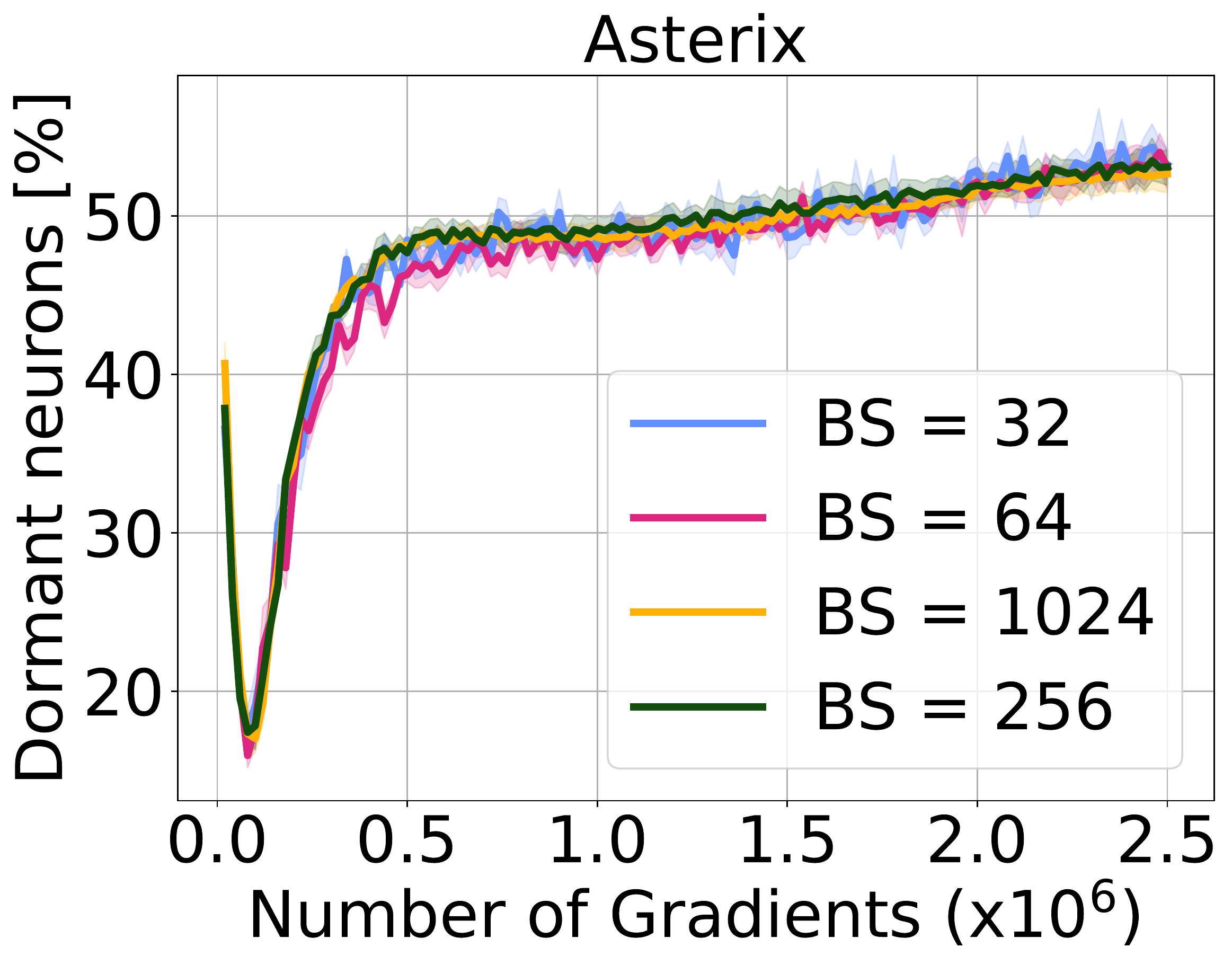}
\includegraphics[width=0.19\columnwidth]{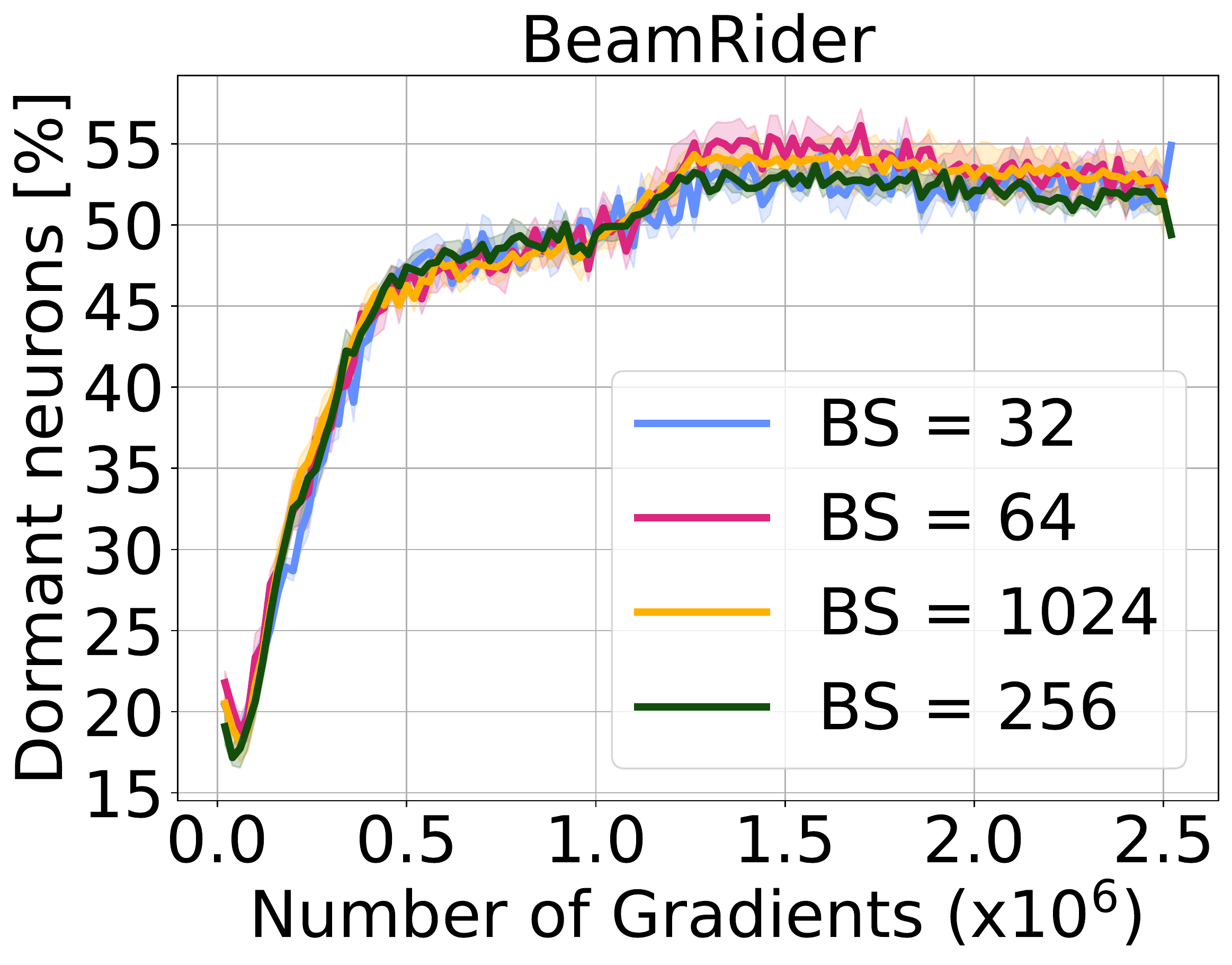}
\includegraphics[width=0.19\columnwidth]{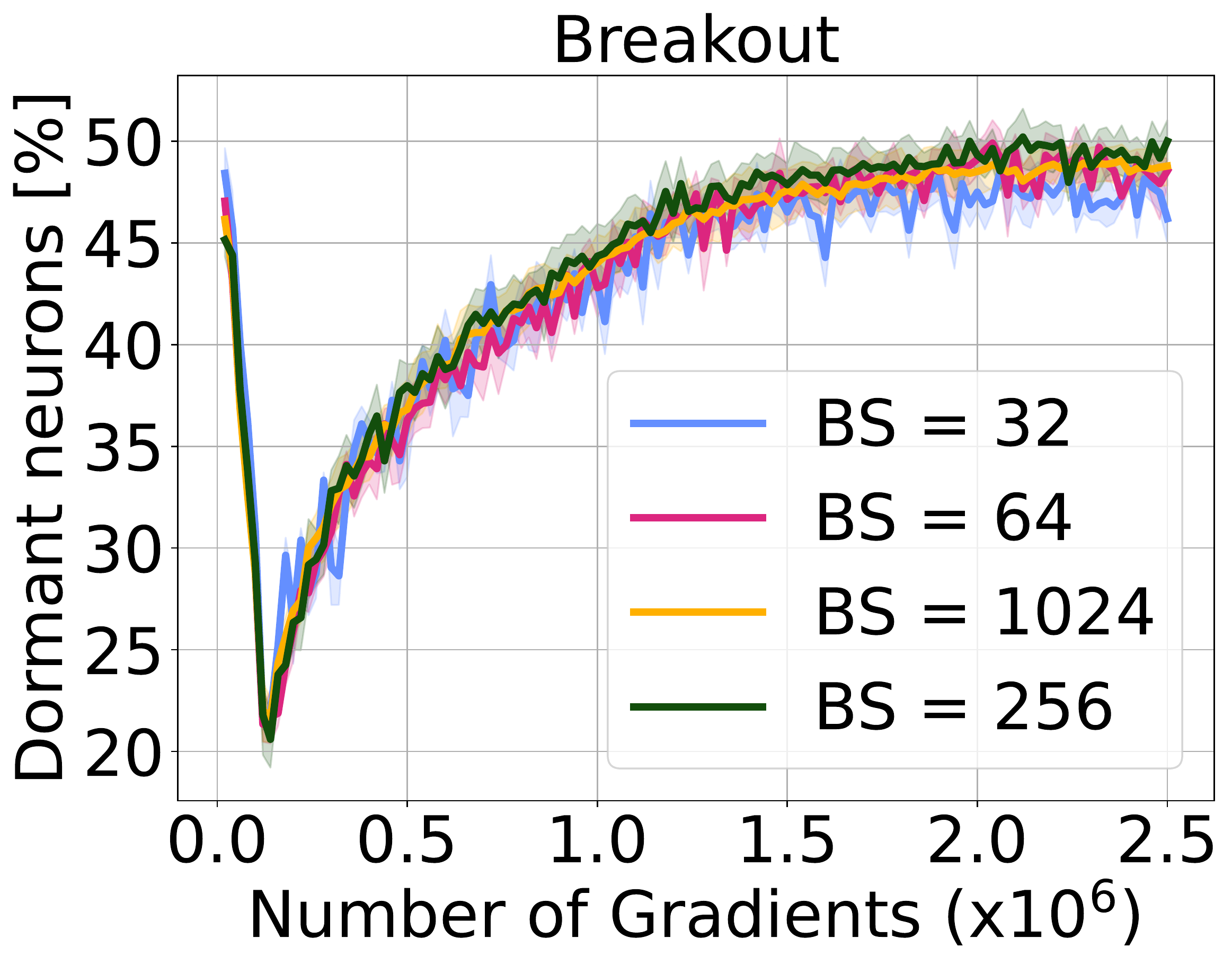}
\includegraphics[width=0.19\columnwidth]{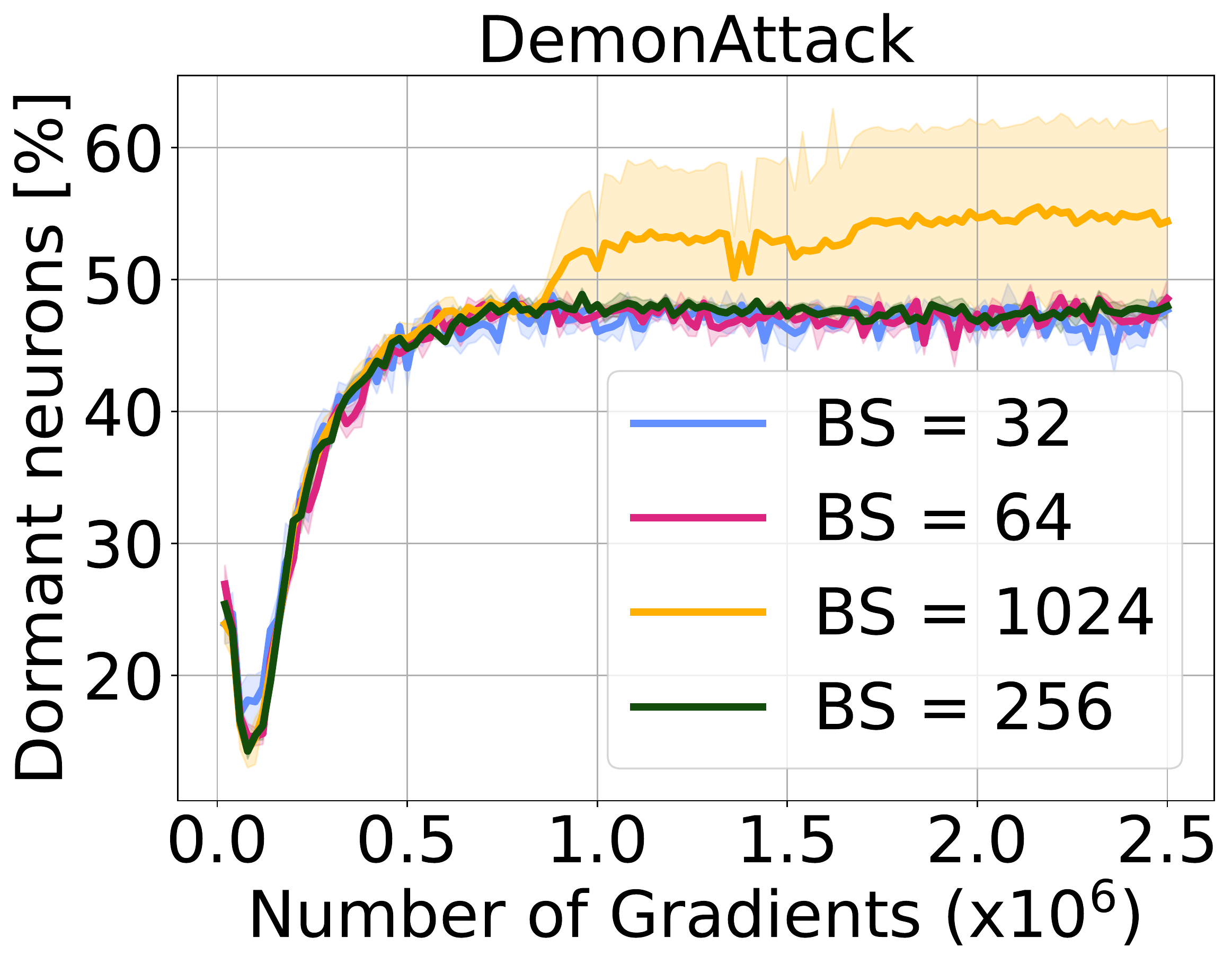}
\includegraphics[width=0.19\columnwidth]{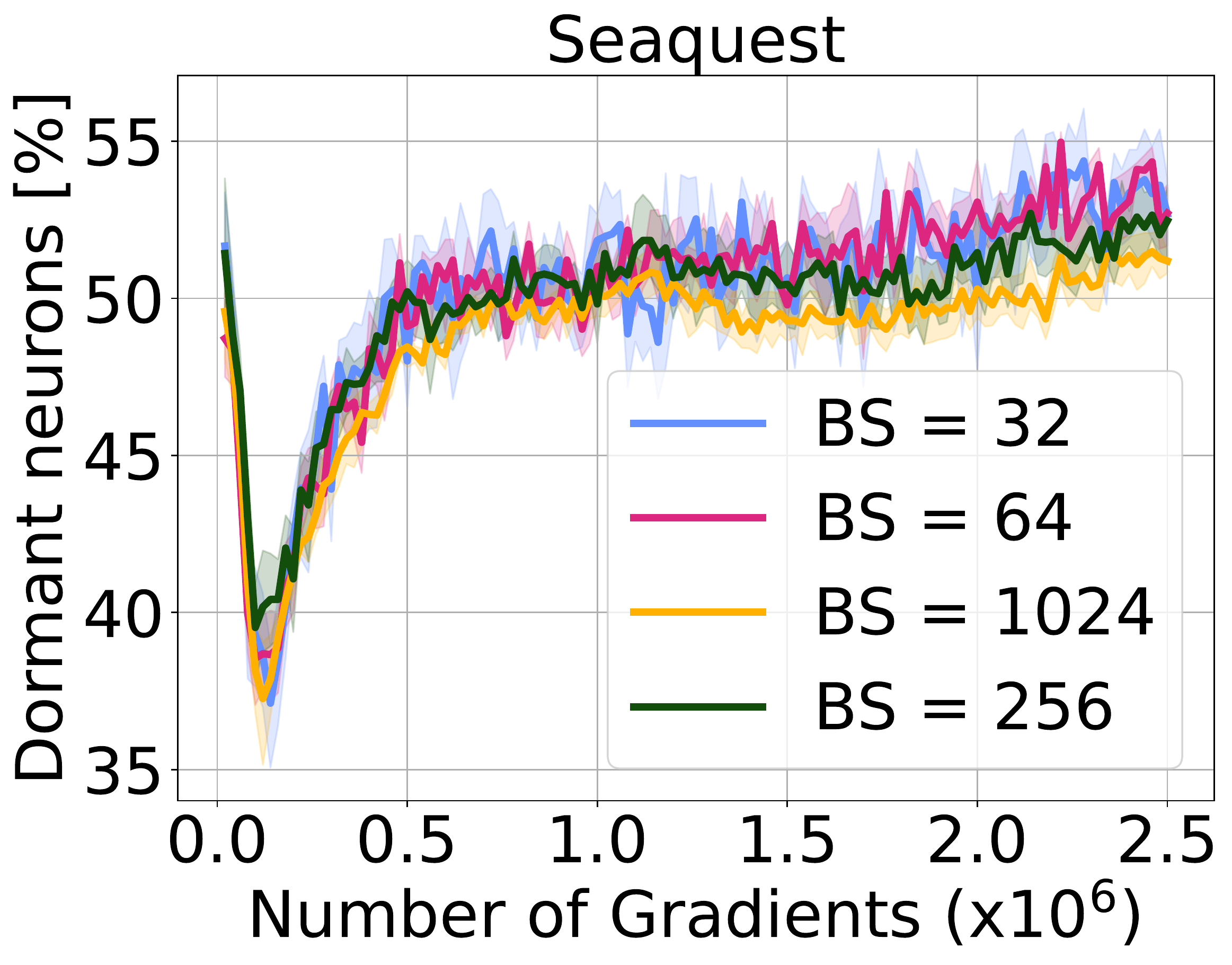}
}
\caption{Effect of the batch size used to detect dormant neurons.}
\label{fig:batch_size_effect}
\end{center}
\vskip -0.2in
\end{figure*}

The score of a neuron is calculated based on a given batch $\mathcal{D}$ of data (Section \ref{sec:analysis}). Here we study the effect of the batch size in determining the percentage of dormant neurons. We study four different values: \{32, 64, 256, 1024\}. \autoref{fig:batch_size_effect} shows that the identified percentage of dormant neurons is approximately the same using different batch sizes.

\subsection{Comparison with Continual Backprop}
\label{appendix:cbp}
\begin{figure}[ht]
\vskip 0.2in
\begin{center}
{
\includegraphics[width=0.27\columnwidth]{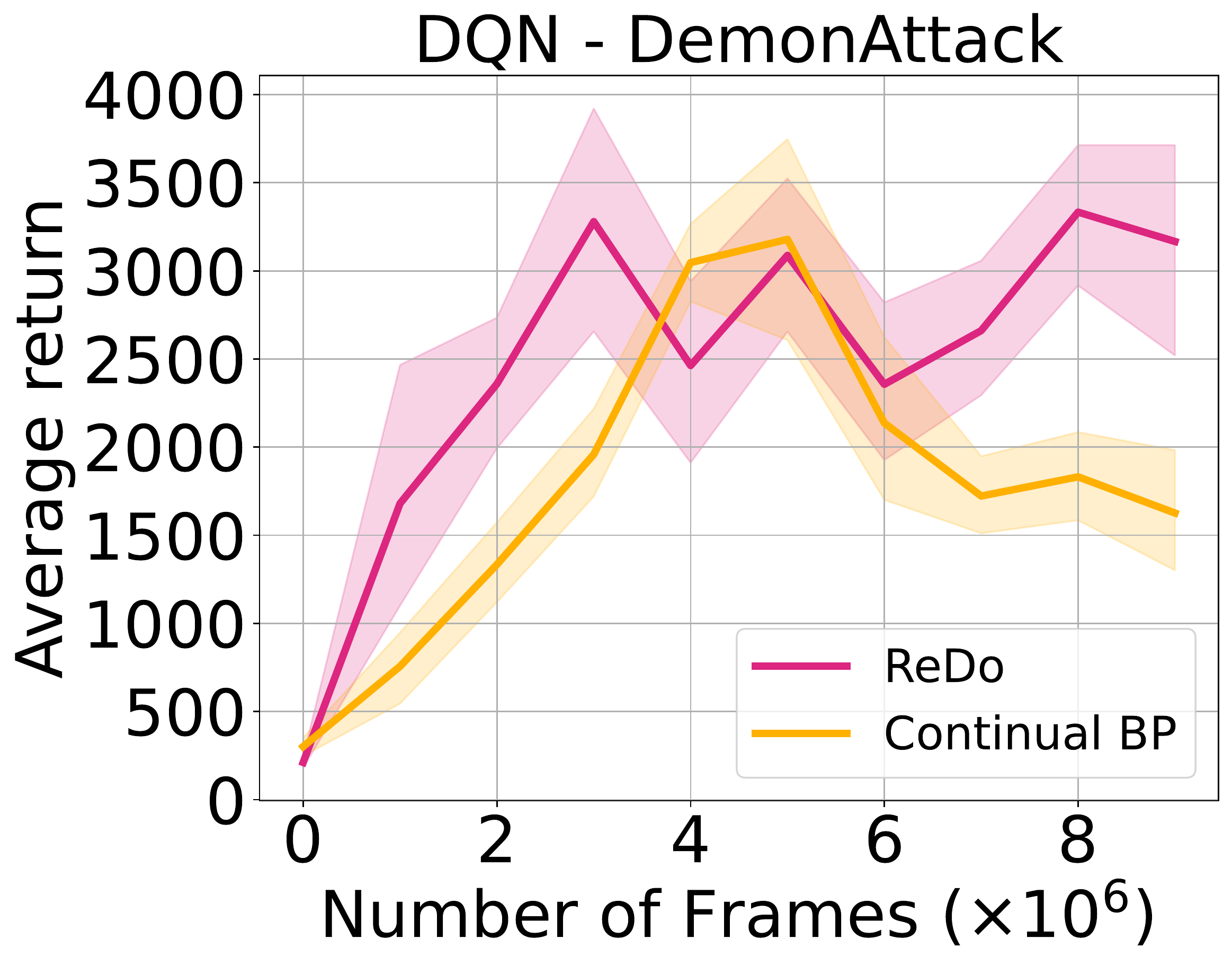}
\hspace{0.5cm}
\includegraphics[width=0.27\columnwidth]{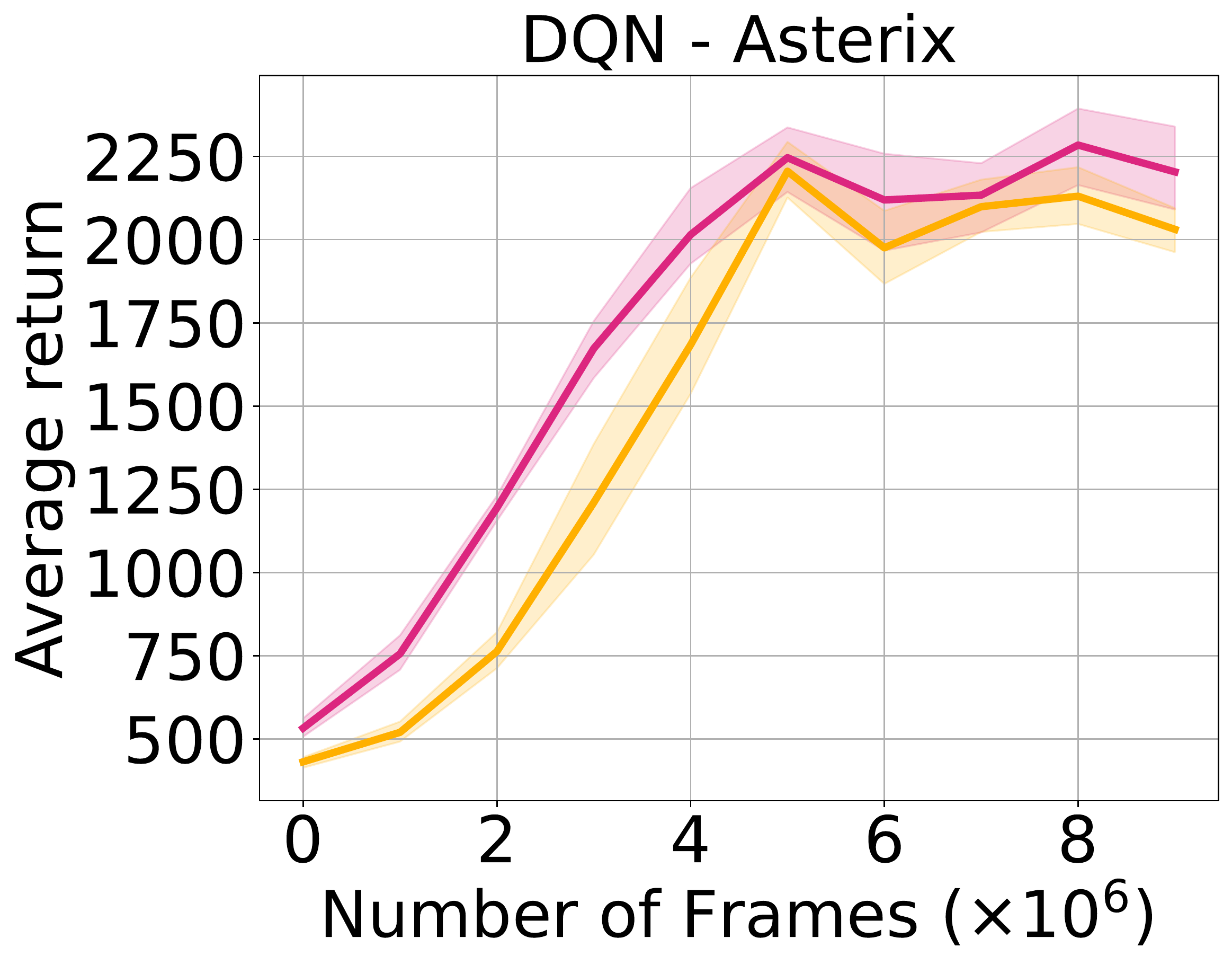}
}
\caption{Comparison of different strategies for selecting the recycled neurons.}
\label{fig:appendix:cbp}
\end{center}
\vskip -0.2in
\end{figure}

Similar to the experiments in \autoref{fig:different_scores}, we use a fixed recycling schedule to compare the activation-based metric used by {\em ReDo} and the utility metric proposed by Continual Backprop \citep{dohare2021continual}. Results shown in \autoref{fig:appendix:cbp} show that both metrics achieve similar results. Note that the original Continual Backprop algorithm calculates neuron scores at every iteration and uses a running average to obtain a better estimate of the neuron saliency. This approach requires additional storage and computing compared to the fixed schedule used by our algorithm. Given the high dormancy threshold preferred by our method (i.e., more neurons are recycled), we expect better saliency estimates to have a limited impact on the results presented here. However, a more thorough analysis is needed to make general conclusions.  

\subsection{Effect of Recycling the Dormant Capacity}

\begin{figure}[ht]
\vskip 0.2in
\begin{center}
{
\includegraphics[width=0.3\columnwidth]{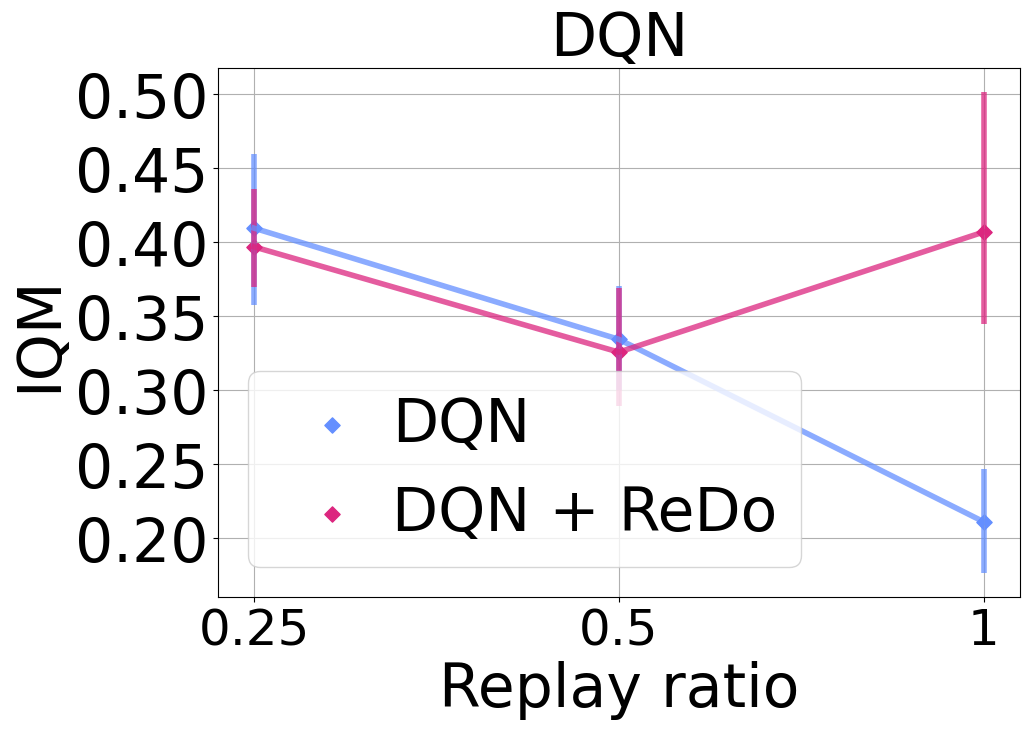}
}
\caption{Comparison of agents with varying replay ratios, while keeping the number of gradient updates constant.}
\label{fig:appendix:fixed_gradient_steps}
\end{center}
\vskip -0.2in
\end{figure}

\paragraph{Improving Sample Efficiency.} 
To examine the impact of recycling dormant neurons on enhancing the agents' sample efficiency, an alternative approach is to compare agents with varying replay ratios, while keeping the number of gradient updates constant during training. Consequently, agents with a higher replay ratio will perform fewer interactions with the environment. 

We performed this analysis on DQN and the 17 Atari games. Agents with a replay ratio of 0.25 run for 10M frames, a replay ratio of 0.5 run for 5M frames, and a replay ratio of 1 run for 2.5M frames. The number of gradient steps are fixed across all agents. \autoref{fig:appendix:fixed_gradient_steps} shows the aggregated results across all games. Interestingly the performance of {\em ReDo} with $RR = 1$ is very close to $RR = 0.25$, while significantly reducing the number of environment steps by four. On the other hand, DQN with $RR = 1$ suffers from a performance drop.  

\paragraph{Improving Networks' expressivity.} 
Our results in the main paper show that recycling dormant neurons improves the learning ability of agents measured by their performance. Here, we did some preliminary experiments to measure the effect of neuron recycling on the learned representations. Following \cite{kumar2020implicit}, we calculate the effective rank, a measure of expressivity, of the feature learned in the penultimate layer of networks trained with and without {\em ReDo}. We performed this analysis on agents trained for 10M frames on DemonAttack using DQN. The results are averaged over 5 seeds. The results in \autoref{table:srank} suggest recycling dormant neurons improves the expressivity, shown by the increased rank of the learned representations. Further investigation of expressivity metrics and analyses on other domains would be an exciting future direction.   

\begin{table}
\caption{Effective rank \cite{kumar2020implicit} of the learned representations of agents trained on DemonAttack.}
\label{table:srank}
\vskip 0.15in
\begin{center}
\begin{tabular}{lc}
\toprule
Agent & Effective rank \\
\midrule
DQN & 449.2 $\pm$ 5.77\\
DQN + {\em ReDo} &  \textbf{470.8} $\pm$ 1.16  \\
\bottomrule
\end{tabular}
\end{center}
\vskip -0.1in
\end{table}

\section{Performance Per Game}
\label{appendix:per_game_curves}
\begin{figure*}[ht]
\vskip 0.2in
\begin{center}
{
\includegraphics[width=\columnwidth]{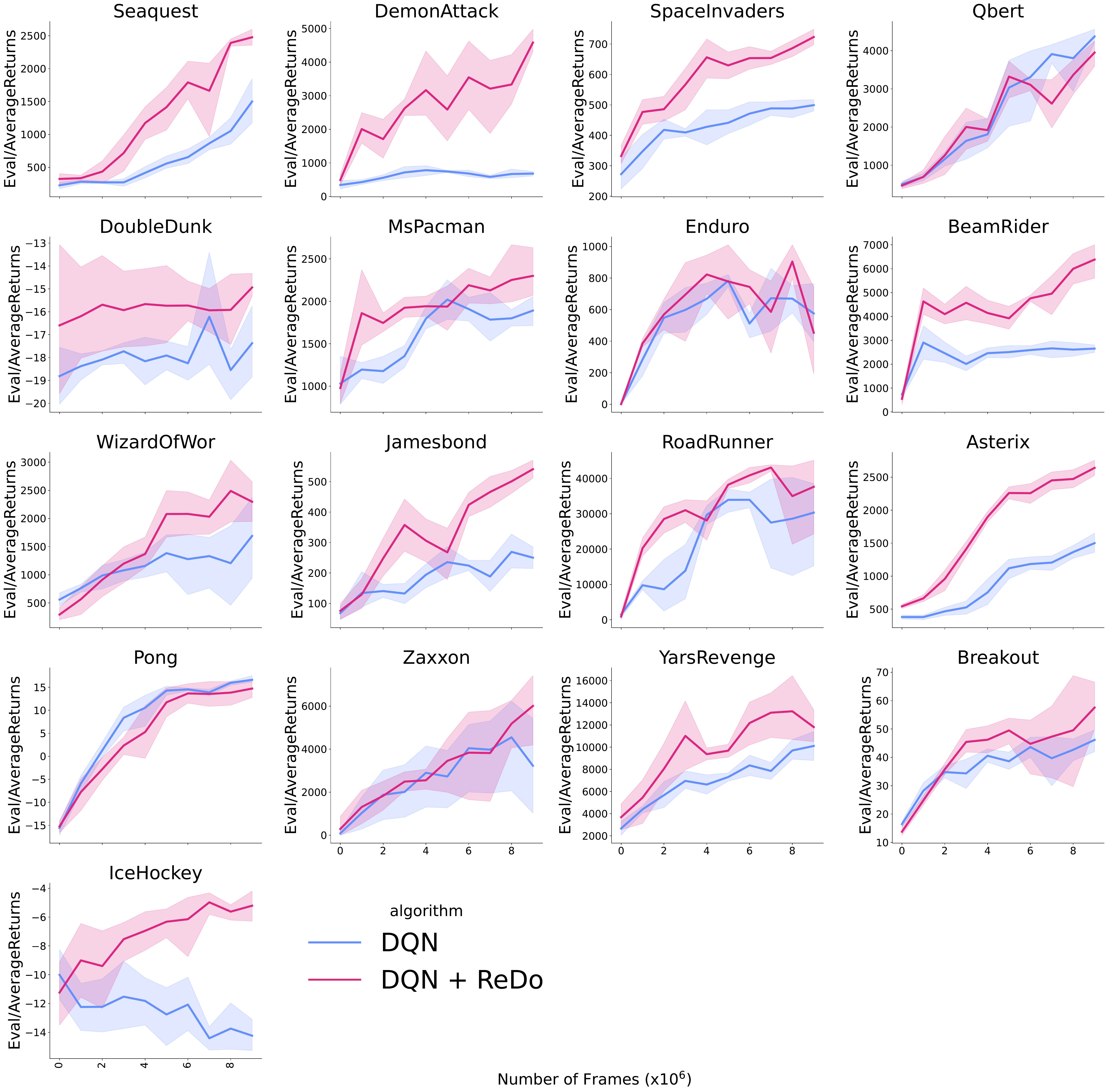}
}
\caption{Training curves for DQN with the nature CNN architecture ($RR = 1$).}
\label{fig:DQN_step1_per_game_rr1}
\end{center}
\vskip -0.2in
\end{figure*}

\begin{figure*}[ht]
\vskip 0.2in
\begin{center}
{
\includegraphics[width=\columnwidth]{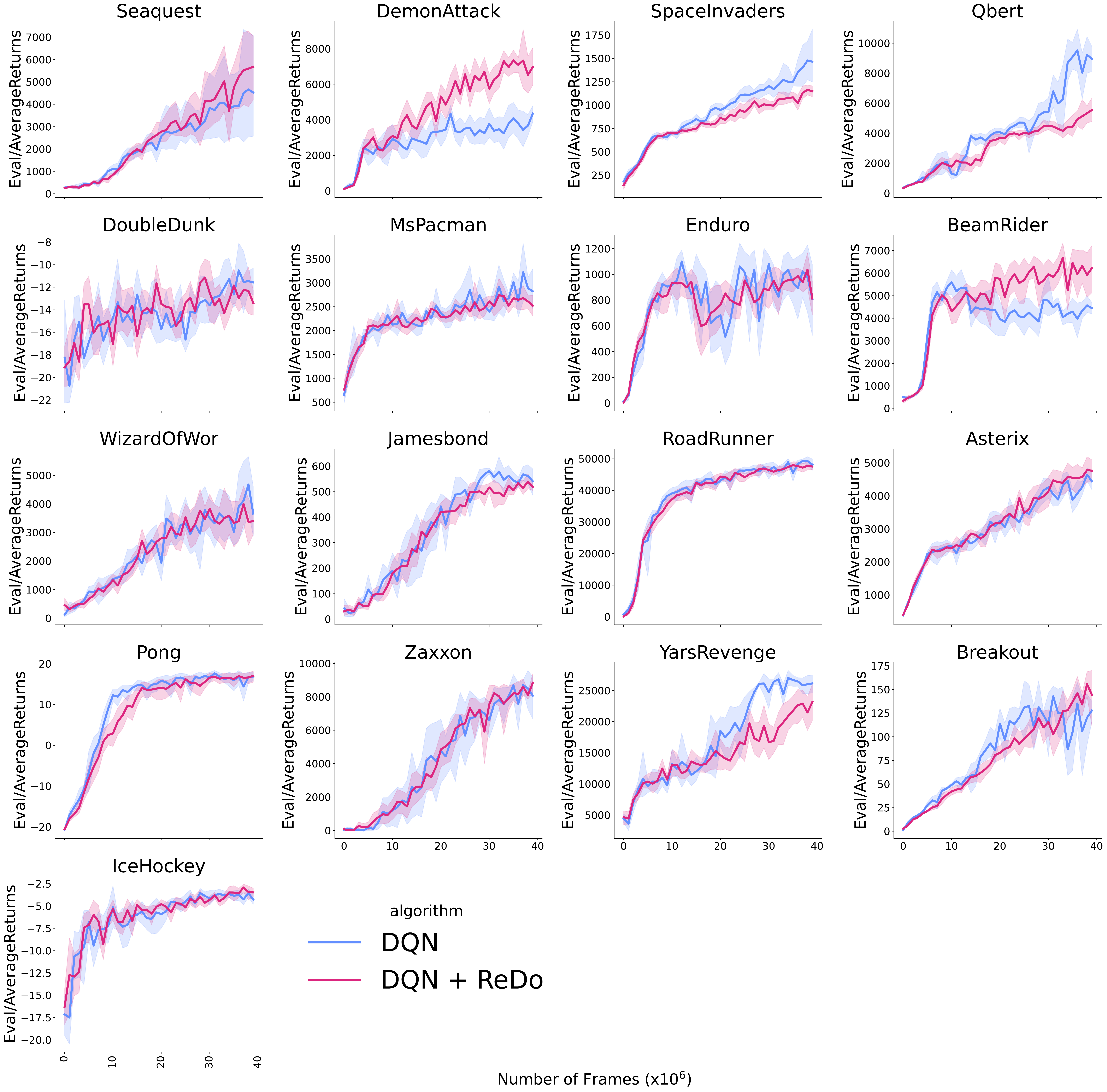}
}
\caption{Training curves for DQN with the nature CNN architecture ($RR = 0.25$).}
\label{fig:DQN_step1_per_game}
\end{center}
\vskip -0.2in
\end{figure*}

\begin{figure*}[ht]
\vskip 0.2in
\begin{center}
{
\includegraphics[width=\columnwidth]{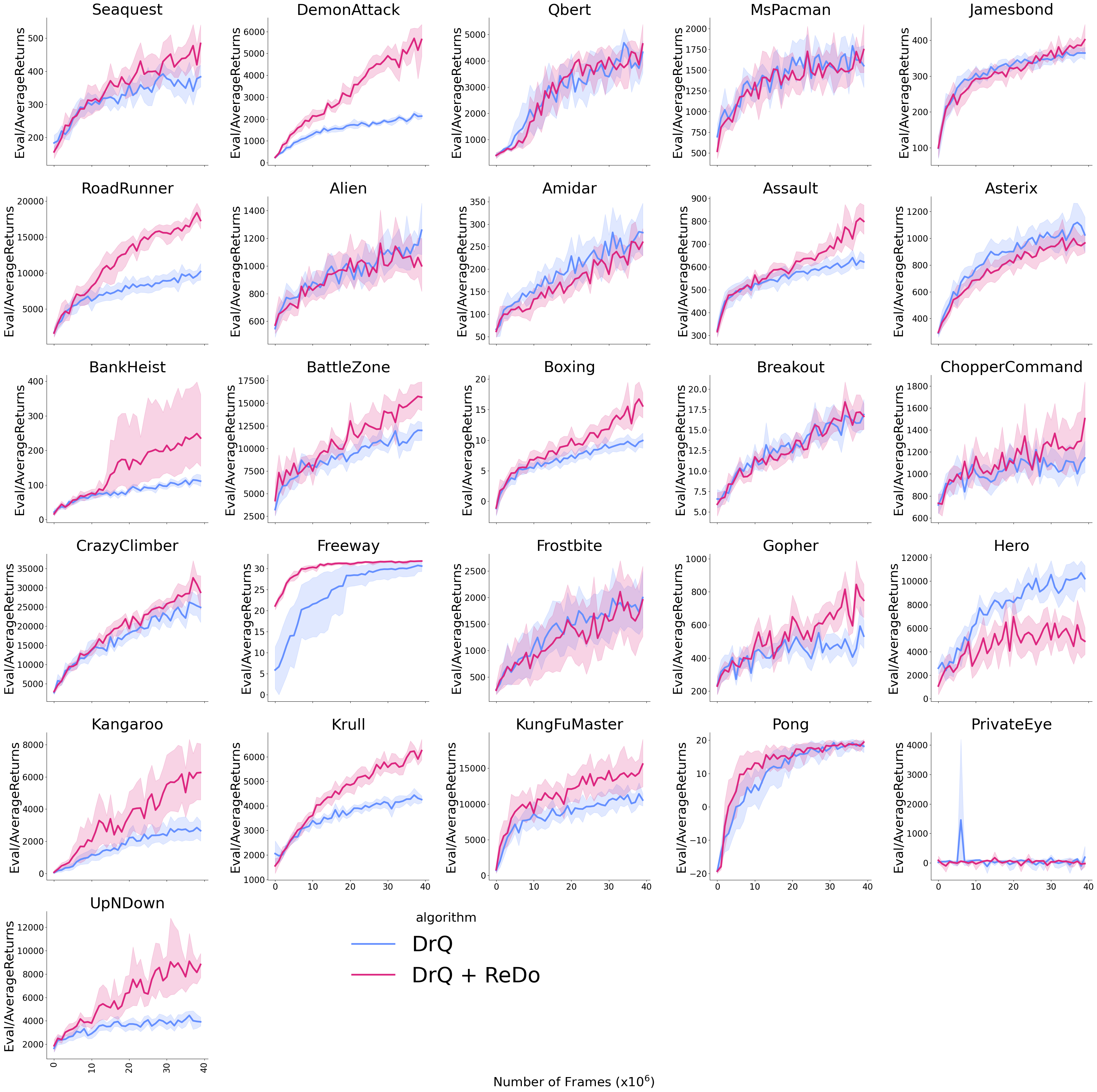}
}
\caption{Training curves for DrQ($\epsilon$) with the nature CNN architecture ($RR = 4$).}
\label{fig:DrQ_per_game_rr_4}
\end{center}
\vskip -0.2in
\end{figure*}

\begin{figure*}[ht]
\vskip 0.2in
\begin{center}
{
\includegraphics[width=\columnwidth]{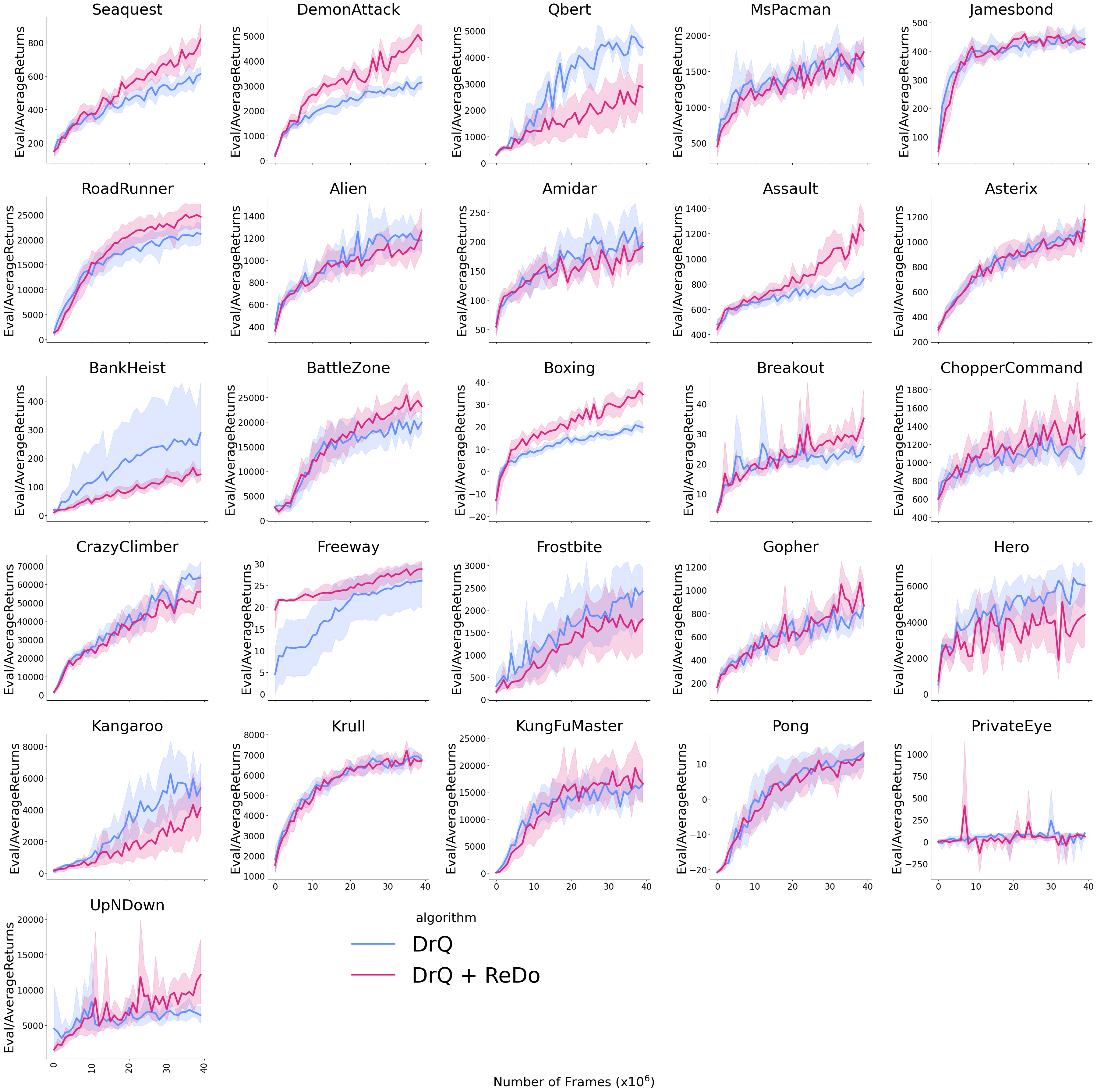}
}
\caption{Training curves for DrQ($\epsilon$) with the nature CNN architecture ($RR = 1$).}
\label{fig:DrQ_per_game_rr_1}
\end{center}
\vskip -0.2in
\end{figure*}

Here we share the training curves of DQN using the CNN architecture for each game in the high replay ratio regime ($RR = 1$) (\autoref{fig:DQN_step1_per_game_rr1}) and the default setting ($RR = 0.25$) (\autoref{fig:DQN_step1_per_game}). Similarly, \autoref{fig:DrQ_per_game_rr_4} and \ref{fig:DrQ_per_game_rr_1} show the training curves of DrQ($\epsilon$) for each game in the high replay ratio regime ($RR = 4$) and the default setting ($RR = 1$), respectively.  

%% file: main.bbl
\begin{thebibliography}{69}
\providecommand{\natexlab}[1]{#1}
\providecommand{\url}[1]{\texttt{#1}}
\expandafter\ifx\csname urlstyle\endcsname\relax
  \providecommand{\doi}[1]{doi: #1}\else
  \providecommand{\doi}{doi: \begingroup \urlstyle{rm}\Url}\fi

\bibitem[Agarwal et~al.(2020)Agarwal, Schuurmans, and
  Norouzi]{agarwal2020optimistic}
Agarwal, R., Schuurmans, D., and Norouzi, M.
\newblock An optimistic perspective on offline reinforcement learning.
\newblock In \emph{International Conference on Machine Learning}, pp.\
  104--114. PMLR, 2020.

\bibitem[Agarwal et~al.(2021)Agarwal, Schwarzer, Castro, Courville, and
  Bellemare]{agarwal2021deep}
Agarwal, R., Schwarzer, M., Castro, P.~S., Courville, A.~C., and Bellemare, M.
\newblock Deep reinforcement learning at the edge of the statistical precipice.
\newblock \emph{Advances in neural information processing systems},
  34:\penalty0 29304--29320, 2021.

\bibitem[Alabdulmohsin et~al.(2021)Alabdulmohsin, Maennel, and
  Keysers]{alabdulmohsin2021impact}
Alabdulmohsin, I., Maennel, H., and Keysers, D.
\newblock The impact of reinitialization on generalization in convolutional
  neural networks.
\newblock \emph{arXiv preprint arXiv:2109.00267}, 2021.

\bibitem[Ara{\'u}jo et~al.(2021)Ara{\'u}jo, Ceron, and
  Castro]{araujo2021lifting}
Ara{\'u}jo, J. G.~M., Ceron, J. S.~O., and Castro, P.~S.
\newblock Lifting the veil on hyper-parameters for value-based deep
  reinforcement learning.
\newblock In \emph{Deep RL Workshop NeurIPS 2021}, 2021.
\newblock URL \url{https://openreview.net/forum?id=Ws4v7nSqqb}.

\bibitem[Arnob et~al.(2021)Arnob, Ohib, Plis, and Precup]{arnob2021single}
Arnob, S.~Y., Ohib, R., Plis, S., and Precup, D.
\newblock Single-shot pruning for offline reinforcement learning.
\newblock \emph{arXiv preprint arXiv:2112.15579}, 2021.

\bibitem[Ash \& Adams(2020)Ash and Adams]{ash2020warm}
Ash, J. and Adams, R.~P.
\newblock On warm-starting neural network training.
\newblock \emph{Advances in Neural Information Processing Systems},
  33:\penalty0 3884--3894, 2020.

\bibitem[Bellemare et~al.(2013)Bellemare, Naddaf, Veness, and
  Bowling]{bellemare2013arcade}
Bellemare, M.~G., Naddaf, Y., Veness, J., and Bowling, M.
\newblock The arcade learning environment: An evaluation platform for general
  agents.
\newblock \emph{Journal of Artificial Intelligence Research}, 47:\penalty0
  253--279, 2013.

\bibitem[Bellemare et~al.(2020)Bellemare, Candido, Castro, Gong, Machado,
  Moitra, Ponda, and Wang]{bellemare2020autonomous}
Bellemare, M.~G., Candido, S., Castro, P.~S., Gong, J., Machado, M.~C., Moitra,
  S., Ponda, S.~S., and Wang, Z.
\newblock Autonomous navigation of stratospheric balloons using reinforcement
  learning.
\newblock \emph{Nature}, 588\penalty0 (7836):\penalty0 77--82, 2020.

\bibitem[Benedetti \& Couillard-Despres(2022)Benedetti and
  Couillard-Despres]{benedetti2022would}
Benedetti, B. and Couillard-Despres, S.
\newblock Why would the brain need dormant neuronal precursors?
\newblock \emph{Frontiers in Neuroscience}, 16, 2022.

\bibitem[Benedetti et~al.(2020)Benedetti, Dannehl, K{\"o}nig, Coviello,
  Kreutzer, Zaunmair, Jakubecova, Weiger, Aigner, Nacher,
  et~al.]{benedetti2020functional}
Benedetti, B., Dannehl, D., K{\"o}nig, R., Coviello, S., Kreutzer, C.,
  Zaunmair, P., Jakubecova, D., Weiger, T.~M., Aigner, L., Nacher, J., et~al.
\newblock Functional integration of neuronal precursors in the adult murine
  piriform cortex.
\newblock \emph{Cerebral cortex}, 30\penalty0 (3):\penalty0 1499--1515, 2020.

\bibitem[Bengio et~al.(2020)Bengio, Pineau, and Precup]{bengio2020interference}
Bengio, E., Pineau, J., and Precup, D.
\newblock Interference and generalization in temporal difference learning.
\newblock In \emph{International Conference on Machine Learning}, pp.\
  767--777. PMLR, 2020.

\bibitem[Berariu et~al.(2021)Berariu, Czarnecki, De, Bornschein, Smith,
  Pascanu, and Clopath]{DBLP:journals/corr/abs-2106-00042}
Berariu, T., Czarnecki, W., De, S., Bornschein, J., Smith, S.~L., Pascanu, R.,
  and Clopath, C.
\newblock A study on the plasticity of neural networks.
\newblock \emph{CoRR}, abs/2106.00042, 2021.
\newblock URL \url{https://arxiv.org/abs/2106.00042}.

\bibitem[Bradbury et~al.(2018)Bradbury, Frostig, Hawkins, Johnson, Leary,
  Maclaurin, Necula, Paszke, VanderPlas, Wanderman-Milne,
  et~al.]{bradbury2018jax}
Bradbury, J., Frostig, R., Hawkins, P., Johnson, M.~J., Leary, C., Maclaurin,
  D., Necula, G., Paszke, A., VanderPlas, J., Wanderman-Milne, S., et~al.
\newblock Jax: composable transformations of python+ numpy programs.
\newblock 2018.

\bibitem[Castro et~al.(2018)Castro, Moitra, Gelada, Kumar, and
  Bellemare]{castro18dopamine}
Castro, P.~S., Moitra, S., Gelada, C., Kumar, S., and Bellemare, M.~G.
\newblock Dopamine: {A} {R}esearch {F}ramework for {D}eep {R}einforcement
  {L}earning.
\newblock 2018.
\newblock URL \url{http://arxiv.org/abs/1812.06110}.

\bibitem[Chen et~al.(2020)Chen, Wang, Zhou, and Ross]{chen2020randomized}
Chen, X., Wang, C., Zhou, Z., and Ross, K.~W.
\newblock Randomized ensembled double q-learning: Learning fast without a
  model.
\newblock In \emph{International Conference on Learning Representations}, 2020.

\bibitem[Dai et~al.(2019)Dai, Yin, and Jha]{dai2019nest}
Dai, X., Yin, H., and Jha, N.~K.
\newblock Nest: A neural network synthesis tool based on a grow-and-prune
  paradigm.
\newblock \emph{IEEE Transactions on Computers}, 68\penalty0 (10):\penalty0
  1487--1497, 2019.

\bibitem[Dohare et~al.(2021)Dohare, Mahmood, and Sutton]{dohare2021continual}
Dohare, S., Mahmood, A.~R., and Sutton, R.~S.
\newblock Continual backprop: Stochastic gradient descent with persistent
  randomness.
\newblock \emph{arXiv preprint arXiv:2108.06325}, 2021.

\bibitem[Espeholt et~al.(2018)Espeholt, Soyer, Munos, Simonyan, Mnih, Ward,
  Doron, Firoiu, Harley, Dunning, et~al.]{espeholt2018impala}
Espeholt, L., Soyer, H., Munos, R., Simonyan, K., Mnih, V., Ward, T., Doron,
  Y., Firoiu, V., Harley, T., Dunning, I., et~al.
\newblock Impala: Scalable distributed deep-rl with importance weighted
  actor-learner architectures.
\newblock In \emph{International conference on machine learning}, pp.\
  1407--1416. PMLR, 2018.

\bibitem[Evci et~al.(2021)Evci, van Merrienboer, Unterthiner, Pedregosa, and
  Vladymyrov]{evci2021gradmax}
Evci, U., van Merrienboer, B., Unterthiner, T., Pedregosa, F., and Vladymyrov,
  M.
\newblock Gradmax: Growing neural networks using gradient information.
\newblock In \emph{International Conference on Learning Representations}, 2021.

\bibitem[Fan et~al.(2021)Fan, Wang, Huang, Yu, Fei-Fei, Zhu, and
  Anandkumar]{fan2021secant}
Fan, L., Wang, G., Huang, D.-A., Yu, Z., Fei-Fei, L., Zhu, Y., and Anandkumar,
  A.
\newblock Secant: Self-expert cloning for zero-shot generalization of visual
  policies.
\newblock In \emph{International Conference on Machine Learning}, pp.\
  3088--3099. PMLR, 2021.

\bibitem[Fedus et~al.(2020)Fedus, Ramachandran, Agarwal, Bengio, Larochelle,
  Rowland, and Dabney]{fedus2020revisiting}
Fedus, W., Ramachandran, P., Agarwal, R., Bengio, Y., Larochelle, H., Rowland,
  M., and Dabney, W.
\newblock Revisiting fundamentals of experience replay.
\newblock In \emph{International Conference on Machine Learning}, pp.\
  3061--3071. PMLR, 2020.

\bibitem[Fu et~al.(2019)Fu, Kumar, Soh, and Levine]{fu2019diagnosing}
Fu, J., Kumar, A., Soh, M., and Levine, S.
\newblock Diagnosing bottlenecks in deep q-learning algorithms.
\newblock In \emph{International Conference on Machine Learning}, pp.\
  2021--2030. PMLR, 2019.

\bibitem[Graesser et~al.(2022)Graesser, Evci, Elsen, and
  Castro]{graesser2022state}
Graesser, L., Evci, U., Elsen, E., and Castro, P.~S.
\newblock The state of sparse training in deep reinforcement learning.
\newblock In \emph{International Conference on Machine Learning}, pp.\
  7766--7792. PMLR, 2022.

\bibitem[Guadarrama et~al.(2018)Guadarrama, Korattikara, Ramirez, Castro,
  Holly, Fishman, Wang, Gonina, Wu, Kokiopoulou, Sbaiz, Smith, Bartók, Berent,
  Harris, Vanhoucke, and Brevdo]{TFAgents}
Guadarrama, S., Korattikara, A., Ramirez, O., Castro, P., Holly, E., Fishman,
  S., Wang, K., Gonina, E., Wu, N., Kokiopoulou, E., Sbaiz, L., Smith, J.,
  Bartók, G., Berent, J., Harris, C., Vanhoucke, V., and Brevdo, E.
\newblock {TF-Agents}: A library for reinforcement learning in tensorflow.
\newblock \url{https://github.com/tensorflow/agents}, 2018.
\newblock URL \url{https://github.com/tensorflow/agents}.
\newblock [Online; accessed 25-June-2019].

\bibitem[Gulcehre et~al.(2022)Gulcehre, Srinivasan, Sygnowski, Ostrovski,
  Farajtabar, Hoffman, Pascanu, and Doucet]{gulcehre2022empirical}
Gulcehre, C., Srinivasan, S., Sygnowski, J., Ostrovski, G., Farajtabar, M.,
  Hoffman, M., Pascanu, R., and Doucet, A.
\newblock An empirical study of implicit regularization in deep offline rl.
\newblock \emph{arXiv preprint arXiv:2207.02099}, 2022.

\bibitem[Haarnoja et~al.(2018)Haarnoja, Zhou, Abbeel, and
  Levine]{haarnoja2018soft}
Haarnoja, T., Zhou, A., Abbeel, P., and Levine, S.
\newblock Soft actor-critic: Off-policy maximum entropy deep reinforcement
  learning with a stochastic actor.
\newblock In \emph{International conference on machine learning}, pp.\
  1861--1870. PMLR, 2018.

\bibitem[Hansen et~al.(2021)Hansen, Su, and Wang]{NEURIPS2021_1e0f65eb}
Hansen, N., Su, H., and Wang, X.
\newblock Stabilizing deep q-learning with convnets and vision transformers
  under data augmentation.
\newblock In Ranzato, M., Beygelzimer, A., Dauphin, Y., Liang, P., and Vaughan,
  J.~W. (eds.), \emph{Advances in Neural Information Processing Systems},
  volume~34, pp.\  3680--3693. Curran Associates, Inc., 2021.
\newblock URL
  \url{https://proceedings.neurips.cc/paper/2021/file/1e0f65eb20acbfb27ee05ddc000b50ec-Paper.pdf}.

\bibitem[Harris et~al.(2020)Harris, Millman, Van Der~Walt, Gommers, Virtanen,
  Cournapeau, Wieser, Taylor, Berg, Smith, et~al.]{harris2020array}
Harris, C.~R., Millman, K.~J., Van Der~Walt, S.~J., Gommers, R., Virtanen, P.,
  Cournapeau, D., Wieser, E., Taylor, J., Berg, S., Smith, N.~J., et~al.
\newblock Array programming with numpy.
\newblock \emph{Nature}, 585\penalty0 (7825):\penalty0 357--362, 2020.

\bibitem[Hessel et~al.(2018)Hessel, Modayil, Van~Hasselt, Schaul, Ostrovski,
  Dabney, Horgan, Piot, Azar, and Silver]{hessel2018rainbow}
Hessel, M., Modayil, J., Van~Hasselt, H., Schaul, T., Ostrovski, G., Dabney,
  W., Horgan, D., Piot, B., Azar, M., and Silver, D.
\newblock Rainbow: Combining improvements in deep reinforcement learning.
\newblock In \emph{Thirty-second AAAI conference on artificial intelligence},
  2018.

\bibitem[Hestness et~al.(2017)Hestness, Narang, Ardalani, Diamos, Jun,
  Kianinejad, Patwary, Ali, Yang, and Zhou]{hestness2017deep}
Hestness, J., Narang, S., Ardalani, N., Diamos, G., Jun, H., Kianinejad, H.,
  Patwary, M., Ali, M., Yang, Y., and Zhou, Y.
\newblock Deep learning scaling is predictable, empirically.
\newblock \emph{arXiv preprint arXiv:1712.00409}, 2017.

\bibitem[Hiraoka et~al.(2021)Hiraoka, Imagawa, Hashimoto, Onishi, and
  Tsuruoka]{hiraoka2021dropout}
Hiraoka, T., Imagawa, T., Hashimoto, T., Onishi, T., and Tsuruoka, Y.
\newblock Dropout q-functions for doubly efficient reinforcement learning.
\newblock In \emph{International Conference on Learning Representations}, 2021.

\bibitem[Hunter(2007)]{hunter2007matplotlib}
Hunter, J.~D.
\newblock Matplotlib: A 2d graphics environment.
\newblock \emph{Computing in science \& engineering}, 9\penalty0 (03):\penalty0
  90--95, 2007.

\bibitem[Igl et~al.(2020)Igl, Farquhar, Luketina, Boehmer, and
  Whiteson]{igl2020transient}
Igl, M., Farquhar, G., Luketina, J., Boehmer, W., and Whiteson, S.
\newblock Transient non-stationarity and generalisation in deep reinforcement
  learning.
\newblock In \emph{International Conference on Learning Representations}, 2020.

\bibitem[Janner et~al.(2019)Janner, Fu, Zhang, and Levine]{janner2019trust}
Janner, M., Fu, J., Zhang, M., and Levine, S.
\newblock When to trust your model: Model-based policy optimization.
\newblock \emph{Advances in Neural Information Processing Systems}, 32, 2019.

\bibitem[Kaiser et~al.(2019)Kaiser, Babaeizadeh, Mi{\l}os, Osi{\'n}ski,
  Campbell, Czechowski, Erhan, Finn, Kozakowski, Levine,
  et~al.]{kaiser2019model}
Kaiser, {\L}., Babaeizadeh, M., Mi{\l}os, P., Osi{\'n}ski, B., Campbell, R.~H.,
  Czechowski, K., Erhan, D., Finn, C., Kozakowski, P., Levine, S., et~al.
\newblock Model based reinforcement learning for atari.
\newblock In \emph{International Conference on Learning Representations}, 2019.

\bibitem[Kaplan et~al.(2020)Kaplan, McCandlish, Henighan, Brown, Chess, Child,
  Gray, Radford, Wu, and Amodei]{kaplan2020scaling}
Kaplan, J., McCandlish, S., Henighan, T., Brown, T.~B., Chess, B., Child, R.,
  Gray, S., Radford, A., Wu, J., and Amodei, D.
\newblock Scaling laws for neural language models.
\newblock \emph{arXiv preprint arXiv:2001.08361}, 2020.

\bibitem[Kingma \& Ba(2015)Kingma and Ba]{KingmaB14}
Kingma, D.~P. and Ba, J.
\newblock Adam: {A} method for stochastic optimization.
\newblock In Bengio, Y. and LeCun, Y. (eds.), \emph{3rd International
  Conference on Learning Representations, {ICLR} 2015, San Diego, CA, USA, May
  7-9, 2015, Conference Track Proceedings}, 2015.
\newblock URL \url{http://arxiv.org/abs/1412.6980}.

\bibitem[Kirk et~al.(2021)Kirk, Zhang, Grefenstette, and
  Rockt{\"a}schel]{kirk2021survey}
Kirk, R., Zhang, A., Grefenstette, E., and Rockt{\"a}schel, T.
\newblock A survey of generalisation in deep reinforcement learning.
\newblock \emph{arXiv preprint arXiv:2111.09794}, 2021.

\bibitem[Krizhevsky et~al.(2009)Krizhevsky, Hinton,
  et~al.]{krizhevsky2009learning}
Krizhevsky, A., Hinton, G., et~al.
\newblock Learning multiple layers of features from tiny images.
\newblock 2009.

\bibitem[Kumar et~al.(2021{\natexlab{a}})Kumar, Agarwal, Ghosh, and
  Levine]{kumar2020implicit}
Kumar, A., Agarwal, R., Ghosh, D., and Levine, S.
\newblock Implicit under-parameterization inhibits data-efficient deep
  reinforcement learning.
\newblock In \emph{International Conference on Learning Representations},
  2021{\natexlab{a}}.

\bibitem[Kumar et~al.(2021{\natexlab{b}})Kumar, Agarwal, Ma, Courville, Tucker,
  and Levine]{kumar2021dr3}
Kumar, A., Agarwal, R., Ma, T., Courville, A., Tucker, G., and Levine, S.
\newblock Dr3: Value-based deep reinforcement learning requires explicit
  regularization.
\newblock In \emph{International Conference on Learning Representations},
  2021{\natexlab{b}}.

\bibitem[Li et~al.(2020)Li, Xiong, An, Xu, and Dou]{li2020rifle}
Li, X., Xiong, H., An, H., Xu, C.-Z., and Dou, D.
\newblock Rifle: Backpropagation in depth for deep transfer learning through
  re-initializing the fully-connected layer.
\newblock In \emph{International Conference on Machine Learning}, pp.\
  6010--6019. PMLR, 2020.

\bibitem[Lillicrap et~al.(2016)Lillicrap, Hunt, Pritzel, Heess, Erez, Tassa,
  Silver, and Wierstra]{lillicrap2016continuous}
Lillicrap, T.~P., Hunt, J.~J., Pritzel, A., Heess, N., Erez, T., Tassa, Y.,
  Silver, D., and Wierstra, D.
\newblock Continuous control with deep reinforcement learning.
\newblock In \emph{ICLR (Poster)}, 2016.

\bibitem[Lin(1992)]{lin1992self}
Lin, L.-J.
\newblock Self-improving reactive agents based on reinforcement learning,
  planning and teaching.
\newblock \emph{Machine learning}, 8\penalty0 (3):\penalty0 293--321, 1992.

\bibitem[Lyle et~al.(2021)Lyle, Rowland, and Dabney]{lyle2021understanding}
Lyle, C., Rowland, M., and Dabney, W.
\newblock Understanding and preventing capacity loss in reinforcement learning.
\newblock In \emph{International Conference on Learning Representations}, 2021.

\bibitem[Mnih et~al.(2015)Mnih, Kavukcuoglu, Silver, Rusu, Veness, Bellemare,
  Graves, Riedmiller, Fidjeland, Ostrovski, et~al.]{mnih2015human}
Mnih, V., Kavukcuoglu, K., Silver, D., Rusu, A.~A., Veness, J., Bellemare,
  M.~G., Graves, A., Riedmiller, M., Fidjeland, A.~K., Ostrovski, G., et~al.
\newblock Human-level control through deep reinforcement learning.
\newblock \emph{nature}, 518\penalty0 (7540):\penalty0 529--533, 2015.

\bibitem[Nikishin et~al.(2022)Nikishin, Schwarzer, D’Oro, Bacon, and
  Courville]{nikishin2022primacy}
Nikishin, E., Schwarzer, M., D’Oro, P., Bacon, P.-L., and Courville, A.
\newblock The primacy bias in deep reinforcement learning.
\newblock In \emph{International Conference on Machine Learning}, pp.\
  16828--16847. PMLR, 2022.

\bibitem[Oliphant(2007)]{4160250}
Oliphant, T.~E.
\newblock Python for scientific computing.
\newblock \emph{Computing in Science \& Engineering}, 9\penalty0 (3):\penalty0
  10--20, 2007.
\newblock \doi{10.1109/MCSE.2007.58}.

\bibitem[Puterman(2014)]{puterman2014markov}
Puterman, M.~L.
\newblock \emph{Markov decision processes: discrete stochastic dynamic
  programming}.
\newblock John Wiley \& Sons, 2014.

\bibitem[Rotheneichner et~al.(2018)Rotheneichner, Belles, Benedetti, K{\"o}nig,
  Dannehl, Kreutzer, Zaunmair, Engelhardt, Aigner, Nacher,
  et~al.]{rotheneichner2018cellular}
Rotheneichner, P., Belles, M., Benedetti, B., K{\"o}nig, R., Dannehl, D.,
  Kreutzer, C., Zaunmair, P., Engelhardt, M., Aigner, L., Nacher, J., et~al.
\newblock Cellular plasticity in the adult murine piriform cortex: continuous
  maturation of dormant precursors into excitatory neurons.
\newblock \emph{Cerebral Cortex}, 28\penalty0 (7):\penalty0 2610--2621, 2018.

\bibitem[Sankararaman et~al.(2020)Sankararaman, De, Xu, Huang, and
  Goldstein]{sankararaman2020impact}
Sankararaman, K.~A., De, S., Xu, Z., Huang, W.~R., and Goldstein, T.
\newblock The impact of neural network overparameterization on gradient
  confusion and stochastic gradient descent.
\newblock In \emph{International conference on machine learning}, pp.\
  8469--8479. PMLR, 2020.

\bibitem[Silver et~al.(2016)Silver, Huang, Maddison, Guez, Sifre, Van
  Den~Driessche, Schrittwieser, Antonoglou, Panneershelvam, Lanctot,
  et~al.]{silver2016mastering}
Silver, D., Huang, A., Maddison, C.~J., Guez, A., Sifre, L., Van Den~Driessche,
  G., Schrittwieser, J., Antonoglou, I., Panneershelvam, V., Lanctot, M.,
  et~al.
\newblock Mastering the game of go with deep neural networks and tree search.
\newblock \emph{nature}, 529\penalty0 (7587):\penalty0 484--489, 2016.

\bibitem[Sokar et~al.(2022)Sokar, Mocanu, Mocanu, Pechenizkiy, and
  Stone]{sokar2022dynamic}
Sokar, G., Mocanu, E., Mocanu, D.~C., Pechenizkiy, M., and Stone, P.
\newblock Dynamic sparse training for deep reinforcement learning.
\newblock In \emph{International Joint Conference on Artificial Intelligence},
  2022.

\bibitem[Sutton(1988)]{sutton1988learning}
Sutton, R.~S.
\newblock Learning to predict by the methods of temporal differences.
\newblock \emph{Machine learning}, 3\penalty0 (1):\penalty0 9--44, 1988.

\bibitem[Sutton \& Barto(2018)Sutton and Barto]{sutton2018reinforcement}
Sutton, R.~S. and Barto, A.~G.
\newblock \emph{Reinforcement learning: An introduction}.
\newblock MIT press, 2018.

\bibitem[Taha et~al.(2021)Taha, Shrivastava, and Davis]{taha2021knowledge}
Taha, A., Shrivastava, A., and Davis, L.~S.
\newblock Knowledge evolution in neural networks.
\newblock In \emph{Proceedings of the IEEE/CVF Conference on Computer Vision
  and Pattern Recognition}, pp.\  12843--12852, 2021.

\bibitem[Tan et~al.(2022)Tan, Hu, Pan, and Huang]{tan2022rlx2}
Tan, Y., Hu, P., Pan, L., and Huang, L.
\newblock Rlx2: Training a sparse deep reinforcement learning model from
  scratch.
\newblock \emph{arXiv preprint arXiv:2205.15043}, 2022.

\bibitem[Todorov et~al.(2012)Todorov, Erez, and Tassa]{todorov2012mujoco}
Todorov, E., Erez, T., and Tassa, Y.
\newblock Mujoco: A physics engine for model-based control.
\newblock In \emph{2012 IEEE/RSJ international conference on intelligent robots
  and systems}, pp.\  5026--5033. IEEE, 2012.

\bibitem[van Hasselt et~al.(2018)van Hasselt, Doron, Strub, Hessel, Sonnerat,
  and Modayil]{Hasselt2018deadlytriad}
van Hasselt, H., Doron, Y., Strub, F., Hessel, M., Sonnerat, N., and Modayil,
  J.
\newblock Deep reinforcement learning and the deadly triad.
\newblock \emph{CoRR}, abs/1812.02648, 2018.
\newblock URL \url{http://arxiv.org/abs/1812.02648}.

\bibitem[Van~Hasselt et~al.(2019)Van~Hasselt, Hessel, and
  Aslanides]{van2019use}
Van~Hasselt, H.~P., Hessel, M., and Aslanides, J.
\newblock When to use parametric models in reinforcement learning?
\newblock \emph{Advances in Neural Information Processing Systems}, 32, 2019.

\bibitem[Van~Rossum \& Drake~Jr(1995)Van~Rossum and Drake~Jr]{van1995python}
Van~Rossum, G. and Drake~Jr, F.~L.
\newblock \emph{Python reference manual}.
\newblock Centrum voor Wiskunde en Informatica Amsterdam, 1995.

\bibitem[Wang et~al.(2020)Wang, Kang, Shao, and Feng]{NEURIPS2020_5a751d6a}
Wang, K., Kang, B., Shao, J., and Feng, J.
\newblock Improving generalization in reinforcement learning with mixture
  regularization.
\newblock In Larochelle, H., Ranzato, M., Hadsell, R., Balcan, M., and Lin, H.
  (eds.), \emph{Advances in Neural Information Processing Systems}, volume~33,
  pp.\  7968--7978. Curran Associates, Inc., 2020.
\newblock URL
  \url{https://proceedings.neurips.cc/paper/2020/file/5a751d6a0b6ef05cfe51b86e5d1458e6-Paper.pdf}.

\bibitem[Wu et~al.(2019)Wu, Wang, and Liu]{wu2019splitting}
Wu, L., Wang, D., and Liu, Q.
\newblock Splitting steepest descent for growing neural architectures.
\newblock \emph{Advances in neural information processing systems}, 32, 2019.

\bibitem[Yarats et~al.(2021)Yarats, Kostrikov, and Fergus]{yarats2021image}
Yarats, D., Kostrikov, I., and Fergus, R.
\newblock Image augmentation is all you need: Regularizing deep reinforcement
  learning from pixels.
\newblock In \emph{International Conference on Learning Representations}, 2021.
\newblock URL \url{https://openreview.net/forum?id=GY6-6sTvGaf}.

\bibitem[Yoon et~al.(2018)Yoon, Yang, Lee, and Hwang]{yoon2018lifelong}
Yoon, J., Yang, E., Lee, J., and Hwang, S.~J.
\newblock Lifelong learning with dynamically expandable networks.
\newblock In \emph{International Conference on Learning Representations}, 2018.

\bibitem[Zaidi et~al.(2022)Zaidi, Berariu, Kim, Bornschein, Clopath, Teh, and
  Pascanu]{zaidi2022does}
Zaidi, S., Berariu, T., Kim, H., Bornschein, J., Clopath, C., Teh, Y.~W., and
  Pascanu, R.
\newblock When does re-initialization work?
\newblock \emph{arXiv preprint arXiv:2206.10011}, 2022.

\bibitem[Zhai et~al.(2022)Zhai, Kolesnikov, Houlsby, and
  Beyer]{zhai2022scaling}
Zhai, X., Kolesnikov, A., Houlsby, N., and Beyer, L.
\newblock Scaling vision transformers.
\newblock In \emph{Proceedings of the IEEE/CVF Conference on Computer Vision
  and Pattern Recognition}, pp.\  12104--12113, 2022.

\bibitem[Zhou et~al.(2012)Zhou, Sohn, and Lee]{zhou2012online}
Zhou, G., Sohn, K., and Lee, H.
\newblock Online incremental feature learning with denoising autoencoders.
\newblock In \emph{Artificial intelligence and statistics}, pp.\  1453--1461.
  PMLR, 2012.

\bibitem[Zhou et~al.(2021)Zhou, Vani, Larochelle, and
  Courville]{zhou2021fortuitous}
Zhou, H., Vani, A., Larochelle, H., and Courville, A.
\newblock Fortuitous forgetting in connectionist networks.
\newblock In \emph{International Conference on Learning Representations}, 2021.

\end{thebibliography}
